%% file: main.tex
\documentclass[10pt,twocolumn,letterpaper]{article}

\usepackage{iccv}
\usepackage{times}
\usepackage{epsfig}
\usepackage{graphicx}
\usepackage{amsmath}
\usepackage{amssymb}
\usepackage{booktabs}
\usepackage{subfigure}
\usepackage{xcolor,colortbl}
\usepackage{marvosym}
\usepackage{wrapfig}

\usepackage{mmstyle}

\usepackage[pagebackref=true,breaklinks=true,letterpaper=true,colorlinks,bookmarks=false]{hyperref}

\usepackage[capitalize]{cleveref}
\crefname{section}{Sec.}{Secs.}
\crefname{section}{Section}{Sections}
\crefname{table}{Table}{Tables}
\crefname{table}{Tab.}{Tabs.}

\iccvfinalcopy

\begin{document}

\title{RoboBEV: Towards Robust Bird's Eye View Perception under Corruptions}

\author{
Shaoyuan Xie$^{1}$\quad Lingdong Kong$^{2,3}$\quad Wenwei Zhang$^{2,4}$\quad Jiawei Ren$^{4}$ \\ Liang Pan$^{4}$\quad
Kai Chen$^{2}$\quad Ziwei Liu$^{4\textrm{\Letter}}$\\
\small{
$^{1}$~Huazhong University of Science and Technology\quad
$^{2}$~Shanghai AI Laboratory\quad $^{3}$~National University of Singapore}\\
\small{$^{4}$~S-Lab, Nanyang Technological University
}
\\
{\scriptsize \texttt{shaoyuanxie@hust.edu.cn} \quad \texttt{\{konglingdong,zhangwenwei,chenkai\}@pjlab.org.cn} \quad {\scriptsize\texttt{\{jiawei011,liang.pan,ziwei.liu\}@ntu.edu.sg}}
\vspace{-0.21cm}
}}

\maketitle

\begin{abstract}
The recent advances in camera-based bird's eye view (BEV) representation exhibit great potential for in-vehicle 3D perception. Despite the substantial progress achieved on standard benchmarks, the robustness of BEV algorithms has not been thoroughly examined, which is critical for safe operations. To bridge this gap, we introduce \textbf{RoboBEV}, a comprehensive benchmark suite that encompasses eight distinct corruptions, including \textit{Bright}, \textit{Dark}, \textit{Fog}, \textit{Snow}, \textit{Motion Blur}, \textit{Color Quant}, \textit{Camera Crash}, and \textit{Frame Lost}. Based on it, we undertake extensive evaluations across a wide range of BEV-based models to understand their resilience and reliability. Our findings indicate a strong correlation between absolute performance on in-distribution and out-of-distribution datasets. Nonetheless, there are considerable variations in relative performance across different approaches. Our experiments further demonstrate that pre-training and depth-free BEV transformation has the potential to enhance out-of-distribution robustness. Additionally, utilizing long and rich temporal information largely helps with robustness. Our findings provide valuable insights for designing future BEV models that can achieve both accuracy and robustness in real-world deployments. \footnote{Code: \url{https://github.com/Daniel-xsy/RoboBEV}.}
\end{abstract}

\input{sections/01_introduction}
\input{sections/02_related_work}

\input{sections/03_approach}
\input{sections/04_experiment}
\input{sections/05_discussion}
\input{sections/06_conclusion}

\input{sections/07_appendix}

\clearpage
\clearpage
{\small
\bibliographystyle{ieee_fullname}
\bibliography{ref}
}

\end{document}

%% file: sections/01_introduction.tex
\section{Introduction}
\label{sec:intro}
Deep neural network-based 3D perception methods have exhibited remarkable performance on various challenging benchmarks~\cite{li2022bevformer, wang2022detr3d, huang2021bevdet, liu2022petr, wang2022probabilistic, wang2021fcos3d, lang2019pointpillars, vora2020pointpainting, zhou2018voxelnet, yan2018second}. The camera-based approaches~\cite{li2022bevformer, wang2022detr3d, huang2021bevdet, liu2022petr, wang2022probabilistic, wang2021fcos3d} have garnered significant attention in comparison to LiDAR-based methods~\cite{lang2019pointpillars, vora2020pointpainting, zhou2018voxelnet, yan2018second}, owing to their low deployment cost, high computational efficiency, and dense semantic information. Moreover, building representations in the bird's eye view (BEV) offers several benefits. Firstly, it allows for joint learning from multi-view images. Secondly, the bird's eye perspective provides a physics-interpretable way for information fusion from different sensors and timestamps~\cite{ma2022vision}. Lastly, BEV output space is easily applicable in many downstream tasks, such as prediction and planning, which leads to significant improvements in performance for BEV-based perception frameworks.

\begin{figure}[t]
    \centering
    \includegraphics[width=\linewidth]{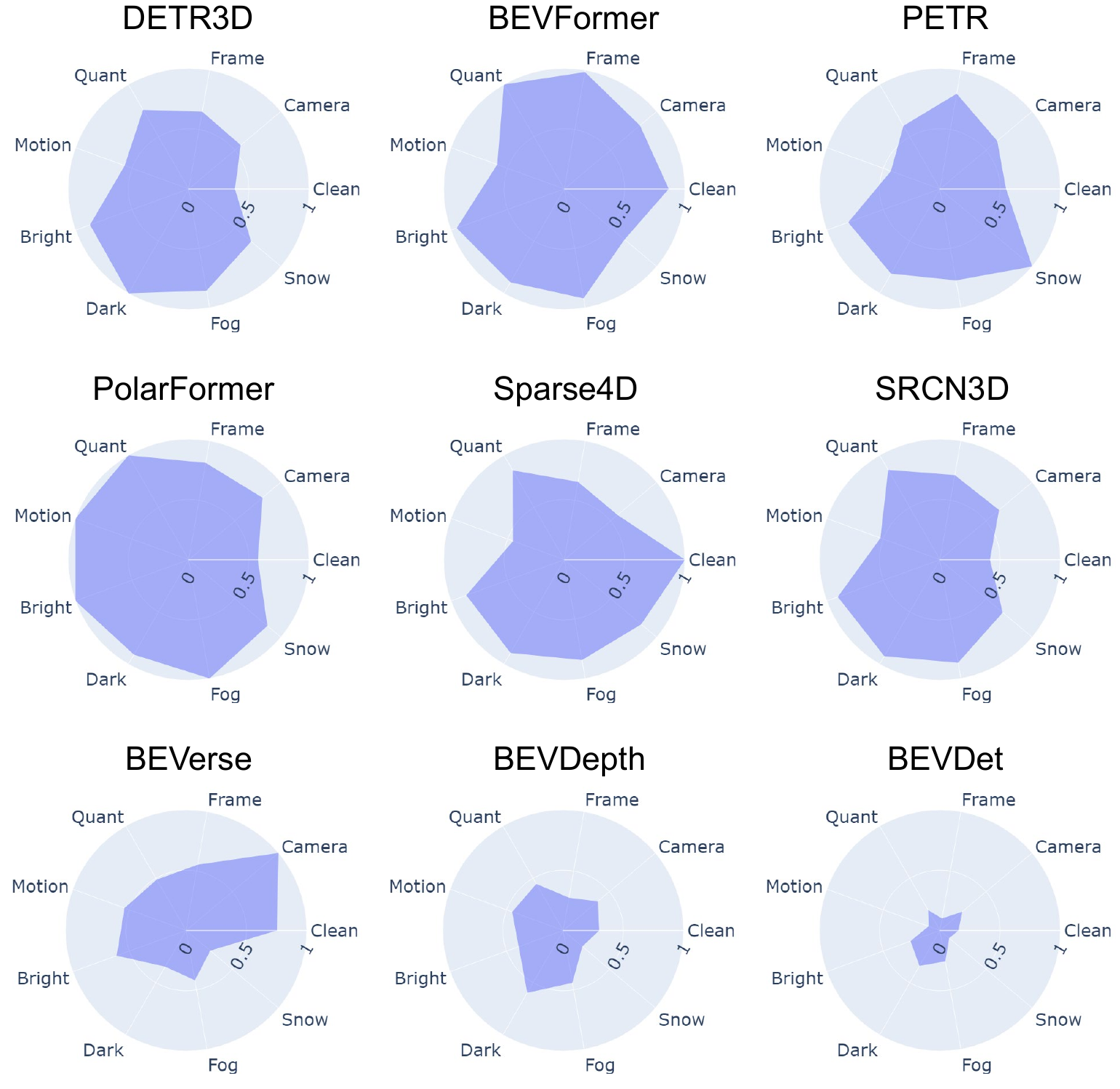}
    \caption{The radar charts of existing BEV detectors' nuScenes Detection Score (NDS)~\cite{caesar2020nuscenes} under eight corruption types. We observe diverse behaviors of different models even with competitive ``clean" performance. The NDS is normalized across all the benchmarking BEV models to lie between 0.1 and 1.}
    \label{fig:radar}
\end{figure}

\begin{figure*}[t]
    \centering
    \includegraphics[width=\linewidth]{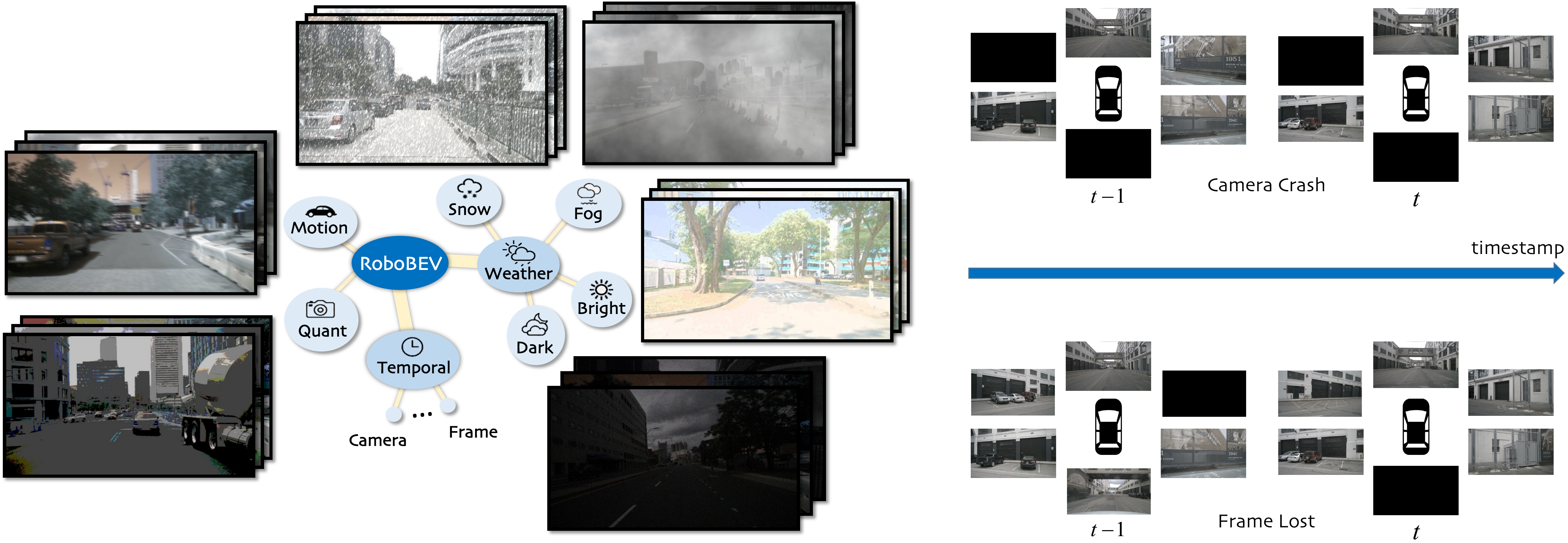}
    \caption{Corruption examples from the \textbf{\textit{RoboBEV}} benchmark. \textbf{Left:} Corruption taxonomy. \textbf{Right:} Temporal corruptions. \textit{Camera Crash} drop fixed set of images along timestamps; \textit{Frame Lost} randomly drop frames along timestamps. More examples are in Appendix~\ref{sec:app-more-visual}.}
    \label{fig:robobev}
\end{figure*}

Despite the remarkable progress achieved by recent BEV perception methods, their robustness against out-of-context scenarios remains inadequately understood. This is particularly concerning given that these methods are typically utilized in highly safety-critical systems (\eg, autonomous driving systems) in the real world. Generally, robustness can be divided into two categories: adversarial robustness~\cite{carlini2017towards} and robustness under natural corruptions~\cite{hendrycks2019benchmarking}. Very recently, Xie \textit{et al.} conduct a comprehensive study on the adversarial robustness of camera-based 3D detection models \cite{xie2023adversarial}, which measures the \textit{worst-case} performance but is difficult to implement in real-world settings. Conversely, we aim to investigate the robustness of camera-based 3D perception algorithms under natural corruptions, which measures \textit{average-case} performance that occurs frequently in reality.

In this work, to address the existing knowledge gap, we present a comprehensive benchmark dubbed \textbf{\textit{RoboBEV}}. This benchmark evaluates the robustness of camera-based BEV perceptions against natural corruptions including exterior environments, interior sensors, and temporal factors. Specifically, the exterior environments include various light and weather conditions, which are simulated by incorporating \textit{Brightness}, \textit{Dark}, \textit{Fog}, and \textit{Snow} weathers. Additionally, the inputs may be corrupted by interior factors caused by sensors, such as motion and color distortions. To this end, we consider \textit{Motion Blur} and \textit{Color Quant} to simulate these settings. Moreover, we propose two novel corruptions in temporal space tailored for BEV-based temporal fusion strategies, namely \textit{Camera Crash} and \textit{Frame Lost}. The study involves a comprehensive investigation of diverse out-of-distribution corruption settings that are highly relevant to real-world autonomous driving applications.

Leveraging the proposed \textit{RoboBEV} benchmark, we conduct an exhaustive analysis of 26 camera-based BEV perception models under 8 corruptions across 3 severities. The benchmark results, as illustrated in Figure~\ref{fig:radar}, demonstrate that models that exhibit competitive performance on standard datasets differ significantly in their performance on corrupted data and different corruptions. We mainly observe the following results: (a) the absolute performances show a strong correlation with the performances under corruption. However, the relative robustness does not necessarily increase as standard performance improves; (b) pre-training together with depth-free BEV transformation has great potential to enhance robustness; (c) utilizing long and rich temporal information largely helps with robustness. Our findings provide valuable insights for the design of future models to strive for better robustness under natural corruptions. The key contributions of this work are summarized as follows:

\begin{itemize}
\item To the best of our knowledge, we make the first attempt to introduce \textbf{\textit{RoboBEV}} for a comprehensive benchmark and evaluation of the robustness of camera-based BEV perceptions under natural corruptions.
\item We conduct exhaustive experiments to evaluate \textbf{26} existing camera-based BEV perception algorithms under \textbf{eight} corruptions across \textbf{three} severities.
\item Our study provides an in-depth analysis of the factors that contribute to superior robustness under corruption scenarios, which sheds light on future model design.
\end{itemize}
    

%% file: sections/02_related_work.tex
\section{Related Work}
\label{sec:rel-work}

\noindent\textbf{Camera-based Bird's Eye View Perception}. 
The existing BEV perception can be divided into two categories based on whether they perform depth estimation explicitly. Inspired by LSS~\cite{philion2020lift}, BEVDet~\cite{huang2021bevdet} utilizes an extra depth estimation branch to perform the transform from perspective view to bird's eye view (PV2BEV). BEVDepth~\cite{li2022bevdepth} makes more accurate depth estimation via explicit depth supervision from point clouds and further improves the performance. BEVerse~\cite{zhang2022beverse} proposes a unified framework for multi-task learning and achieving state-of-the-art performance.
Another line of work performs PV2BEV without explicit depth estimation. Following DETR~\cite{carion2020end}, DETR3D~\cite{wang2022detr3d} and ORA3D~\cite{roh2022ora3d} represent 3D objects as queries and perform cross-attention via Transformer decoder. PETR~\cite{liu2022petr} further improves performance by proposing 3D position-aware representations. BEVFormer~\cite{li2022bevformer} introduces temporal cross-attention to extract BEV features from multi-timestamps images. PolarFormer~\cite{jiang2022polarformer} explores predicting 3D targets in polar coordinates. Inspired by the success of Sparse RCNN~\cite{sun2021sparse}, SRCN3D~\cite{shi2022srcn3d} and Sparse4D~\cite{lin2022sparse4d} introduce sparse proposals for feature aggregation. SOLOFusion~\cite{Park2022TimeWT} attempts to fuse more history information in temporal modeling. Despite these works showing competitive results on the standard datasets, it's unclear how they behave in front of natural corruptions.

\noindent\textbf{Robustness under Adversarial Attacks}. Contemporary neural networks are prone to adversarial attacks, where a deliberately created perturbation added to the input can lead the network to produce incorrect predictions~\cite{szegedy2013intriguing,goodfellow2014explaining, moosavi2017universal}. Adversarial examples have been extensively researched in various vision tasks, including classification~\cite{szegedy2013intriguing,goodfellow2014explaining, moosavi2017universal,madry2017towards,brown2017adversarial,liu2016delving}, detection~\cite{xie2017adversarial,liu2018dpatch,tu2020physically,cao2021invisible}, and segmentation~\cite{xie2017adversarial, rossolini2022real}. These adversarial examples can be generated both in digital~\cite{szegedy2013intriguing,goodfellow2014explaining,moosavi2017universal,madry2017towards,brown2017adversarial,liu2016delving,xie2017adversarial} and physical environments~\cite{rossolini2022real,kurakin2018adversarial,cao2021invisible}. The recent study demonstrates that the 3D perception system tends to crash when exposed to adversarial examples, which could pose potential safety risks in the deployment stages~\cite{rossolini2022real, cao2021invisible, tu2020physically}. While Xie \etal~\cite{xie2023adversarial} conducts a comprehensive study of the adversarial robustness of camera-based detectors, we focus on natural corruptions, which are more likely to occur in the real world.

\noindent\textbf{Robustness under Natural Corruptions}. The evaluation of model robustness has garnered significant research attention. Various benchmarks have been proposed to evaluate the robustness of 2D image classification models, \eg, ImageNet-C~\cite{hendrycks2019benchmarking}, ObjectNet~\cite{barbu2019objectnet}, ImageNetV2~\cite{recht2019imagenet}, ImageNet-A~\cite{hendrycks2021natural}, and ImageNet-R~\cite{recht2019imagenet}. ImageNet-C intentionally corrupts the ``clean" ImageNet samples with simulated corruptions like compression loss and motion blur, while ObjectNet collects a test set with rich variations in rotation, background, and viewpoint. ImageNetV2 re-collects a test set following ImageNet's protocol and assesses the performance gap due to the natural distribution shift. ImageNet-A and ImageNet-R benchmark the classifier's robustness to natural adversarial examples and abstract visual renditions, respectively. Hendrycks \etal~\cite{hendrycks2021many} notes the correlation between the robustness of synthetic corruptions and the improvement of real-world cases. However, there is a lack of comparable benchmarks to evaluate the out-of-distribution robustness of the 3D BEV perception models, which are commonly employed in safety-critical applications. In this work, we present the first effort to investigate the robustness of these models under natural corruptions.

%% file: sections/03_approach.tex
\input{tables/robodet_model}
\section{RoboBEV Benchmark}
\label{sec:approach}

\subsection{Benchmark Design}
We present \textit{nuScenes-C} as our benchmark dataset, which is generated by corrupting the validation set of the nuScenes dataset~\cite{caesar2020nuscenes}. We choose nuScenes since it has been widely utilized among almost all the recent BEV models. Our benchmark dataset comprises \textbf{\textit{eight}} different corruptions, which include various exterior environments, interior sensor distortions, and the newly introduced temporal corruptions.
Following the experimental protocol established in \cite{hendrycks2019benchmarking}, we establish \textbf{\textit{three}} different levels of corruption intensity (\ie, easy, moderate, and hard) for each type of corruption. The severity levels are carefully determined to avoid significant performance drops that could undermine the conclusions' reliability. Additionally, we incorporate variations within each severity level of corruption to augment the benchmark's diversity.

\subsection{Natural Corruptions}
The corruption taxonomy is shown in Figure~\ref{fig:robobev}. Our first category of corruption involves diverse exterior environment conditions, such as lighting and extreme weather conditions. To simulate these situations, we introduce \textit{Brightness}, \textit{Dark}, \textit{Fog}, and \textit{Snow}. Since the majority of images in autonomous driving datasets are captured under mild meteorological conditions with temperate lighting, it is crucial to ensure that the perception system could perform well under different exterior conditions.

Secondly, sensor distortions are also prevalent in the real world. Images captured by cameras can easily become blurred at high moving speeds, and images can be quantized to reduce memory when deployed on devices with strict resource constraints. To simulate these scenarios, we incorporate \textit{Motion Blur} and \textit{Color Quantization}. 

Lastly, we consider the case where cameras may crash or drop certain frames due to physical problems. We simulate this scenario by introducing the \textit{Camera Crash} corruption, where we randomly discard images from a fixed set of cameras based on the corruption intensity. Additionally, we propose \textit{Frame Lost}, where we randomly drop multiple cameras independently with the same probability at each timestamp, emulating situations where some frames are lost in consecutive time sequences. Figure~\ref{fig:robobev} illustrates these procedures.

In addition, we conduct an analysis of the pixel histogram. It is worth noting that while \textit{Motion Blur} incurs the least amount of pixel distribution shift, it results in a significant reduction in performance. Further experimental results are provided in Section~\ref{sec:exp}. For more visualization results, please refer to Appendix~\ref{sec:app-corruptions}.

\subsection{Robustness Metrics}
We follow the official nuScenes metric~\cite{caesar2020nuscenes} to calculate robustness metrics on the \textit{nuScenes-C} dataset. We report nuScenes Detection Score (NDS) and mean Average Precision (mAP), along with mean Average Translation Error (mATE), mean Average Scale Error (mASE), mean Average Orientation Error (mAOE), mean Average Velocity Error (mAVE) and mean Average Attribute Error (mAAE).

To better compare the robustness among different BEV detectors, we introduce two new metrics inspired by \cite{hendrycks2019benchmarking} based on NDS. The first metric is the mean corruption error (mCE), which is applied to measure the \textit{relative robustness} of candidate models compared to the baseline model:
\begin{equation}
    \text{CE}_i = \frac{\sum^{3}_{l=1}(1 - \text{NDS})_{i,l}}{\sum^{3}_{l=1}(1 - \text{NDS}_{i,l}^{\text{baseline}})}~,~
    \text{mCE} = \frac{1}{N}\sum^N_{i=1}\text{CE}_i~,
\end{equation}
where $i$ denotes the corruption type and $l$ is the severity level; $N$ denotes the number of corruption types in our benchmark.
To compare the \textit{performance discrepancy} between \textit{nuScenes-C} and the standard nuScenes dataset, we define a simple mean resilience rate (mRR) metric, which is calculated across three severity levels as follows:
\begin{equation}
    \text{RR}_i = \frac{\sum^{3}_{l=1}\text{NDS}_{i,l}}{3\times \text{NDS}_{\text{clean}}}~,~~
  \text{mRR} = \frac{1}{N}\sum^N_{i=1}\text{RR}_i~.
\end{equation}

In our benchmark, we report both metrics for each candidate BEV algorithm and draw key analyses upon them.

\input{tables/robodet_ce}
\begin{figure*}[htbp]
    \centering
    \begin{minipage}[t!]{0.55\textwidth}
        \centering
        \subfigure[mCE \vs NDS]{
            \label{fig:nds-mce}
            \includegraphics[width=0.46\linewidth]{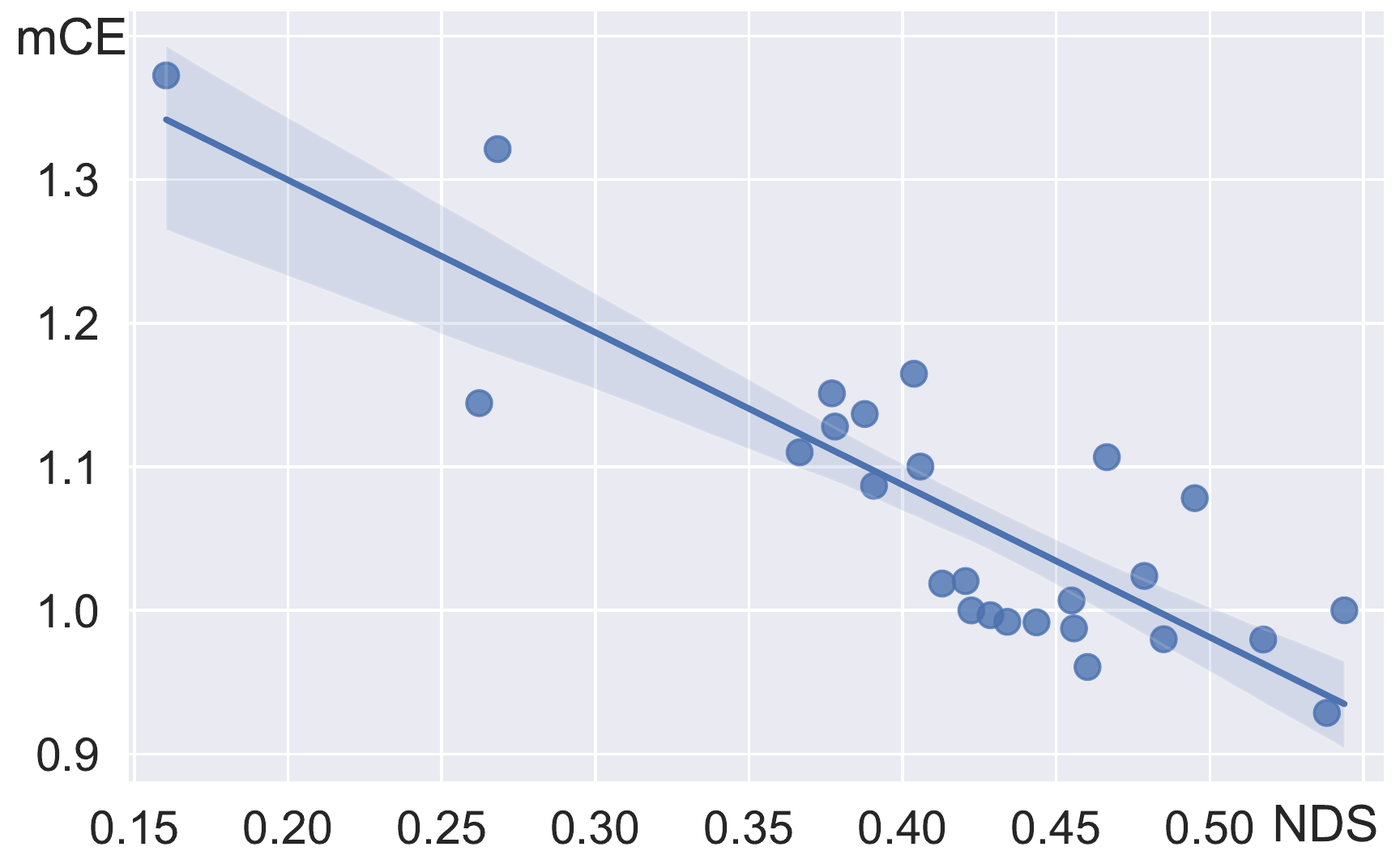}
        }
        \subfigure[mRR \vs NDS]{
            \label{fig:nds-mrr}
            \includegraphics[width=0.46\linewidth]{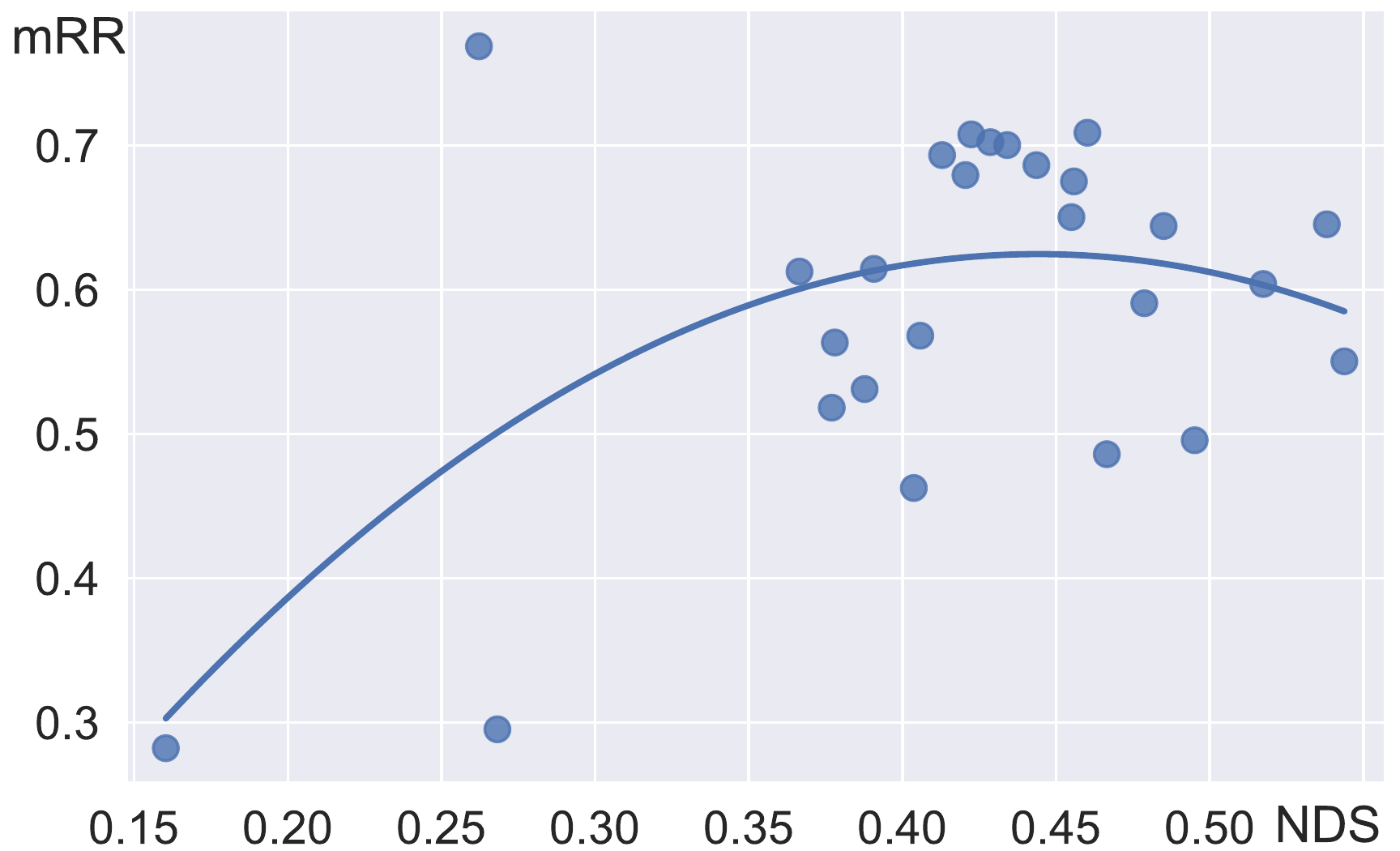}
        }
        \vspace{0.cm}
        \caption{The performance on \textit{nuScenes-C} is improved as the performance on the ``clean" nuScenes~\cite{caesar2020nuscenes} dataset. The relation of absolute performance is close to linear. However, when considering the relative performance, the mRR metric is more randomly distributed without a clear trend to increase.}
        \label{fig:overall}
    \end{minipage}
    \hspace{0.2cm}
    \begin{minipage}[t!]{0.4\textwidth}
        \centering
        \includegraphics[width=0.75\linewidth]{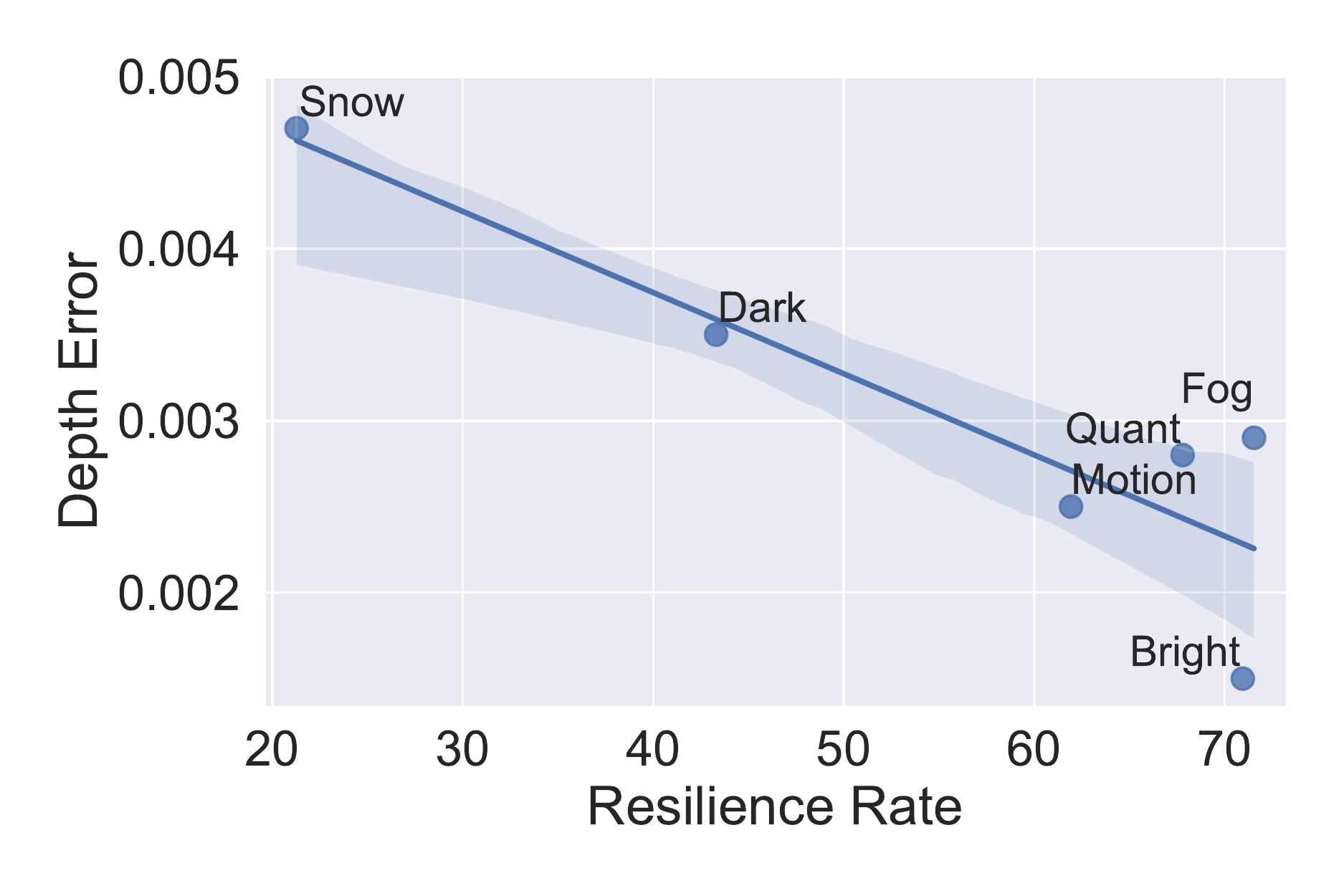}
        \vspace{0.2cm}
        \caption{Depth estimation error \vs Resilience Rate. We observe strong correlations where large depth estimation errors under \textit{Snow} and \textit{Dark} tend to cause drastic performance drops.}
        \label{fig:bevdepth-depth-error}
    \end{minipage}
\end{figure*}

%% file: tables/robodet_model.tex
\begin{table*}[t]
    \centering
    \caption{BEV model calibration. \textbf{Pretrain:} model initialized from pretrained FCOS3D~\cite{wang2021fcos3d} checkpoint; \textbf{Temporal:} model utilizes temporal information; \textbf{Depth:} model with explicit depth estimation branch used in the pipeline; \textbf{CBGS:} model uses the class-balanced group-sampling training strategy~\cite{zhu2019class}. \textbf{Bold}: Best in the category. \underline{Underline}: Second best in the category.}
    \label{tab:robodet_model}
    \vspace{0.2cm}
    \scalebox{0.76}{
    \begin{tabular}{r|cccc|c|c|c|ccc}
    \toprule
    \textbf{Model} & \textbf{Pretrain} & \textbf{Temporal} & \textbf{Depth} & \textbf{CBGS} &\textbf{Backbone} & \textbf{BEV Encoder} & \textbf{Image Size} & \textbf{NDS} $\uparrow$ & \textbf{mCE} (\%) $\downarrow$ & \textbf{mRR} (\%) $\uparrow$\\
    \midrule\midrule
    \rowcolor{gray!10}DETR3D~\cite{wang2022detr3d} & \checkmark &  &  &  & ResNet & Attention & $1600\times 900$ & $0.4224$ & $100.00$ & $70.77$ \\
    DETR3D$_{\text{CBGS}}$~\cite{wang2022detr3d}  & \checkmark &  &  & \checkmark & ResNet & Attention & $1600\times 900$ & $0.4341$ & $99.21$ & $70.02$   \\
    \rowcolor{gray!10}BEVFormer~{\small (small)}~\cite{li2022bevformer}  & \checkmark & \checkmark &  &  & ResNet & Attention & $1280\times 720$ & \underline{$0.4787$}  & $101.23$ & $59.07$  \\
    BEVFormer-S~{\small (small)}~\cite{li2022bevformer}   & \checkmark &  &  &  & ResNet & Attention & $1280\times 720$ & $0.2622$ & $114.43$ & $\mathbf{76.87}$  \\
    \rowcolor{gray!10}BEVFormer~{\small (base)}~\cite{li2022bevformer}  & \checkmark & \checkmark &  &  & ResNet & Attention & $1600\times 900$ & $\mathbf{0.5174}$ & \underline{$97.97$} & $60.40$  \\
    BEVFormer-S~{\small (base)}~\cite{li2022bevformer} & \checkmark &  &  &  & ResNet & Attention & $1600\times 900$ & $0.4129$ & $101.87$ & $69.33$   \\
    \rowcolor{gray!10}PETR~{\small (r50)}~\cite{liu2022petr}  &  &  &  &  & ResNet & Attention & $1408\times 512$ & $0.3665$ & $111.01$ & $61.26$  \\
    PETR~{\small (vov)}~\cite{liu2022petr}  & \checkmark &  &  &  & VoVNet-V2 & Attention & $1600\times 640$ & $0.4550$ & $100.69$ & $65.03$   \\
    \rowcolor{gray!10}ORA3D~\cite{roh2022ora3d}  & \checkmark &  &  &  & ResNet & Attention & $1600\times 900$ & $0.4436$ & $99.17$ & $68.63$   \\
    PolarFormer~{\small (r101)}~\cite{jiang2022polarformer}  & \checkmark &  &  &  & ResNet & Attention & $1600\times 900$ & $0.4602$ & $\mathbf{96.06}$ & \underline{$70.88$}   \\
    \rowcolor{gray!10}PolarFormer~{\small (vov)}~\cite{jiang2022polarformer}    & \checkmark &  &  &  & VoVNet-V2 & Attention & $1600\times 900$ & $0.4558$ & $98.75$ & $67.51$  \\
    \midrule
    SRCN3D~{\small (r101)}~\cite{shi2022srcn3d}  & \checkmark &  &  &  & ResNet & CNN + Attn. & $1600\times 900$ & $0.4286$ & $\mathbf{99.67}$ & $\mathbf{70.23}$   \\
    \rowcolor{gray!10}SRCN3D~{\small (vov)}~\cite{shi2022srcn3d}  & \checkmark &  &  &  & VoVNet-V2 & CNN + Attn. & $1600\times 900$ & $0.4205$ & $102.04$ & $67.95$
    \\
    Sparse4D~{\small (r101)}~\cite{lin2022sparse4d}  & \checkmark & \checkmark &  &  & ResNet & CNN + Attn. & $1600\times 640$ & $\mathbf{0.5438}$  & $100.01$ & $55.04$   \\
    \midrule
    \rowcolor{gray!10}BEVDet~{\small (r50)}~\cite{huang2021bevdet}  &  &  & \checkmark & \checkmark & ResNet & CNN & $704\times 256$ & $0.3770$ & $115.12$ & $51.83$   \\
    BEVDet~{\small (r101)}~\cite{huang2021bevdet}  &  &  & \checkmark & \checkmark & ResNet & CNN & $704\times 256$ & $0.3877$ & $113.68$ & $53.12$    \\
    \rowcolor{gray!10}BEVDet~{\small (r101)}~\cite{huang2021bevdet}  & \checkmark &  & \checkmark & \checkmark & ResNet & CNN & $704\times 256$ & $0.3780$ & $112.80$ & $56.35$    \\
    BEVDet~{\small (tiny)}~\cite{huang2021bevdet}  &  &  & \checkmark & \checkmark & SwinTrans & CNN & $704\times 256$ & $0.4037$ & $116.48$ & $46.26$   \\
    \rowcolor{gray!10}BEVDepth~{\small (r50)}~\cite{li2022bevdepth}  &  &  & \checkmark & \checkmark & ResNet & CNN & $704\times 256$ & $0.4058$ & $110.02$ & $56.82$  \\
    BEVerse~{\small (swin-t)}~\cite{zhang2022beverse}  &  & \checkmark & \checkmark & \checkmark & SwinTrans & CNN & $704\times 256$ & $0.4665$ & $110.67$ & $48.60$  \\
    \rowcolor{gray!10}BEVerse-S~{\small (swin-t)}~\cite{zhang2022beverse} &  &  & \checkmark & \checkmark & SwinTrans & CNN & $704\times 256$ & $0.1603$ & $137.25$ & $28.24$ \\
    BEVerse~{\small (swin-s)}~\cite{zhang2022beverse}  &  & \checkmark & \checkmark & \checkmark & SwinTrans & CNN & $1408\times 512$ & \underline{$0.4951$} & $117.82$ & $49.57$   \\
    \rowcolor{gray!10}BEVerse-S~{\small (swin-s)}~\cite{zhang2022beverse} &  &  & \checkmark & \checkmark & SwinTrans & CNN & $1408\times 512$ & $0.2682$ & $132.13$ & $29.54$ \\
    SOLOFusion~{\small (short)}~\cite{Park2022TimeWT} &  & \checkmark & \checkmark &  & ResNet & CNN & $704\times 256$ & $0.3907$ & $108.68$ & $61.45$ \\
    \rowcolor{gray!10}SOLOFusion~{\small (long)}~\cite{Park2022TimeWT} &  & \checkmark & \checkmark &  & ResNet & CNN & $704\times 256$ & $0.4850$ & \underline{$97.99$} & \underline{$64.42$} \\
    SOLOFusion~{\small (fusion)}~\cite{Park2022TimeWT} &  & \checkmark & \checkmark & \checkmark & ResNet & CNN & $704\times 256$ & $\mathbf{0.5381}$ & $\mathbf{92.86}$ & $\mathbf{64.53}$ \\
    \bottomrule
    \end{tabular}
    }
\end{table*}

%% file: tables/robodet_ce.tex
\begin{table*}[t]
    \centering
    \caption{The \textbf{Corruption Error (CE)} of each BEV detector in our \textit{RoboBEV} benchmark. \textbf{Bold}: Best in the catogory. \colorbox{blue!9.5}{Blue}: Best in the row if improve upon baseline. \colorbox{yellow!20}{Yellow}: Worst in the row if decline upon baseline. \dag: distinguish pre-training version BEVDet.}
    \vspace{0.2cm}
    \label{tab:robodet_ce}
    \scalebox{0.93}{
    \footnotesize
    \begin{tabular}{r|c|c|ccccccccc}
    \toprule
    \textbf{Model} & \textbf{NDS} $\uparrow$ & \textbf{mCE} (\%) $\downarrow$ & \textbf{Camera} & \textbf{Frame} & \textbf{Quant} & \textbf{Motion} & \textbf{Bright} & \textbf{Dark} & \textbf{Fog} & \textbf{Snow} \\
    \midrule\midrule
    \rowcolor{gray!10}DETR3D~\cite{wang2022detr3d} & $0.4224$ & $100.00$ & $100.00$ & $100.00$ & $100.00$ & $100.00$ & $100.00$ & $100.00$ & $100.00$ & $100.00$ \\\midrule
    DETR3D$_{\text{CBGS}}$~\cite{wang2022detr3d} & $0.4341$  & $99.21$  & $98.15$ & $98.90$ & $99.15$ & \cellcolor{yellow!20}$101.62$ & \cellcolor{blue!9.5}$97.47$ & $\mathbf{100.28}$ & $98.23$ & $99.85$ \\
    BEVFormer~{\scriptsize (small)}~\cite{li2022bevformer} & $0.4787$  & $102.40$  & $101.23$ & $101.96$ & \cellcolor{blue!9.5}$98.56$ & $101.24$ & $104.35$ & $105.17$ & \cellcolor{yellow!20}$105.40$ & $101.29$  \\
    BEVFormer~{\scriptsize (base)}~\cite{li2022bevformer} &$\mathbf{0.5174}$  &  $97.97$ & $\mathbf{95.87}$  & \cellcolor{blue!9.5}$\mathbf{94.42}$ & $\mathbf{95.13}$ & $99.54$ & $96.97$ & \cellcolor{yellow!20}$103.76$ & $97.42$ & $100.69$ \\
    PETR~{\scriptsize (r50)}~\cite{liu2022petr} & $0.3665$  & $111.01$  & $107.55$ & $105.92$ & $110.33$ & $104.93$ & \cellcolor{yellow!20}$119.36$ & $116.84$ & $117.02$ & $106.13$ \\
    PETR~{\scriptsize (vov)}~\cite{liu2022petr} & $0.4550$  & $100.69$  & $99.09$ & $97.46$ & $103.06$ & $102.33$ & $102.40$ & \cellcolor{yellow!20}$106.67$ & $103.43$ & \cellcolor{blue!9.5}$\mathbf{91.11}$ \\
    ORA3D~\cite{roh2022ora3d} & $0.4436$ & $99.17$ & \cellcolor{blue!9.5}$97.26$ & $98.03$ & $97.32$ & $100.19$ & $98.78$ & \cellcolor{yellow!20}$102.40$ & $99.23$ & $100.19$ \\
    PolarFormer~{\scriptsize (r101)}~\cite{jiang2022polarformer} & $0.4602$ & $\mathbf{96.06}
$  & $96.16$ & $97.24$ & $\mathbf{95.13}$ & \cellcolor{blue!9.5}$\mathbf{92.37}$ & $\mathbf{94.96}$ & \cellcolor{yellow!20}$103.22$ & $\mathbf{94.25}$ & $95.17$ \\
    PolarFormer~{\scriptsize (vov)}~\cite{jiang2022polarformer} & $0.4558$ & $98.75$  & $96.13$ & $97.20$ & $101.48$ & $104.32$ & $95.37$ & \cellcolor{yellow!20}$104.78$ & $97.55$ & \cellcolor{blue!9.5}$93.14$  \\
    \midrule
    SRCN3D~{\scriptsize (r101)}~\cite{shi2022srcn3d} & $0.4286$ & $\mathbf{99.67}$  & $\mathbf{98.77}$ & $\mathbf{98.96}$ & \cellcolor{blue!9.5}$\mathbf{97.93}$ & $\mathbf{100.71}$ & $\mathbf{98.80}$ & \cellcolor{yellow!20}$\mathbf{102.72}$ & $\mathbf{99.54}$ & $99.91$ \\
    SRCN3D~{\scriptsize (vov)}~\cite{shi2022srcn3d}  & $0.4205$ & $102.04$  & $99.78$ & $100.34$ & $105.13$ & $107.06$ & $101.93$ & \cellcolor{yellow!20}$107.10$ & $102.27$ & \cellcolor{blue!9.5}$\mathbf{92.75}$ \\
    Sparse4D~{\scriptsize (r101)}~\cite{lin2022sparse4d} & $\mathbf{0.5438}$ & $100.01$  & $99.80$ & $99.91$ & $98.05$ & $102.00$ & $100.30$ & \cellcolor{yellow!20}$103.83$ & $100.46$ & \cellcolor{blue!9.5}$95.72$ \\
    \midrule
    BEVDet~{\scriptsize (r50)}~\cite{huang2021bevdet} & $0.3770$  & $115.12$  & $105.22$ & $109.19$ & $111.27$ & $108.18$ & \cellcolor{yellow!20}$123.96$ & $123.34$ & $123.83$ & $115.93$ \\
    BEVDet~{\scriptsize (tiny)}~\cite{huang2021bevdet} & $0.4037$  & $116.48$  & $103.50$ & $106.61$ & $113.18$ & $107.26$ & $130.19$ & \cellcolor{yellow!20}$131.83$ & $124.01$ & $115.25$ \\
    BEVDet~{\scriptsize (r101)}~\cite{huang2021bevdet} & $0.3877$  & $113.68$  & $103.32$ & $107.29$ & $109.25$ & $105.40$ & \cellcolor{yellow!20}$124.14$ & $123.12$ & $123.28$ & $113.64$ \\
    BEVDet~{\scriptsize (r101\dag)}~\cite{huang2021bevdet} & $0.3780$ & $112.80$  & $105.84$ & $108.68$ & $101.99$ & $100.97$ & $123.39$ & $119.31$ & \cellcolor{yellow!20}$130.21$ & $112.04$ \\
    BEVDepth~{\scriptsize (r50)}~\cite{li2022bevdepth} & $0.4058$  & $110.02$  & $103.09$ & $106.26$ & $106.24$ & $102.02$ & \cellcolor{yellow!20}$118.72$ & $114.26$ & $116.57$ & $112.98$ \\
    BEVerse~{\scriptsize (swin-t)}\cite{zhang2022beverse} & $0.4665$ & $110.67$  & $95.49$ & \cellcolor{blue!9.5}$94.15$ & $108.46$ & $100.19$ & $122.44$ & \cellcolor{yellow!20}$130.40$ & $118.58$ & $115.69$ \\
    BEVerse~{\scriptsize (swin-s)}~\cite{zhang2022beverse} & $0.4951$ & $107.82$  & \cellcolor{blue!9.5}$92.93$ & $101.61$ & $105.42$ & $100.40$ & $110.14$ & \cellcolor{yellow!20}$123.12$ & $117.46$ & $111.48$ \\
    SOLOFusion~{\scriptsize (short)}~\cite{Park2022TimeWT} & $0.3907$ & $108.68$  & $104.45$ & $105.53$ & $105.47$ & $100.79$ & $117.27$ & $110.44$ & \cellcolor{yellow!20}$115.01$ & $110.47$ \\
    SOLOFusion~{\scriptsize (long)}~\cite{Park2022TimeWT} & $0.4850$ & $97.99$  & $95.80$ & $101.54$ & $93.83$ & \cellcolor{blue!9.5}$89.11$ & $100.00$ & $\mathbf{99.61}$ & $98.70$ & \cellcolor{yellow!20}$\mathbf{105.35}$ \\
    SOLOFusion~{\scriptsize (fusion)}~\cite{Park2022TimeWT} & $\mathbf{0.5381}$ & $\mathbf{92.86}$  & $\mathbf{86.74}$ & $\mathbf{88.37}$ & $\mathbf{87.09}$ & \cellcolor{blue!9.5}$\mathbf{86.63}$ & $\mathbf{94.55}$ & $102.22$ & $\mathbf{90.67}$ & \cellcolor{yellow!20}$106.64$ \\
    \bottomrule
    \end{tabular}
    }
\end{table*}

%% file: sections/04_experiment.tex
\begin{figure*}[ht]
    \centering
    \includegraphics[width=\linewidth]{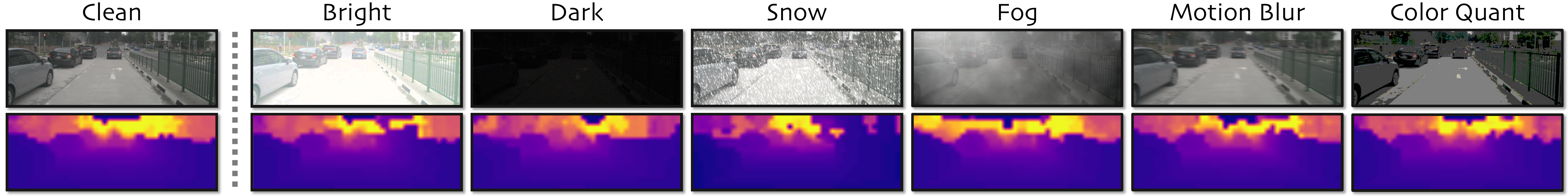}
    \caption{Depth estimation results of BEVDepth~\cite{li2022bevdepth} under different corruptions. The results exhibit a different sensitivity for each type.}
    \label{fig:depth-estimation}
\end{figure*}
\input{tables/robodet_rr_2}
\input{tables/multimodal}

\section{Experiments}
\label{sec:exp}
\subsection{Experimental Setup}
In our study, we use the official model configurations and public weight checkpoints provided by open-sourced codebases, whenever applicable; while we also train additional model variants with minimal modifications to conduct experiments under controlled settings. To ensure high consistency and reproducibility, we set a fixed random seed for all experiments. Furthermore, we report metrics for each corruption type by averaging over three severities. We adopt DETR3D~\cite{wang2022detr3d} as our baseline for the mCE metric, considering its wide adoption as a state-of-the-art approach. Our code is built upon
MMDetection3D codebase~\cite{mmdet3d2020}.

\subsection{Main Benchmarking Results}

In this study, we conduct a comprehensive benchmarking analysis of 26 existing BEV detectors on the \textit{nuScenes-C} dataset. The main results of our experiments are presented in Tables~\ref{tab:robodet_ce} and~\ref{tab:robodet_rr}.
Our findings indicate that all models exhibit varying degrees of performance declines on the corruption set. We observed that \textit{Bright}, which causes a much larger shift in pixel distribution than \textit{Motion Blur}, resulted in the smallest performance drop. For most of the models, the resilience rate of \textit{Bright} remains the highest.

{\textit{We notice a strong correlation of the absolute performances between nuScenes-C and the ``clean" dataset.}} Specifically, BEV detectors that perform well on the standard dataset are also likely to perform better on the out-of-distribution dataset, as illustrated in Figure~\ref{fig:nds-mce}. However, a closer examination of the results revealed a more complex situation. Models with similar performance on the ``clean" dataset exhibit diverse robustness under different types of corruption. For example, BEVerse (swin-s) \cite{zhang2022beverse} demonstrates significantly improved robustness in \textit{Camera Crash} compared to the baseline, while PETR (vov) \cite{liu2022petr} performs well in \textit{Snow} weather. However, both models struggle to make accurate predictions under \textit{Dark} conditions. 

{\textit{We further find that resilience rates under corruptions confront the risk of decreasing. Superior performances on the standard dataset do not necessarily lead to better resilience rates.}} While the mCE metric reveals a linear relationship between absolute performance on the nuScenes dataset and the \textit{nuScenes-C} dataset, the mRR metric shows large variations among models exhibiting competitive standard performance, suggesting that some models may overfit the nuScenes dataset and struggle to generalize well to the \textit{nuScenes-C} dataset.

This is exemplified by Sparse4D~\cite{lin2022sparse4d} outperforming DETR3D~\cite{wang2022detr3d} on the ``clean" dataset while having lower mRR metrics for all corruption types. Furthermore, we observe that DETR3D exhibits the most robust performance under \textit{Dark} conditions, while BEVerse (swin-t) with better ``clean" performance only achieves a relative performance of approximately 12\% under dark lighting conditions. Therefore, we suggest that benchmarking state-of-the-art models from multiple perspectives is critical to comprehensively evaluate their performance.

To further investigate the behavior of the model's robustness, we break down BEV detectors into various components, namely, training strategies (\eg, FCOS3D~\cite{wang2021fcos3d} pretraining and CBGS~\cite{zhu2019class} resampling strategy for addressing unbalanced classes), model architectures (\eg, backbone), and approach pipelines (\eg, temporal cue learning and depth estimation). The results of this analysis are in Table~\ref{tab:robodet_model}. 

Specifically, we focus on FCOS3D~\cite{wang2021fcos3d} pre-training and CBGS~\cite{zhu2019class} resampling strategy, as they are commonly utilized in these approaches and have demonstrated effectiveness in improving the performance of the ``clean" dataset.

\subsection{Camera-LiDAR Fusion}
Apart from building BEV representations only from image inputs, BEVFusion~\cite{liu2022bevfusion} extracts a shared BEV space to fuse features from both image and point clouds. We consider the settings where the cameras are affected while the LiDAR works well, which are common happened in the real world. For instance, the capture process of point clouds is not disturbed by the lighting conditions but the images are of low quality in the dark. In terms of weather like \textit{Snow} and \textit{Fog}, which can add noise to both camera and LiDAR, we do not consider this setting in this work. The results are shown in Table~\ref{tab:multimodal}. It's interesting to notice that the multi-modality fusion model still keeps high performance even when the imaging modality is corrupted. The performance of the fusion model under most natural corruptions outperforms that of the LiDAR-based model. This reveals that multi-modality inputs can be complementary when one of them degenerates.

%% file: tables/robodet_rr_2.tex
\begin{table*}[t]
    \centering
    \caption{The \textbf{Resilience Rate (RR)} of each BEV detector in our \emph{RoboBEV} benchmark. \textbf{Bold}: best within the category. \colorbox{blue!9.5}{Blue}: Best across category. \dag: distinguish pre-training version BEVDet.}
    \label{tab:robodet_rr}
    \vspace{0.2cm}
    \scalebox{0.96}{
    \footnotesize
    \begin{tabular}{r|c|c|ccccccccc}
    \toprule
    \textbf{Model} & \textbf{NDS} & \textbf{mRR} (\%) $\uparrow$ & \textbf{Camera} & \textbf{Frame} & \textbf{Quant} & \textbf{Motion} & \textbf{Bright} & \textbf{Dark} & \textbf{Fog} & \textbf{Snow} \\
    \midrule\midrule
    DETR3D~\cite{wang2022detr3d} & $0.4224$  & $70.77$  & $67.68$ & $61.65$ & $75.21$ & $63.00$ & $94.74$ & \cellcolor{blue!9.5}$\mathbf{65.96}$ & \cellcolor{blue!9.5}$\mathbf{92.61}$ & $45.29$ \\
    DETR3D$_{\text{CBGS}}$~\cite{wang2022detr3d} & $0.4341$ & $70.02$  & $\mathbf{68.90}$ & $61.85$ & $74.52$ & $58.56$ & \cellcolor{blue!9.5}$\mathbf{95.69}$ & $63.72$ & \cellcolor{blue!9.5}$\mathbf{92.61}$ & $44.34$ \\
    BEVFormer~{\scriptsize (small)}~\cite{li2022bevformer} & $0.4787$ & $59.07$  & $57.89$ & $51.37$ & $68.41$ & $53.69$ & $78.15$ & $50.41$ & $74.85$ & $37.79$  \\
    BEVFormer~{\scriptsize (base)}~\cite{li2022bevformer} &$\mathbf{0.5174}$ &  $60.40$ & $60.96$  & $58.31$ & $67.82$ & $52.09$ & $80.87$ & $48.61$ & $78.64$ & $35.89$ \\
    PETR~{\scriptsize (r50)}~\cite{liu2022petr} & $0.3665$ & $61.26$  & $63.30$ & $59.10$ & $67.45$ & $62.73$ & $77.52$ & $42.86$ & $78.47$ & $38.66$ \\
    PETR~{\scriptsize (vov)}~\cite{liu2022petr} & $0.4550$ & $65.03$  & $64.26$ & $61.36$ & $65.23$ & $54.73$ & $84.79$ & $50.66$ & $81.38$ & $\mathbf{57.85}$ \\
    ORA3D~\cite{roh2022ora3d} & $0.4436$ & $68.63$  & $68.87$ & $\mathbf{61.99}$ & $75.74$ & $59.67$ & $91.86$ & $58.90$ & $89.25$ & $42.79$ \\
    PolarFormer~{\scriptsize (r101)}~\cite{jiang2022polarformer} & $0.4602$ & \cellcolor{blue!9.5}$\mathbf{70.88}$  & $68.08$ & $61.02$ & $\mathbf{76.25}$ & $\mathbf{69.99}$ & $93.52$ & $55.50$ & \cellcolor{blue!9.5}$\mathbf{92.61}$ & $50.07$ \\
    PolarFormer~{\scriptsize (vov)}~\cite{jiang2022polarformer} & $0.4558$ & $67.51$  & $68.78$ & $61.67$ & $67.49$ & $51.43$ & $93.90$ & $53.55$ & $89.10$ & $54.15$  \\
    \midrule
    SRCN3D~{\scriptsize (r101)}~\cite{shi2022srcn3d} & $0.4286$ & $\mathbf{70.23}$  & $\mathbf{68.76}$ & $\mathbf{62.55}$ & $\mathbf{77.41}$ & $\mathbf{60.87}$ & $\mathbf{95.05}$ & $\mathbf{60.43}$ & $\mathbf{91.93}$ & $44.80$ \\
    SRCN3D~{\scriptsize (vov)}~\cite{shi2022srcn3d} & $0.4205$ & $67.95$  & $68.37$ & $61.33$ & $67.23$ & $50.96$ & $92.41$ & $54.08$ & $89.75$ & \cellcolor{blue!9.5}$\mathbf{59.43}$ \\
    Sparse4D{\scriptsize (r101)}~\cite{lin2022sparse4d} & $\mathbf{0.5438}$ & $55.04$  & $52.83$ & $48.01$ & $60.87$ & $46.23$ & $73.26$ & $46.16$ & $71.42$ & $41.54$ \\
    \midrule
    BEVDet~{\scriptsize (r50)}~\cite{huang2021bevdet} & $0.3770$ & $51.83$  & $65.94$ & $51.03$ & $63.87$ & $54.67$ & $68.04$ & $29.23$ & $65.28$ & $16.58$ \\
    BEVDet~{\scriptsize (tiny)}~\cite{huang2021bevdet} & $0.4037$ & $46.26$  & $64.63$ & $52.39$ & $56.43$ & $52.71$ & $54.27$ & $12.14$ & $60.69$ & $16.84$ \\
    BEVDet~{\scriptsize (r101)}~\cite{huang2021bevdet} & $0.3877$ & $53.12$  & $67.63$ & $53.26$ & $65.67$ & $58.42$ & $65.88$ & $28.84$ & $64.35$ & $20.89$ \\
    BEVDet~{\scriptsize (r101\dag)}~\cite{huang2021bevdet}  & $0.3780$ & $56.35$  & $64.60$ & $51.90$ & \cellcolor{blue!9.5}$\mathbf{80.45}$ & $68.52$ & $68.76$ & $36.85$ & $54.84$ & $24.84$ \\
    BEVDepth~{\scriptsize (r50)}~\cite{li2022bevdepth}  & $0.4058$ & $56.82$  & $65.01$ & $52.76$ & $67.79$ & $61.93$ & $70.95$ & $43.30$ & $71.54$ & $21.27$ \\
    BEVerse~{\scriptsize (swin-t)}~\cite{zhang2022beverse} & $0.4665$ & $48.60$  & $68.19$ & \cellcolor{blue!9.5}$\mathbf{65.10}$ & $55.73$ & $56.74$ & $56.93$ & $12.71$ & $59.61$ & $13.80$ \\
    BEVerse~{\scriptsize (swin-s)}~\cite{zhang2022beverse} & $0.4951$ & $49.57$  & $67.95$ & $50.19$ & $56.70$ & $53.16$ & $68.55$ & $22.58$ & $57.54$ & $19.89$ \\
    SOLOFusion~{\scriptsize (short)}~\cite{Park2022TimeWT} & $0.3907$ & $61.45$  & $65.04$ & $56.18$ & $71.77$ & $66.62$ & $75.92$ & $52.03$ & $76.73$ & $27.28$ \\
    SOLOFusion~{\scriptsize (long)}~\cite{Park2022TimeWT} & $0.4850$ & $64.42$  & $65.13$ & $51.34$ & $74.19$ & \cellcolor{blue!9.5}$\mathbf{71.34}$ & $\mathbf{82.52}$ & $\mathbf{58.02}$ & $82.29$ & $\mathbf{30.52}$ \\
    SOLOFusion~{\scriptsize (fusion)}~\cite{Park2022TimeWT} & $\mathbf{0.5381}$ & $\mathbf{64.53}$  & \cellcolor{blue!9.5}$\mathbf{70.73}$ & $64.37$ & $75.41$ & $67.68$ & $80.45$ & $48.80$ & $\mathbf{83.26}$ & $25.57$ \\
    \bottomrule
    \end{tabular}
    }
\end{table*}

%% file: tables/multimodal.tex
\begin{table*}[t]
    \centering
    \caption{NDS results of BEVFusion~\cite{liu2022bevfusion} under different input modalities. Since \textit{Fog} and \textit{Snow} can also affect the LiDAR sensors, we do not consider these two corruptions of the fusion model.}
    \vspace{0.2cm}
    \label{tab:multimodal}
    \scalebox{0.93}{
    \footnotesize
    \begin{tabular}{cc|c|cccccccc}
    \toprule
    \textbf{Camera} & \textbf{LiDAR} & \textbf{Clean} & \textbf{Camera} & \textbf{Frame} & \textbf{Quant} & \textbf{Motion} & \textbf{Bright} & \textbf{Dark} & \textbf{Fog} & \textbf{Snow} \\
    \midrule
    \checkmark &  & $0.4121$ & $0.2777$ & $0.2255$ & $0.2763$ & $0.2788$ & $0.2902$ & $0.1076$ & $0.3041$ & $0.1461$ \\
     & \checkmark & $0.6928$ & $-$ & $-$ & $-$ & $-$ & $-$ & $-$ & $-$ & $-$ \\
    \checkmark & \checkmark & $0.7138$ & $0.6963$ & $0.6931$ & $0.7044$ & $0.6977$ & $0.7018$ & $0.6787$ & $-$ & $-$\\
    \bottomrule
    \end{tabular}
    }
\end{table*}

%% file: sections/05_discussion.tex
\section{Further Analysis}
\label{sec:dis}

\begin{figure*}[ht]
    \centering
    \subfigure[Pretrain - mCE]{
        \label{fig:nds-mce-pretrain}
        \includegraphics[width=0.23\linewidth]{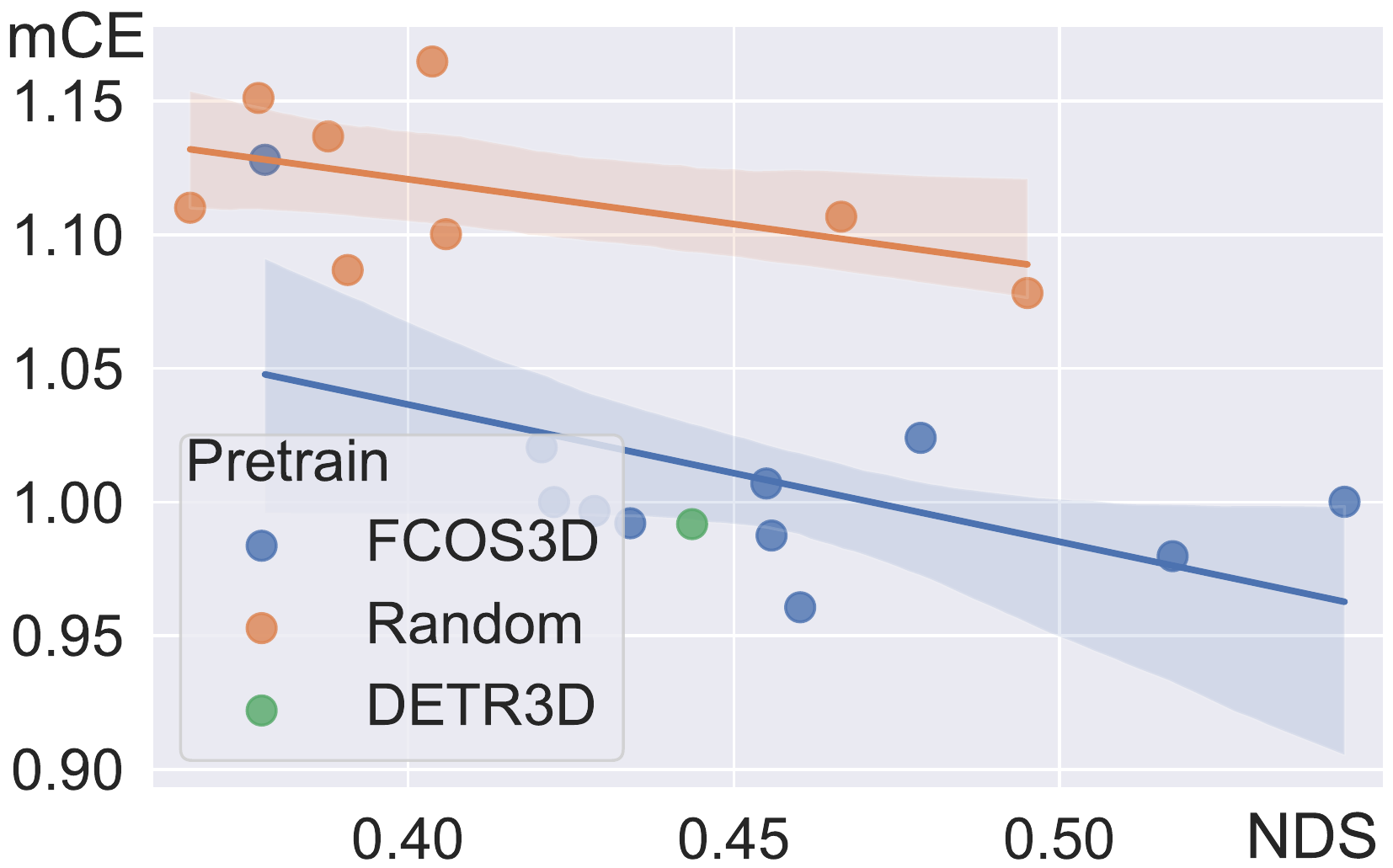}
    }
    \subfigure[Pretrain - mRR]{
        \label{fig:nds-mrr-pretrain}
        \includegraphics[width=0.23\linewidth]{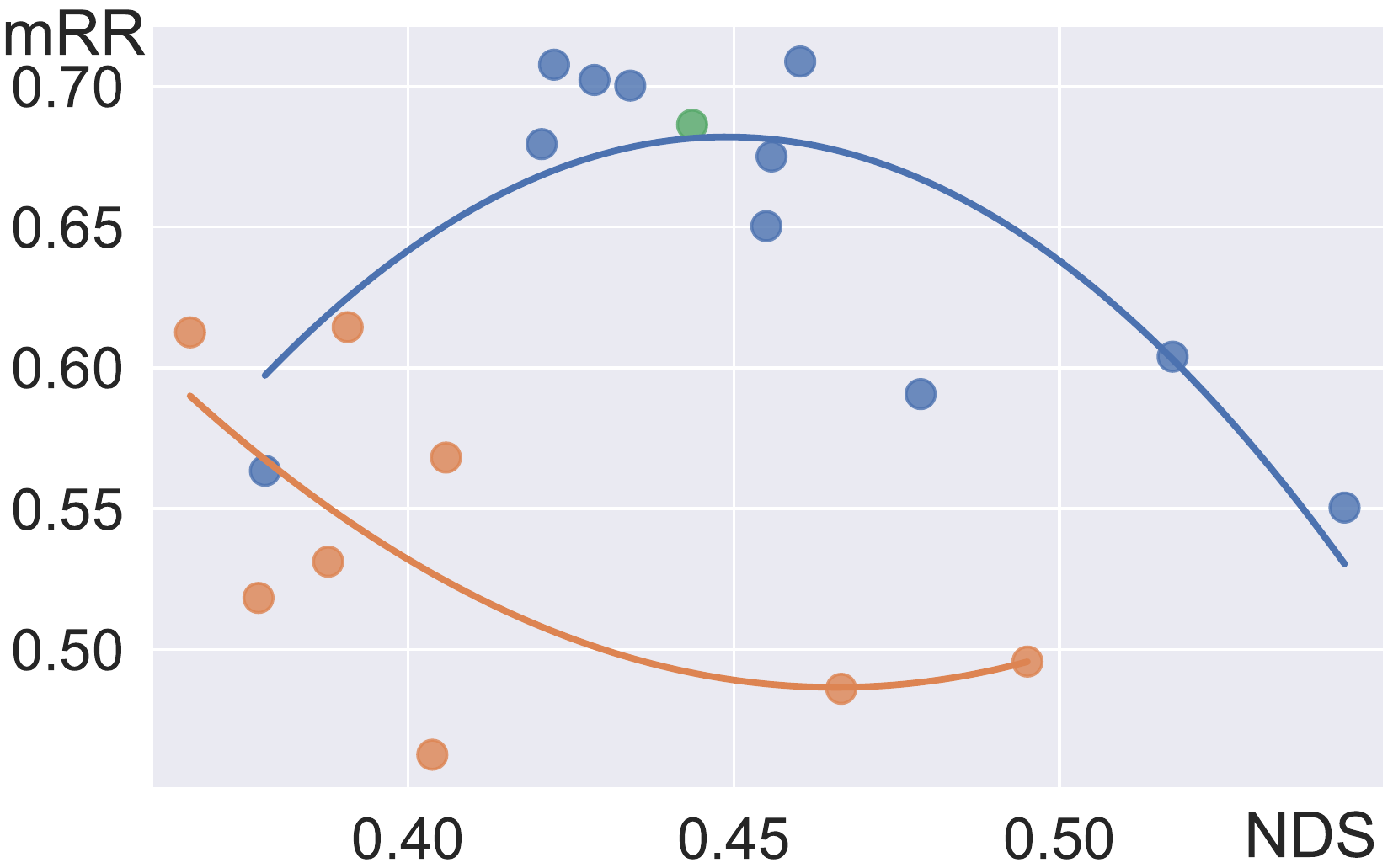}
    }
    \subfigure[Depth - mCE]{
        \label{fig:nds-mce-depth}
        \includegraphics[width=0.23\linewidth]{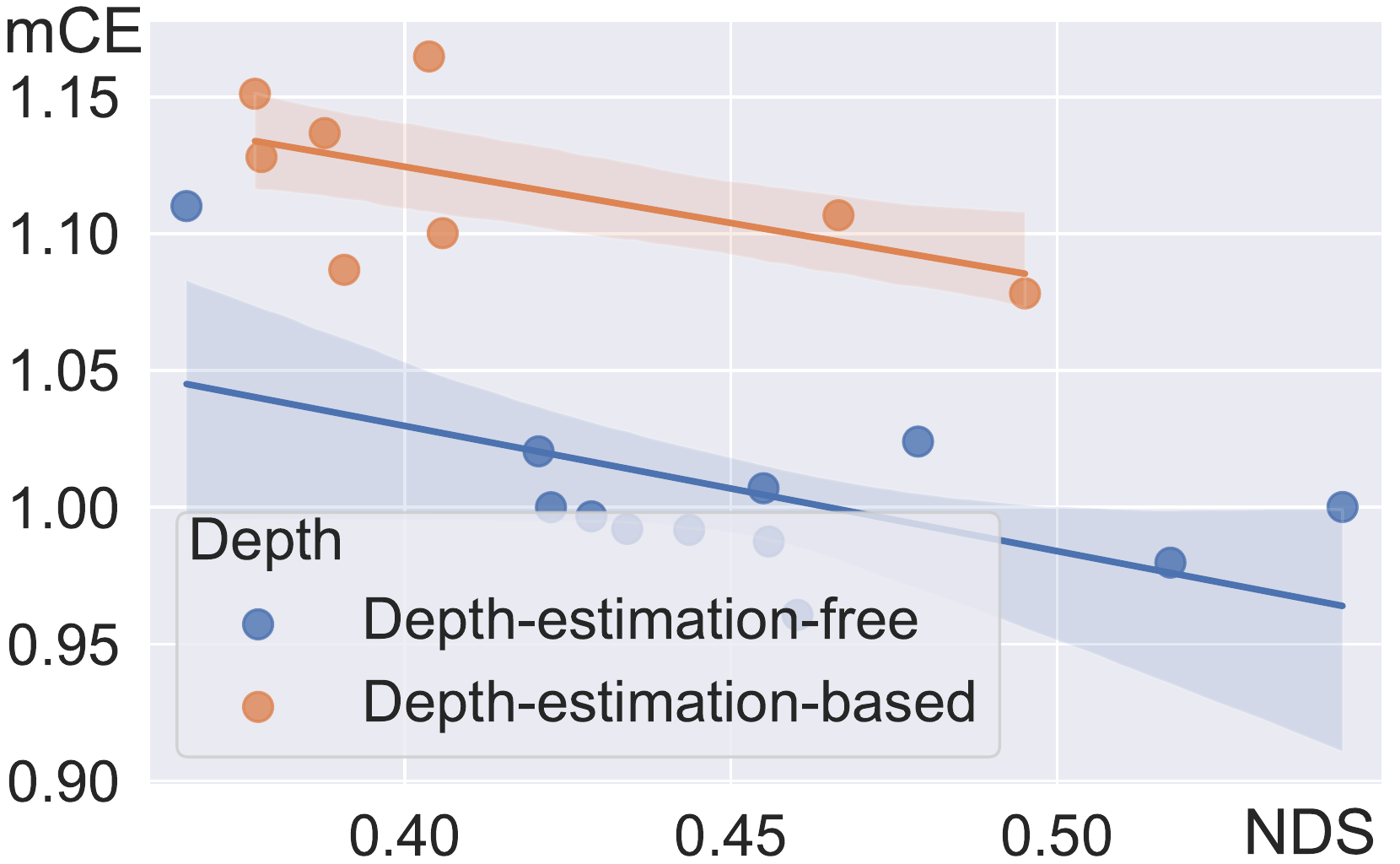}
    }
    \subfigure[Depth - mRR]{
        \label{fig:nds-mrr-depth}
        \includegraphics[width=0.23\linewidth]{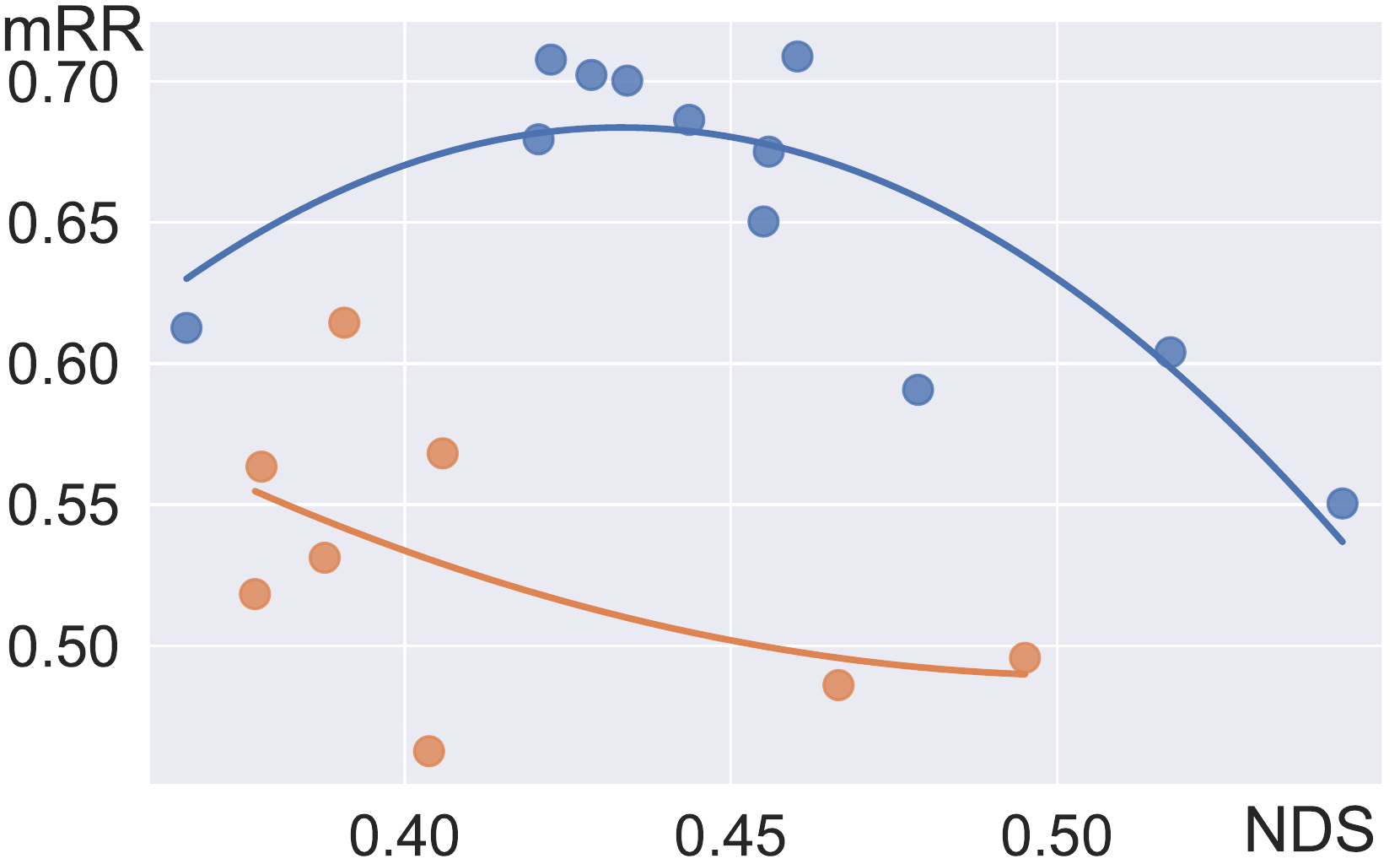}
    }
    \label{fig:pretrain}
    \vspace{-0.1cm}
    \caption{Pre-training strategies together with depth-free bird's eye view transformation provide the models with better robustness. We do not consider SOLOFusion~\cite{Park2022TimeWT} long-term fusion here since it utilizes 16 frames, which is much larger than other methods.}
\end{figure*}

\subsection{Depth Estimation}
\textbf{\textit{Depth-free BEV transformations show better robustness.}}
Prior works in this area can be classified into two categories based on how they exploit depth estimation. The first category, including works such as BEVDet~\cite{huang2021bevdet}, BEVDepth~\cite{li2022bevdepth}, and BEVerse~\cite{zhang2022beverse}, incorporates an explicit depth-estimation branch in the pipeline. This is done because predicting 3D bounding boxes from monocular images is an ill-posed problem. These approaches first predict a per-pixel depth map, which is then used to map image features to corresponding 3D locations. Subsequently, they predict 3D targets in the bird's-eye view (BEV) perspective by aggregating BEV features in a bottom-up fashion~\cite{wang2022detr3d}.

The second category of works uses pre-defined object queries~\cite{li2022bevformer, wang2022detr3d} or sparse proposals~\cite{lin2022sparse4d, shi2022srcn3d} to index 2D features in a top-down manner. These two distinct designs demonstrate competitive performance on ``clean" data, and we expand the analysis by evaluating their performance on out-of-distribution datasets.
Our analysis reveals that depth-based approaches suffer from severe performance degradation when exposed to corrupted images as shown in Figure~\ref{fig:nds-mce-depth} and~\ref{fig:nds-mrr-depth}. The observed degradation could be due to inaccurate depth estimation, as illustrated in Figure~\ref{fig:depth-estimation}. 

Moreover, we undertake a comparative study to evaluate the intermediate depth estimation results of BEVDepth~\cite{li2022bevdepth} under corruptions. To this end, we compute the mean square error (MSE) between ``clean" inputs and corrupted inputs. Our findings indicate an explicit correlation between vulnerability and depth estimation error, as presented in Figure~\ref{fig:bevdepth-depth-error}. Specifically, \textit{Snow} and \textit{Dark} corruptions significantly affect accurate depth estimation, leading to the largest performance drop. These results provide further support for our conclusion that the performance of depth-based approaches can suffer significantly if the depth estimation is not accurate enough.

\subsection{Pre-Training}
\textbf{\textit{Pre-training improves robustness across a wide range of semantic corruptions while does not help with temporal corruptions.}}
In recent years, pre-training has emerged as a promising technique for enhancing the performance of computer vision models across various tasks. In the context of 3D detection, it is common to use the FCOS3D~\cite{wang2021fcos3d} weights as initialization for the ResNet backbone. FCOS3D employs a depth weight of 0.2 for stable training, which is then switched to 1 for fine-tuning, as described in~\cite{wang2021fcos3d}. Alternatively, the VoVNet-V2~\cite{lee2020centermask} backbone is first trained on the DDAD15M~\cite{guizilini20203d} dataset for depth estimation and then fine-tuned on the nuScenes train set for detection. These two pre-training strategies can be categorized as semantic and depth pre-training, respectively. 

\begin{figure}
    \centering
    \includegraphics[width=\linewidth]{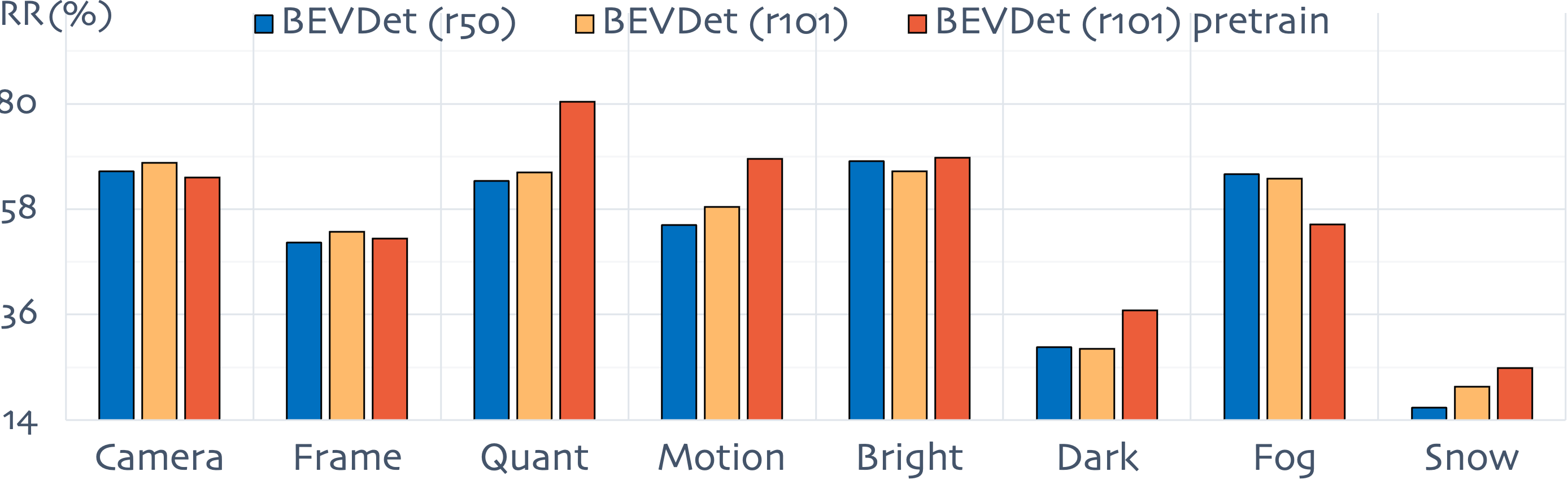}
    \caption{Resilience rate comparisons of BEVDet~\cite{huang2021bevdet} with and without pre-training. The higher the better.}
    \label{fig:bevdet-pretrain}
\end{figure}

The effectiveness of these strategies for improving model robustness is illustrated in Figure~\ref{fig:nds-mce-pretrain} and Figure~\ref{fig:nds-mrr-pretrain}, where models utilize pre-training largely outperform those not. For controlled comparison, we re-implement the BEVDet (r101) model using the FCOS3D checkpoint as initialization. Our results, presented in Figure~\ref{fig:bevdet-pretrain}, show that pre-training can significantly improve mRR across a wide range of corruptions (except \textit{Fog}) even if it has lower ``clean" NDS (0.3780 \vs 0.3877). Specifically, under \textit{Color Quant}, \textit{Motion Blur}, and \textit{Dark} corruptions, the mRR metric improves by 22.5\%, 17.2\%, and 27.8\%, respectively. It's worth noting that pre-training mainly improves most semantic corruptions and doesn't improve temporal corruptions. 

Even though, the pre-trained BEVDet still largely lags behind those depth-free counterparts. Therefore, pre-training together with the depth-free bird's eye view transformation provides models with strong robustness.

\begin{figure}[ht]
    \centering
    \subfigure[Camera Crash - CE]{
        \label{fig:nds-mce-cam-temp}
        \includegraphics[width=0.46\linewidth]{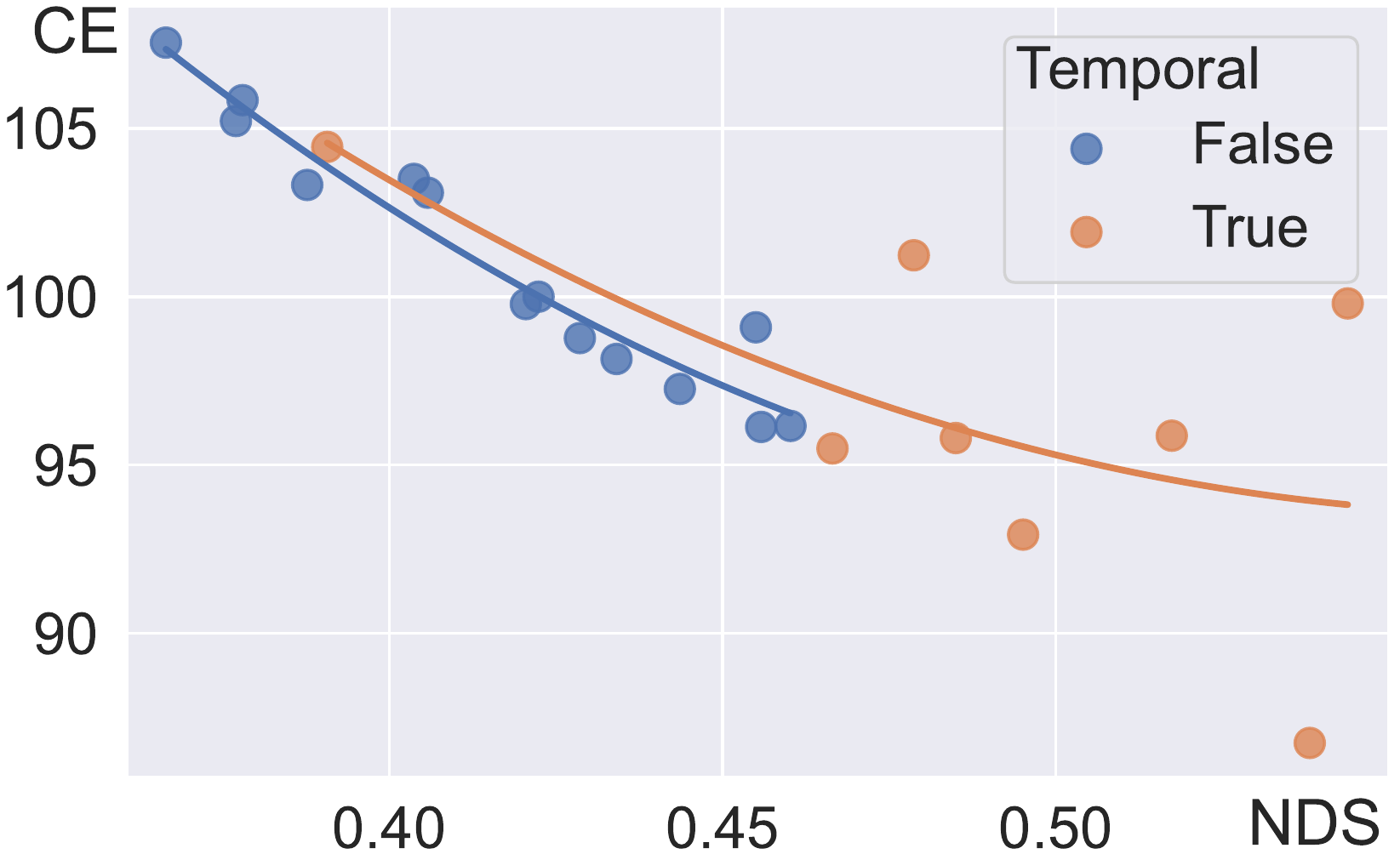}
    }
    \subfigure[Camera Crash - RR]{
        \label{fig:nds-mrr-cam-temp}
        \includegraphics[width=0.46\linewidth]{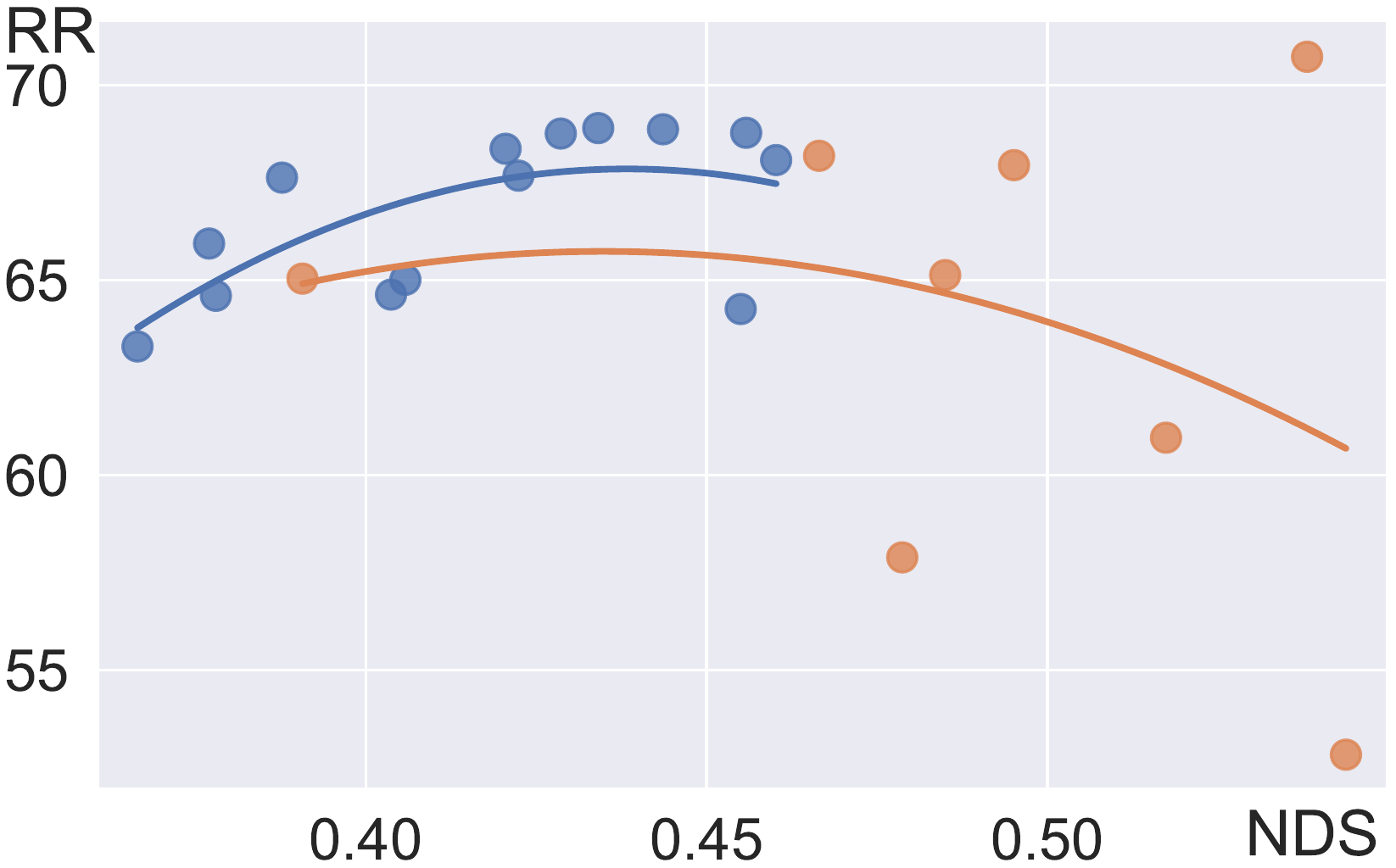}
    }
    \subfigure[Frame Lost - CE]{
        \label{fig:nds-mce-frame-temp}
        \includegraphics[width=0.46\linewidth]{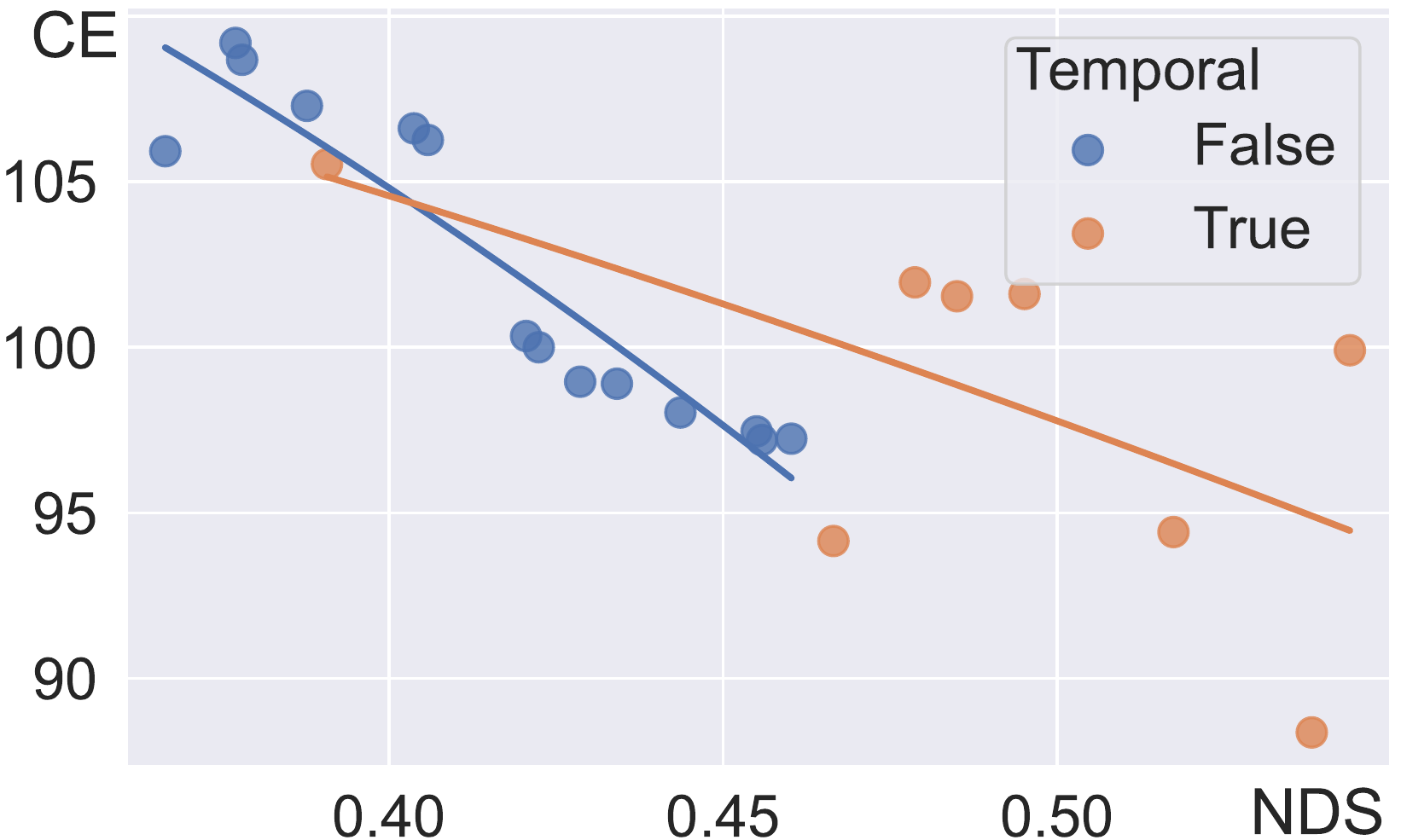}
    }
    \subfigure[Frame Lost - RR]{
        \label{fig:nds-mrr-frame-temp}
        \includegraphics[width=0.46\linewidth]{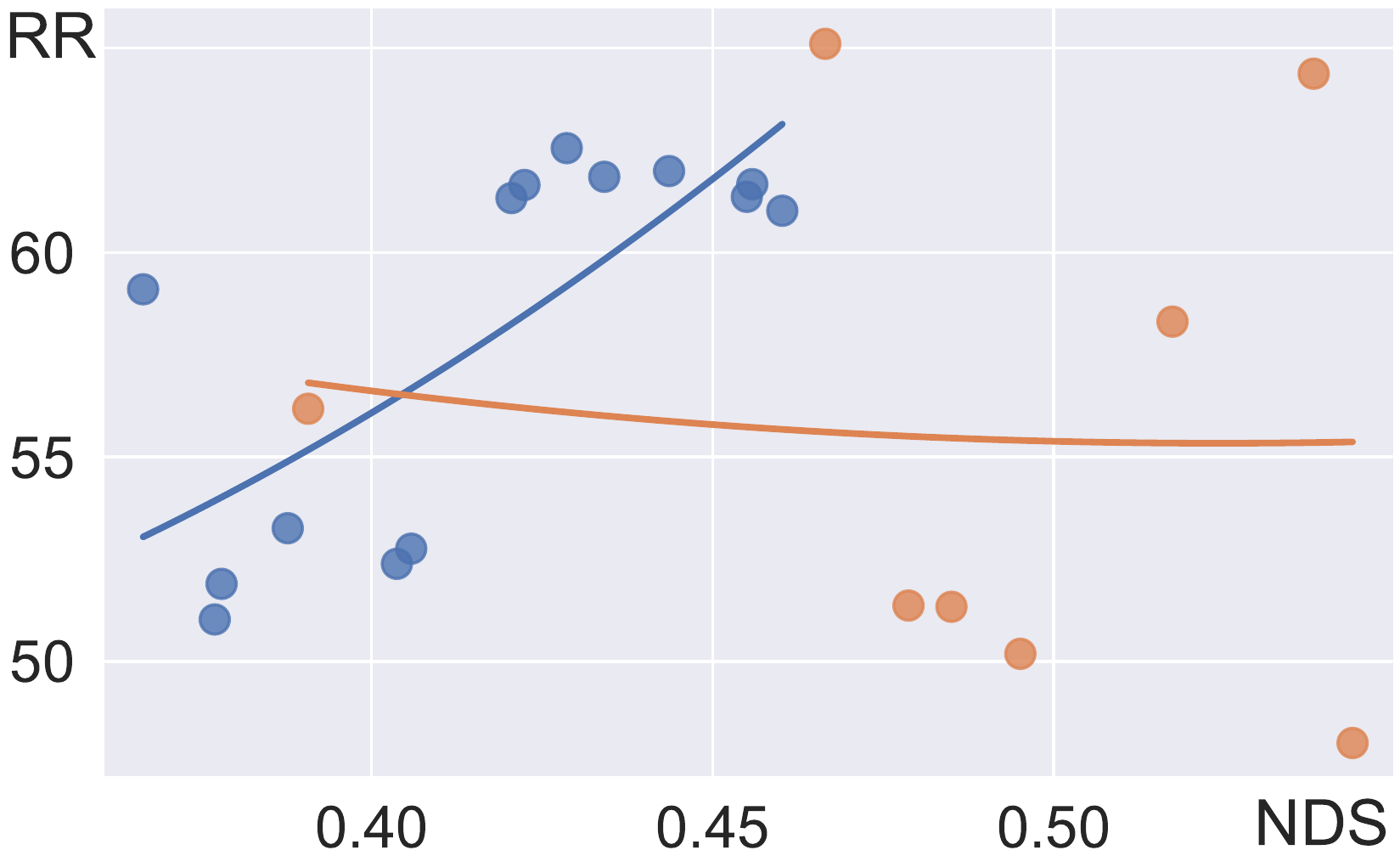}
    }
    \vspace{0.cm}
    \caption{Not all the models with temporal fusion exhibit better robustness under \textit{Camera Crash} and \textit{Frame Lost}. However, they have the potential since the lowest mCE metric models are always those that utilize temporal information.}
    \label{fig:temporal}
\end{figure}

\begin{figure}[ht]
    \centering
    \subfigure[ResNet \vs VoVNet-V2. Since the two versions have similar ``clean" performances, we compare the absolute corruption error (the lower the better).]{
        \label{fig:backbone}
        \includegraphics[width=\linewidth]{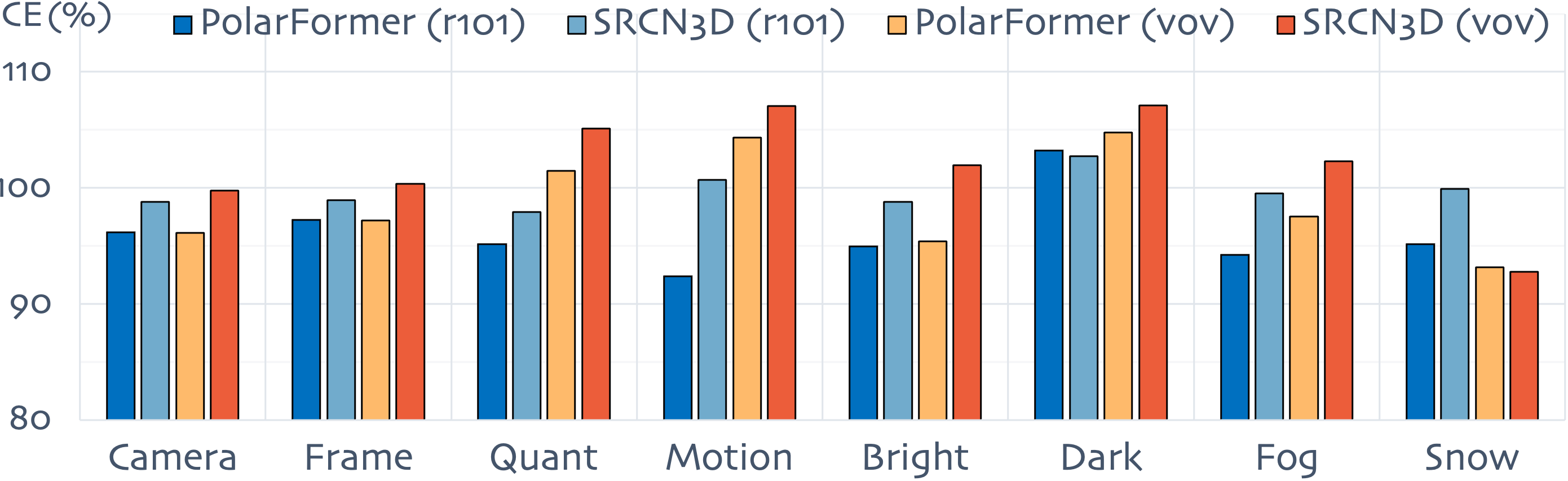}
    }
    \subfigure[ResNet \vs SwinTransformer. Since the models have different ``clean" performances, we compare the relative resilience rate (the higher the better).]{
        \label{fig:swin-backbone}
        \includegraphics[width=\linewidth]{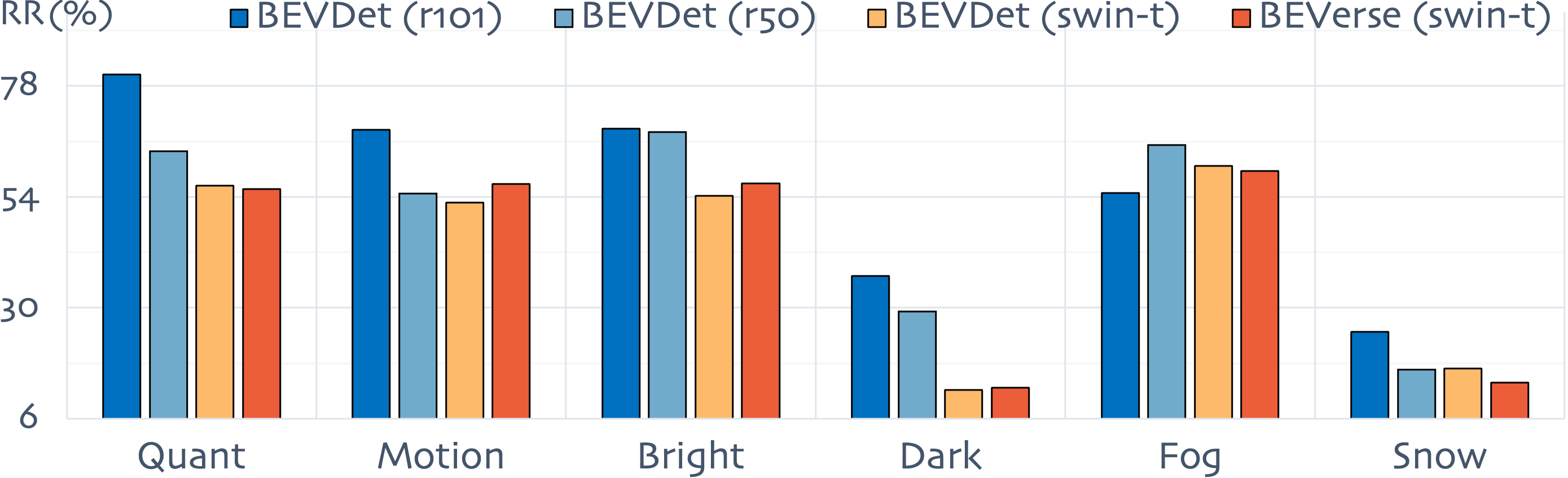}
    }
    \caption{Comparisons of backbones under different corruptions.}
    \vspace{-0.5cm}
\end{figure}

\subsection{Temporal Fusion}
\textbf{\textit{Temporal fusion has the potential to yield better absolute performance under corruptions. Fusing longer temporal information largely helps with robustness.}}
In the context of autonomous driving, accurately estimating the velocity of moving objects based on a single-frame input is considerably challenging. Therefore, leveraging temporal cues to better perceive the surrounding environment is crucial for vision systems. Previous studies have proposed several approaches to learning effective temporal cues. In this investigation, we analyze whether temporal information can alleviate the impact of corrupted images. To this end, we choose BEVFormer~\cite{li2022bevformer} and BEVerse~\cite{zhang2022beverse} for examination since both have single-frame and multi-frame versions, enabling us to examine the influence of temporal information solely. Our experimental findings reveal that the use of multi-frame information substantially enhances the robustness of both the mCE and mRR metrics for BEVerse. For instance, the BEVerse Small improves from 132.12 to 117.82 on mCE and from 29.54 to 49.57 on mRR. However, for BEVFormer, the temporal cross-attention operation enhances the mCE metric but deteriorates the mRR metric from 69.33 to 60.40. We conjecture this might be caused by accumulation errors inside the history BEV features since BEVFormer stores the history information inside and update it on-the-fly while BEVerse consumes a fixed-size temporal window. For SOLOFusion~\cite{Park2022TimeWT}, the RR metric of its long-only version outperforms its short-only counterpart on a wide range of corruption types, which indicates the great potential of utilizing longer temporal information.

\begin{figure*}[ht]
    \centering
    \subfigure[DETR3D]{
        \includegraphics[width=0.23\linewidth]{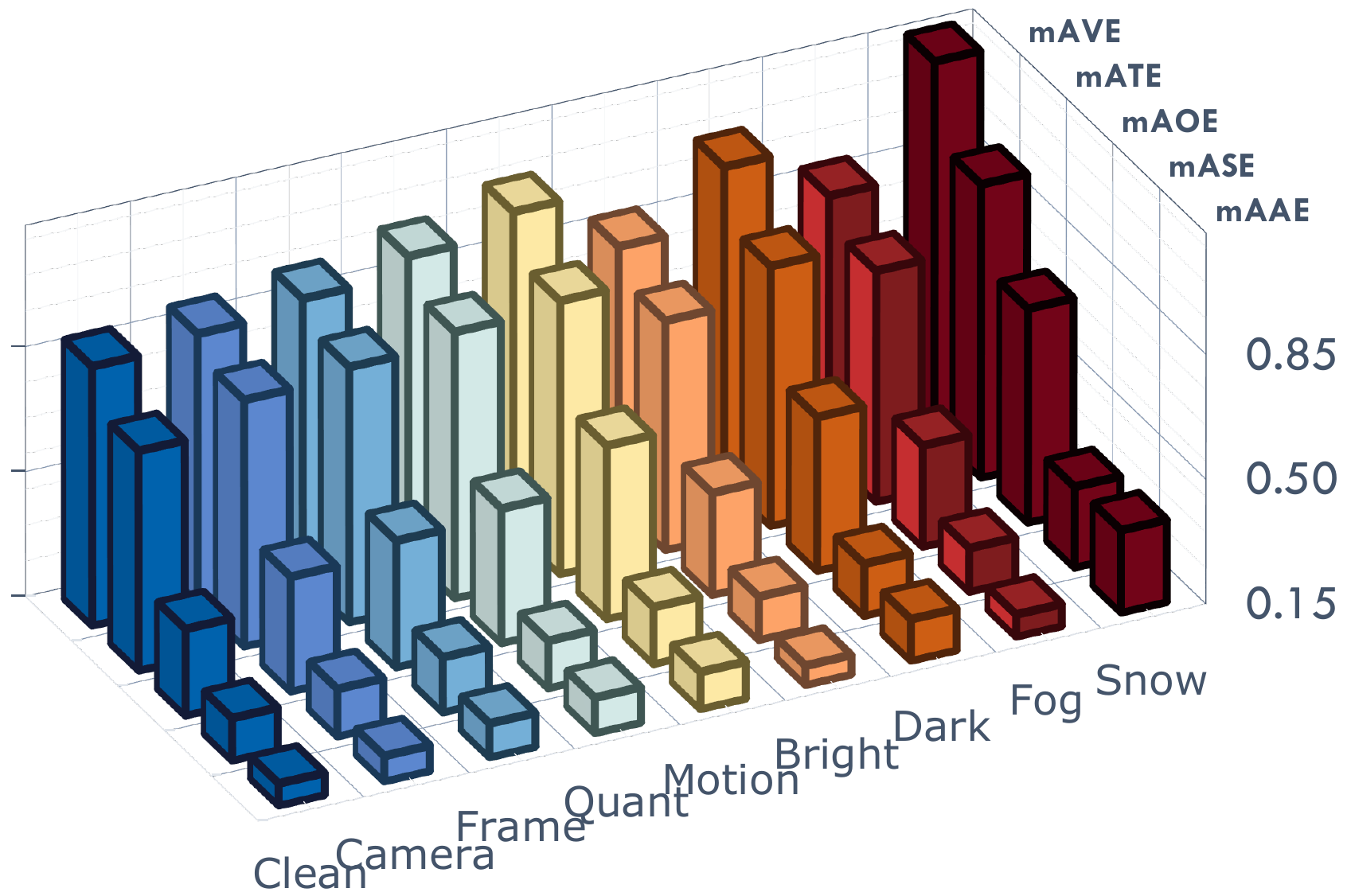}
        \label{fig:3d-detr}
    }
    \subfigure[BEVFormer]{
        \includegraphics[width=0.23\linewidth]{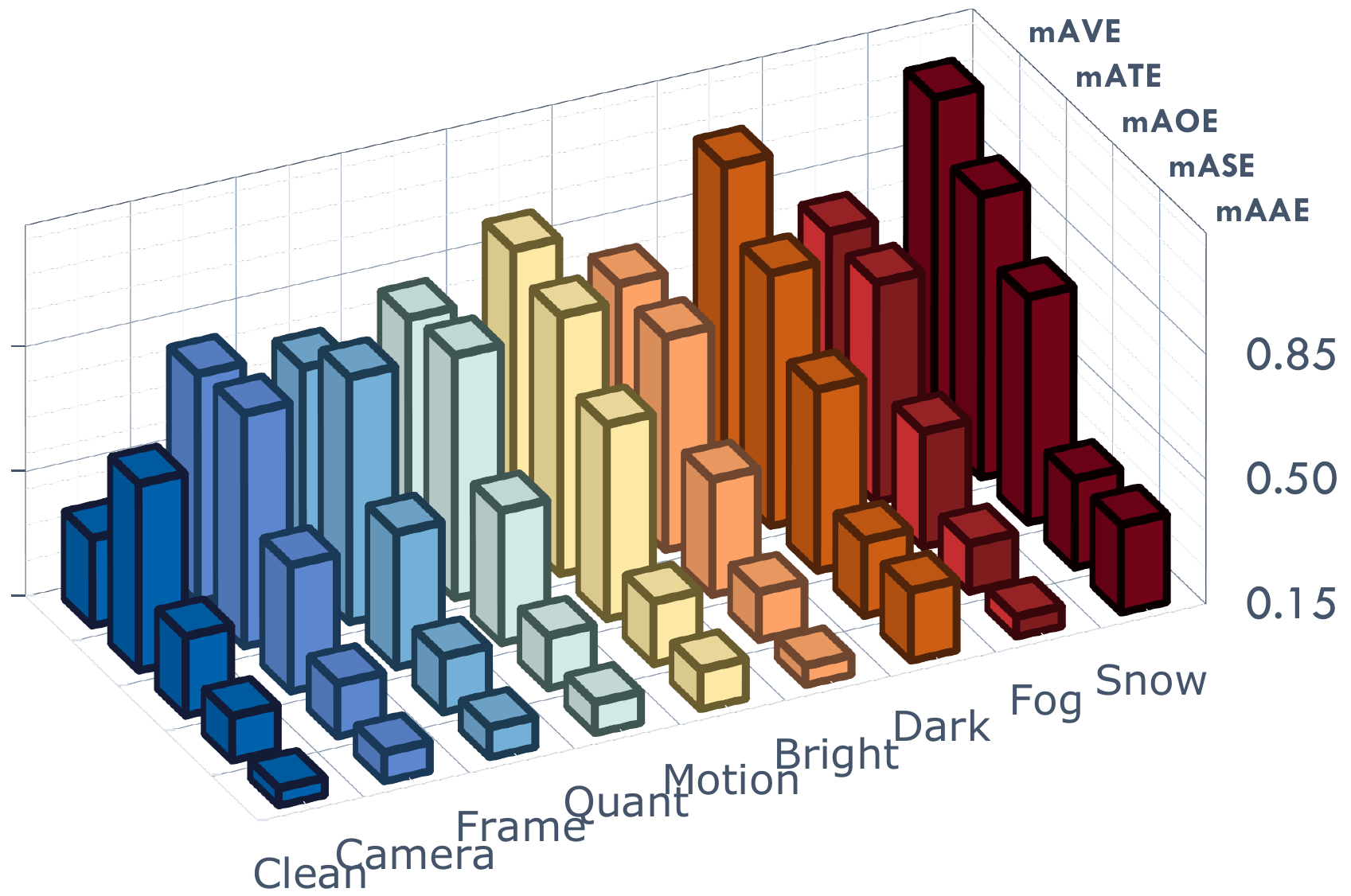}
        \label{fig:3d-bevformer}
    }
    \subfigure[BEVDepth]{
        \includegraphics[width=0.23\linewidth]{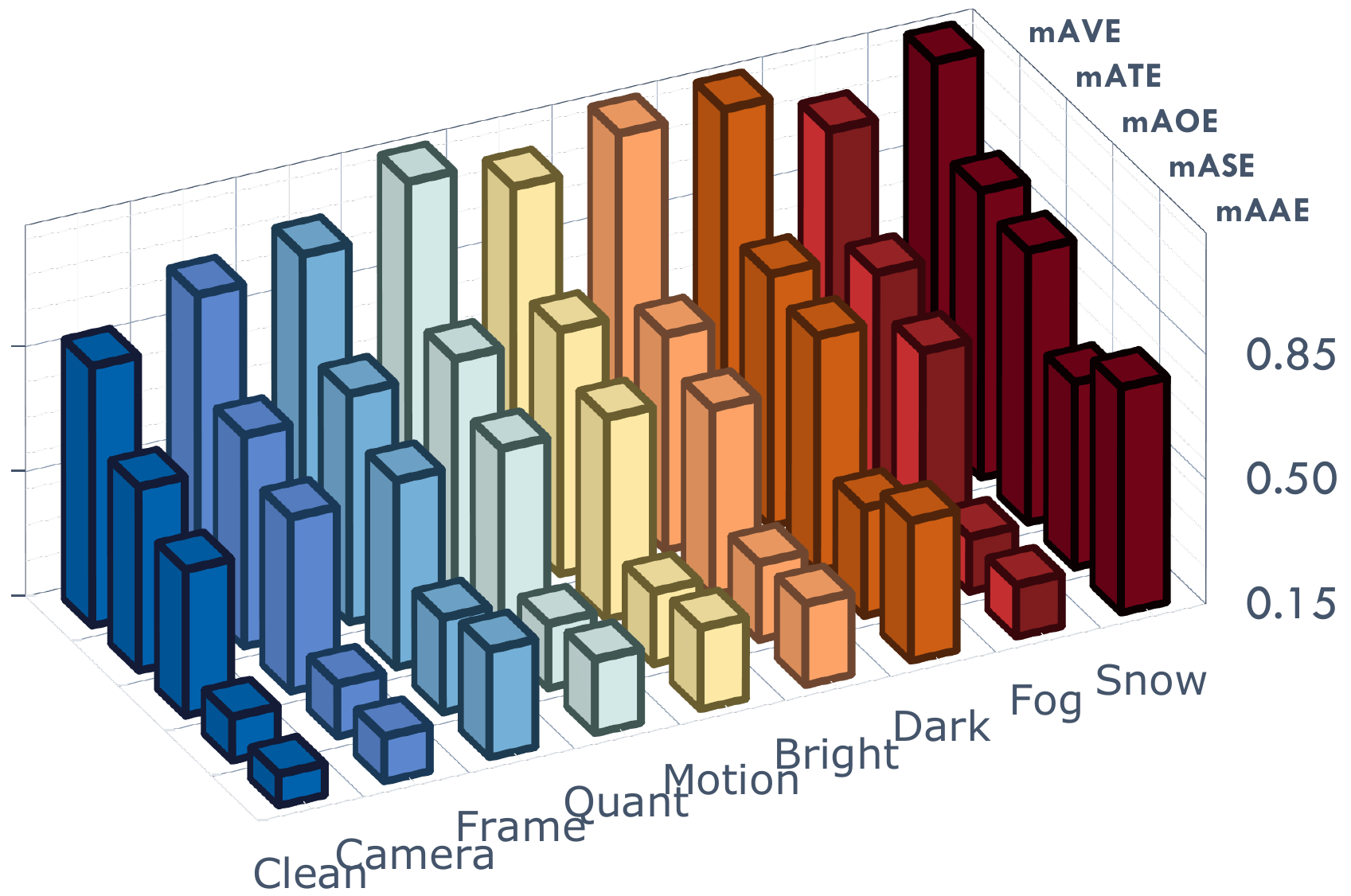}
        \label{fig:3d-bevdepth}
    }
    \subfigure[BEVerse]{
        \includegraphics[width=0.23\linewidth]{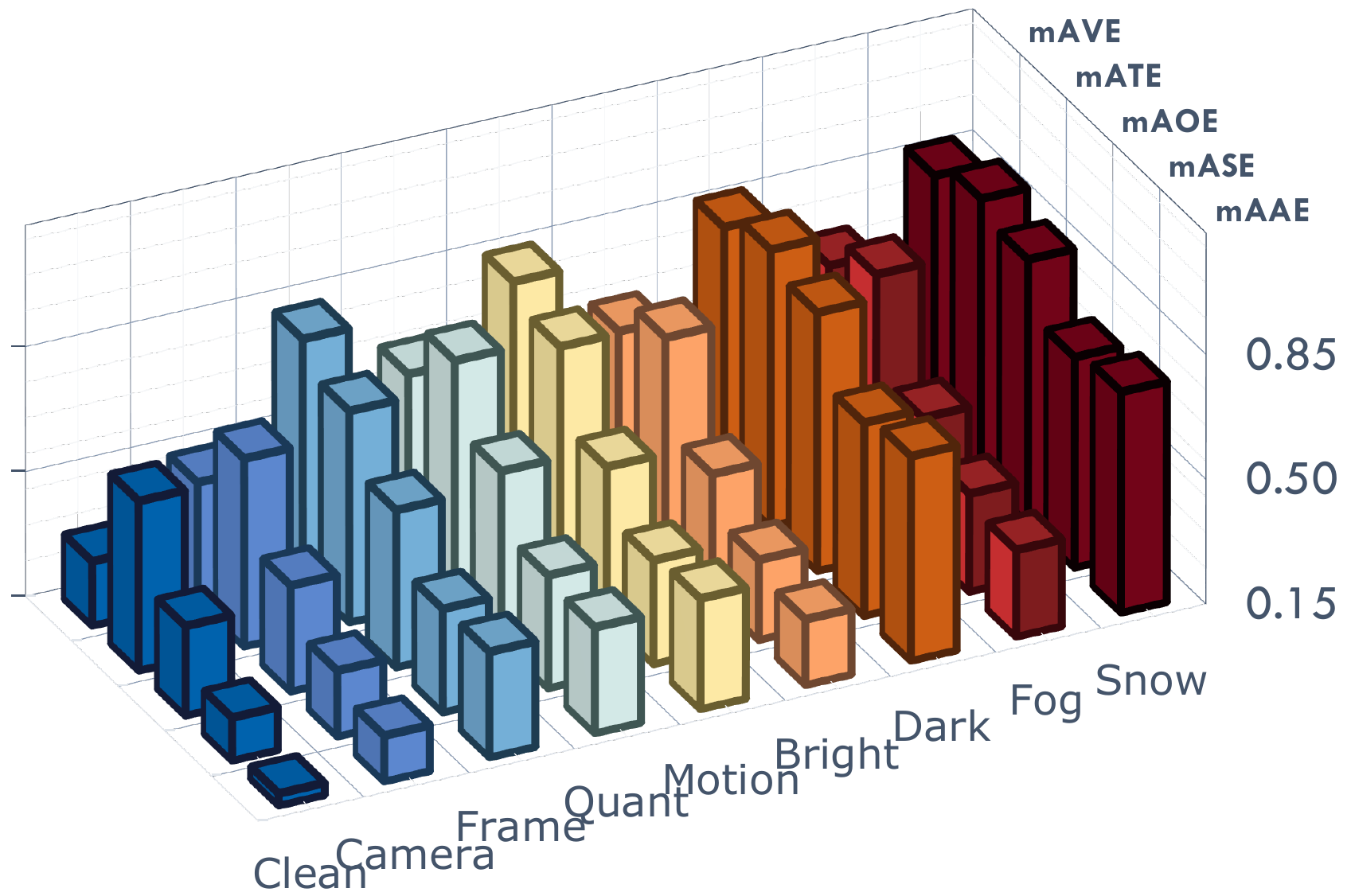}
        \label{fig:3d-beverse}
    }
    \caption{The detailed 3D detection metrics (mAVE, mATE, mAOE, mASE, and mAAE) reported on \textit{nuScenes-C} other than NDS.}
    \label{fig:stat-3d}
\end{figure*}

We are particularly interested in examining how models utilizing temporal information perform under temporal corruptions. We find SOLOFusion which fuses wider and richer temporal information performs extremely well compared to its short-only and long-only versions. In terms of \textit{Camera Crash}, the short-only and long-only versions have close resilience rate performance (65.04 \vs 65.13). However, the fusion version improves to 70.73, which is the highest among all the candidate models. Similarly, the fusion version improves the resilience rate by almost 10\%  compared to the other two versions under \textit{Frame Lost} corruption.

Surprisingly, we further find that not all models with temporal fusion exhibit better robustness under \textit{Camera Crash} and \textit{Frame Lost}. The robustness is highly correlated to how to fuse history frames and how many frames are used, which emphasizes the importance of evaluating temporal fusion strategies from wider perspectives. The results can be seen in Figure~\ref{fig:temporal}. Nonetheless, temporal fusion remains a potential method to enhance temporal robustness since the models with the lowest Corruption Error (or the highest Resilience Rate) are consistently those that utilize temporal information.

\subsection{Backbone}
\textbf{\textit{The Swin Transformer is more vulnerable towards the lighting changings; VoVNet-V2 is more robust against Snow while ResNet shows better robustness across a wide range of corruptions.}}
It is noteworthy that various backbones demonstrate distinct responses to corruptions. PolarFormer~\cite{jiang2022polarformer} and SRCN3D~\cite{shi2022srcn3d}, each having two backbone versions (\ie, ResNet and VoVNet-V2~\cite{lee2020centermask}), exhibit similar standard performance. Nevertheless, ResNet-based detectors exhibit consistently superior robustness across a wide range of corruptions. For example, PolarFormer (r101) and SRCN3D (r101) demonstrate consistent improvements in the \textit{Color Quant}, \textit{Motion}, \textit{Dark}, and \textit{Fog}, as illustrated in Figure~\ref{fig:backbone}. Conversely, the VoVNet-V2~\cite{lee2020centermask} backbone consistently exhibits better robustness under \textit{Snow} corruptions. 

Moreover,  Swin Transformer~\cite{liu2021swin} based BEVDet demonstrates significant vulnerability towards changes in lighting conditions (\eg, \textit{Bright} and \textit{Dark}). A clear comparison can be found between the ResNet-based BEVDet and the Swin-based BEVDet as shown in Figure~\ref{fig:swin-backbone}. Additionally, the Swin-based BEVerse shows similar vulnerability towards these two corruptions, further supporting this conclusion.

\subsection{Detailed nuScenes Metrics}
\textbf{\textit{The velocity prediction error significantly increases under corruptions, even for models with temporal fusion. The attribution and scale errors vary among models.}}
The study primarily presents the NDS metrics, while additional detailed information regarding the model's robustness can be found in Figure~\ref{fig:stat-3d}. It is observed that models employing temporal information have significantly lower mAVE metrics than those without such incorporation (\eg, BEVFormer~\cite{li2022bevformer} and BEVerse~\cite{zhang2022beverse}). However, in the presence of corrupted images, even under mild illumination conditions, the velocity prediction error significantly increases. Notably, the \textit{Motion Blur} corruption also adversely affects the velocity predictions for BEVFormer and BEVerse, as shown in Figure~\ref{fig:3d-bevformer} and \ref{fig:3d-beverse}. This highlights the vulnerability of models, which are even equipped with temporal information, in evaluating velocity under corrupted images.

Furthermore, attribution and scale errors show considerable variations across different models. For depth-free approaches, the errors of these metrics remain consistent, whereas there is a notable disparity for depth-based approaches. These findings further emphasize the vulnerability of depth-based approaches to corruptions.

%% file: sections/06_conclusion.tex
\section{Conclusions}
\label{sec:con}

In this work, we introduce the \textit{RoboBEV} benchmark by utilizing a corpus of eight diverse natural corruptions to create the \textit{nuScenes-C} dataset, which is used to assess the out-of-distribution robustness of camera-only BEV perception approaches. Exhaustive experiments are conducted to investigate the factors that impact the model's robustness. We hope the findings of this study provide valuable insights for designing future models that can achieve better out-of-distribution robustness.

\section{Potential Limitations}
Despite the eight distinct corruptions we introduce, they still cannot cover all the out-of-distribution contexts in real-world applications due to their unpredictable complexity. 
Additionally, we mainly analyze coarse-grained designs between models (\eg, depth estimation) since it is considerably non-trivial to identify the trade-off between fine-grained network architecture designs (\eg, the difference between SRCN3D and Sparse4D, and the different temporal fusion strategies for BEVFormer and BEVerse). 

%% file: sections/07_appendix.tex
\section*{Appendix}

In this appendix, we supplement the following materials to better support the findings and conclusions in the main body of this paper:

\begin{itemize}
    \item \cref{sec:addition-implement} includes more details about the corruption setting, severity level, and model training.
    \item \cref{sec:additional_experimental_result} provides the full \textit{RoboBEV} benchmark results and some visualization results.
    \item \cref{sec:more-analysis} covers additional analyses of the experimental results in our benchmark.
    \item \cref{sec:public-resources-used} acknowledges the public resources used during the course of this work.
\end{itemize}

\section{Additional Implementation Details}
\label{sec:addition-implement}

\subsection{Corruption Definition}
We follow the severity protocol (\ie, from severity 1 to 5) setting in ImageNet-C~\cite{hendrycks2019benchmarking}. We choose the severity carefully to avoid drastic performance degeneration, which might impede us from drawing reliable conclusions. The details can be seen in Table~\ref{tab:corruption-setting}. 

For \textit{Camera Crash}, we first randomly choose a set of cameras, and the number of the dropped cameras is equal to the severity (\eg, from one to five). In terms of \textit{Frame Lost}, we randomly drop each camera with an identical probability at every single input frame in Equation~\ref{eq:frame}.
\begin{equation}
    p_{\rm drop} = {\rm severity} \times \frac{1}{6}.
    \label{eq:frame}
\end{equation}

The \textit{Camera Crash} corruption evaluates the model's behaviors when images from certain viewpoints are continuously lost, while the \textit{Frame Lost} evaluates whether the model can make predictions by utilizing the last frame from the same sensors.

\input{tables/corruption_setting}

\subsection{Re-Implementation}
We re-implement the BEVDet (r101) \cite{huang2021bevdet}, PolarFormer (vov) \cite{jiang2022polarformer}, and SRCN3D (vov) \cite{shi2022srcn3d} models. For BEVDet (r101), we keep the input resolution the same as BEVDet(r50) for a fair comparison. Although this leads to lower results than the official report~\cite{huang2021bevdet}, we focus more on the robustness metric and draw meaningful conclusions rather than better performance on nuScenes~\cite{caesar2020nuscenes} dataset. 

The original PolarFormer (vov) and SRCN3D (vov) are initialized from DD3D~\cite{park2021pseudo} checkpoint, which uses nuScenes \textit{trainval} set for training. This causes information leakage since the nuScenes-C dataset is corrupted from the nuScenes validation set. For fair comparisons, we re-implement these two models without modification but use the FCOS3D~\cite{wang2021fcos3d} model as initialization. The VoVNet-V2 \cite{lee2020centermask} version FCOS3D models are first trained with depth estimation on the DDAD15M~\cite{guizilini20203d} dataset and then fine-tuned on the nuScenes train set.

\begin{figure*}[ht]
    \centering
    \includegraphics[width=\linewidth]{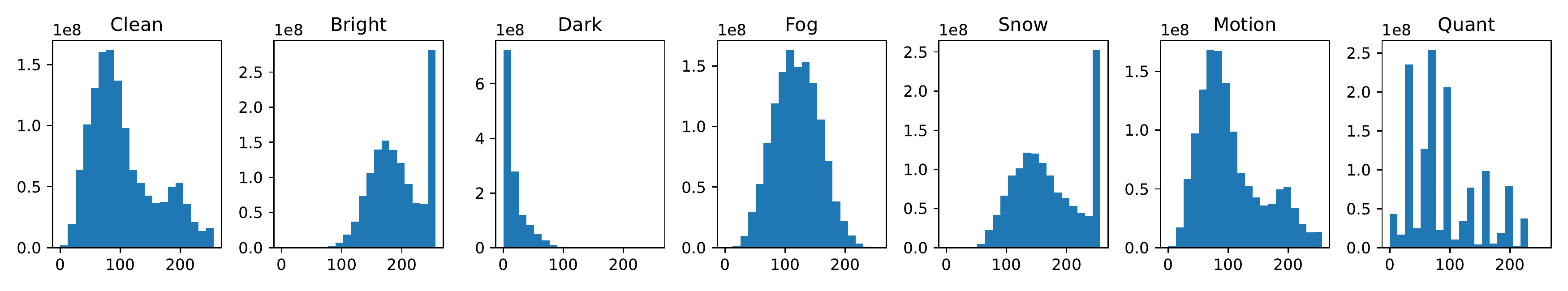}
    \caption{The pixel histogram of each corruption type in the \textit{nuScenes-C} dataset.}
    \label{fig:hist}
\end{figure*}

\section{Additional Experimental Results}
\label{sec:additional_experimental_result}
In this section, we provide the full evaluation results of benchmarked models in the \textit{RoboBEV} benchmark.

\subsection{Map Segmentation}
We further consider BEV-based map segmentation tasks. The results can be seen in Table~\ref{tab:appe-map-seg}. We use setting 1 as mentioned in \cite{pan2020cross} to report the Intersection over Union (IoU) with a threshold of 0.4 of vehicle map-view segmentation results. The performance of BEV-based map segmentation Cross-View-Transformer is also severally hampered by corruption such as \textit{Frame Lost} and \textit{Snow}.

\input{tables/map_seg}

\subsection{Corruption Robustness}
We attach the full results, including monocular FCOS3D~\cite{wang2021fcos3d}, please refer to \cref{tab:detr3d-results} to \cref{tab:fcos3d-results}.

\begin{itemize}
    \item DETR3D~\cite{wang2022detr3d}: See \cref{tab:detr3d-results} for the results of the standard model; see \cref{tab:detr3d-cbgs-results} for the results of DETR3D$_{\text{CBGS}}$.
    \item ORA3D~\cite{roh2022ora3d}: See \cref{tab:ora3d-results} for ORA3D model.
    \item BEVFormer~\cite{li2022bevformer}: Refer to \cref{tab:bevformer-base} and \cref{tab:bevformer-small} for models that turn on multi-frame inference and \cref{tab:bevformer-s-base} and \cref{tab:bevformer-s-small} for single-frame versions.
    \item PETR~\cite{liu2022petr}: Table~\ref{tab:petr-r50} and Table~\ref{tab:petr-vov} show results of PETR (r50) and PETR (vov), respectively.
    \item PolarFormer~\cite{jiang2022polarformer}: Table~\ref{tab:polarformer-r50} and Table~\ref{tab:polarformer-vov} show results of PolarFormer (r101) and our re-implement PolarFormer (vov).
    \item SRCN3D~\cite{shi2022srcn3d}: See \cref{tab:srcn3d-r101} and \cref{tab:srcn3d-vov} for SRCN3D (r101) model and our re-implement SRCN3D (vov) model.
    \item Sparse4D~\cite{lin2022sparse4d}: See \cref{tab:sparse4d-results} for Sparse4D (r101) model.
    \item BEVDet~\cite{huang2021bevdet}: Refer to \cref{tab:bevdet-r50-results} and \cref{tab:bevdet-swint-results} for official implement; for our ResNet101 version implement, see \cref{tab:bevdet-r101-results} and \cref{tab:bevdet-r101-fcos3d-results}.
    \item BEVDepth~\cite{li2022bevdepth}: See \cref{tab:bevdepth-r50-results} for BEVDepth (r50) model.
    \item BEVerse~\cite{zhang2022beverse}: See \cref{tab:beverse-tiny-results} and \cref{tab:beverse-small-results} for temporal models; see \cref{tab:beverse-s-tiny-results} and \cref{tab:beverse-s-small-results} for single-frame models.
    \item SOLOFusion~\cite{Park2022TimeWT}: See \cref{tab:solofusion-s}, \ref{tab:solofusion-l}, and \ref{tab:solofusion}.
    \item FCOS3D~\cite{wang2021fcos3d}: See \cref{tab:fcos3d-results}.
\end{itemize}

\subsection{More Visualization Results}
\label{sec:app-more-visual}
\begin{figure*}[ht]
    \centering
    \includegraphics[width=0.9\linewidth]{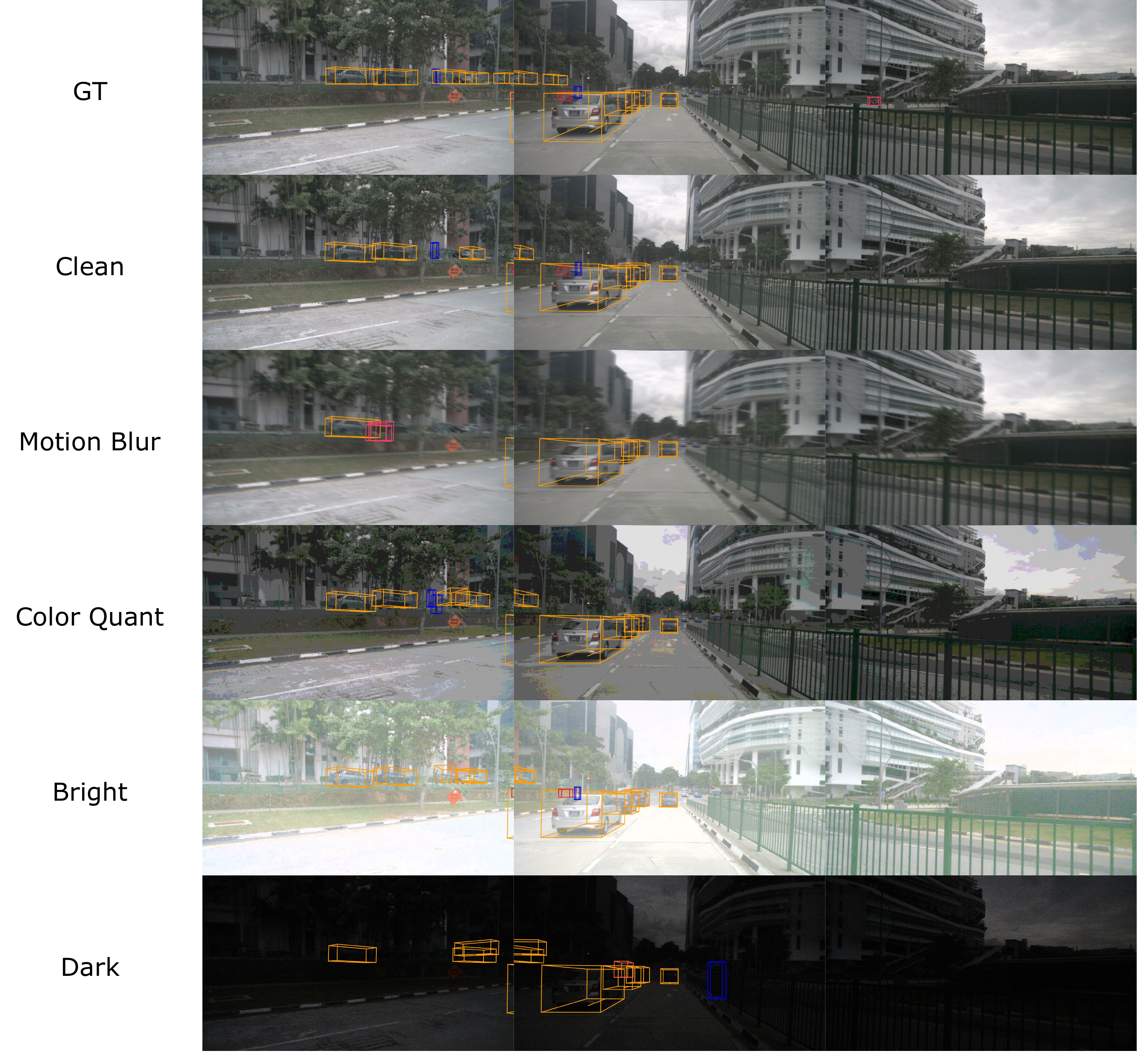}
    \caption{Visualization results of BEVFormer~\cite{li2022bevformer} under moderate corruptions.}
    \label{fig:demo-1}
\end{figure*}

\begin{figure*}[htbp]
    \centering
    \includegraphics[width=0.95\linewidth]{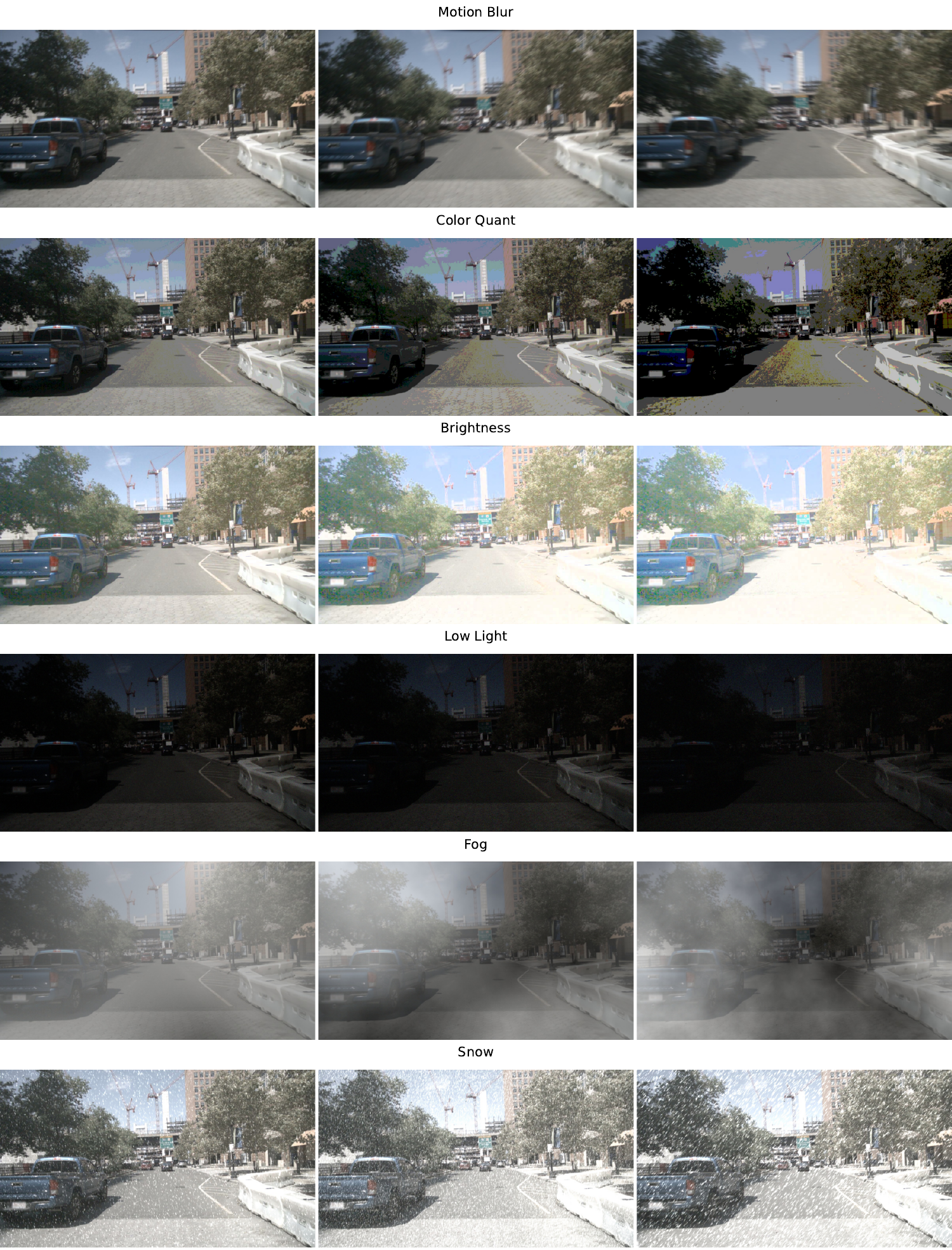}
    \vspace{0.2cm}
    \caption{More visualization results of the \textit{nuScenes-C} dataset. Severity levels from left to right: easy, moderate, and hard.}
    \label{fig:nuscenes-c}
\end{figure*}

We provide more visualization results of the \textit{nuScenes-C} dataset, as shown in Figure~\ref{fig:nuscenes-c}. The prediction results of BEVFormer~\cite{li2022bevformer} under moderate level corruptions can be seen in Figrue~\ref{fig:demo-1}, \ref{fig:demo-2} and \ref{fig:demo-3}.

\section{Additional Analysis}
\label{sec:more-analysis}
\subsection{Corruptions}
\label{sec:app-corruptions}
\begin{figure*}
    \centering
    \subfigure[Camera Crash: CE \vs NDS.]{
        \includegraphics[width=0.23\linewidth]{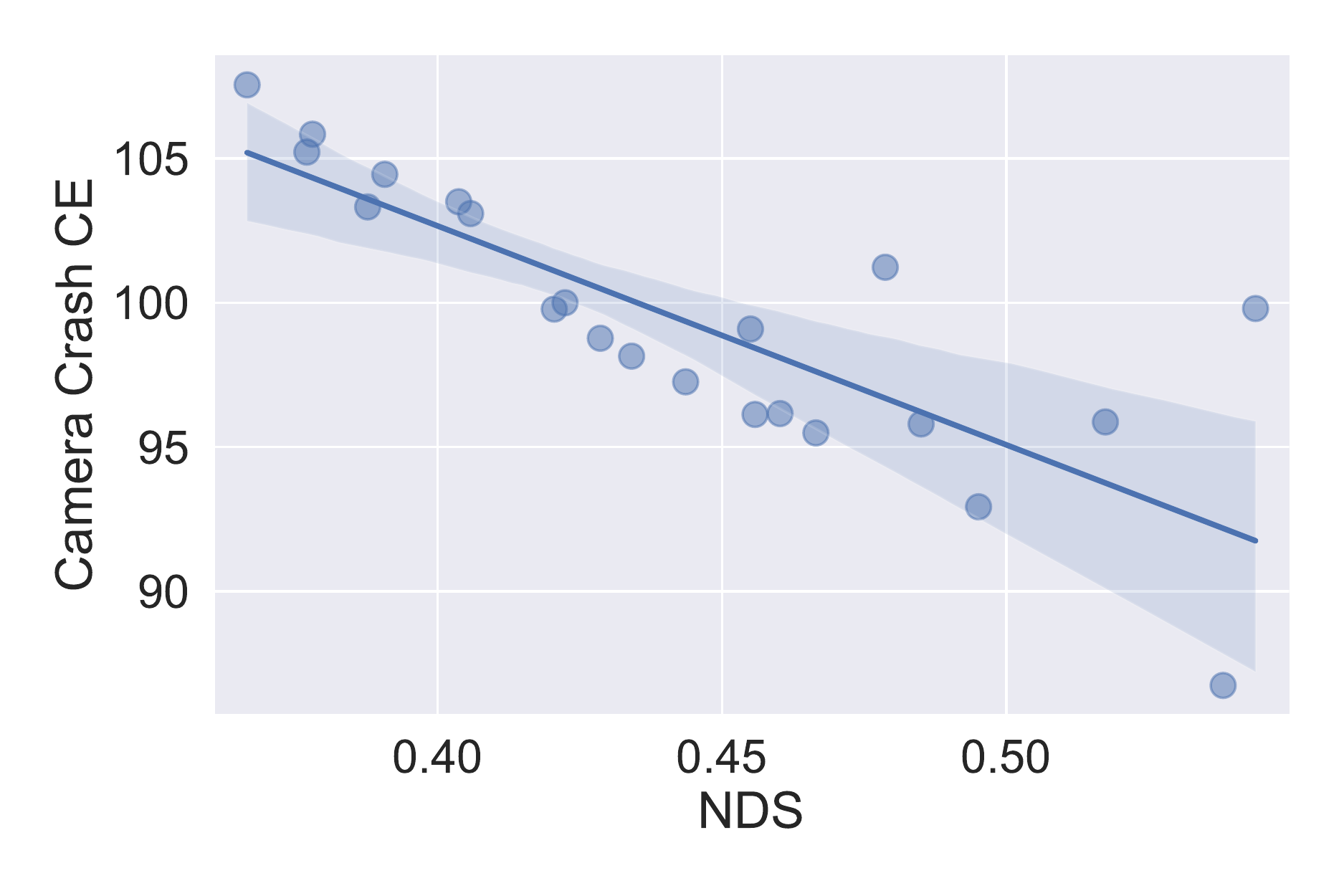}
        \label{fig:camera-ce-nds}
    }
    \subfigure[Camera Crash: RR \vs NDS.]{
        \includegraphics[width=0.23\linewidth]{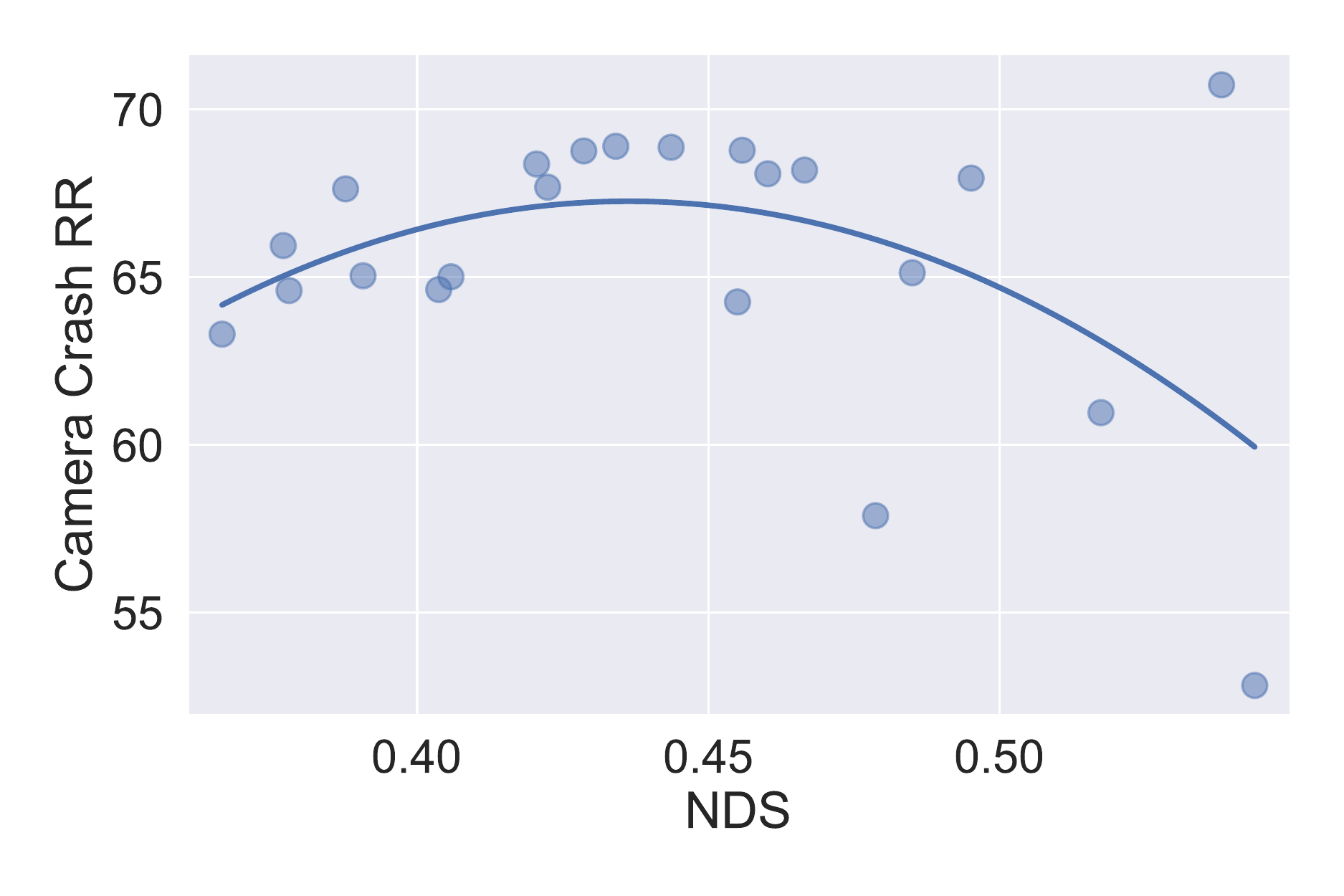}
    }
    \subfigure[Frame Lost: CE \vs NDS.]{
        \includegraphics[width=0.23\linewidth]{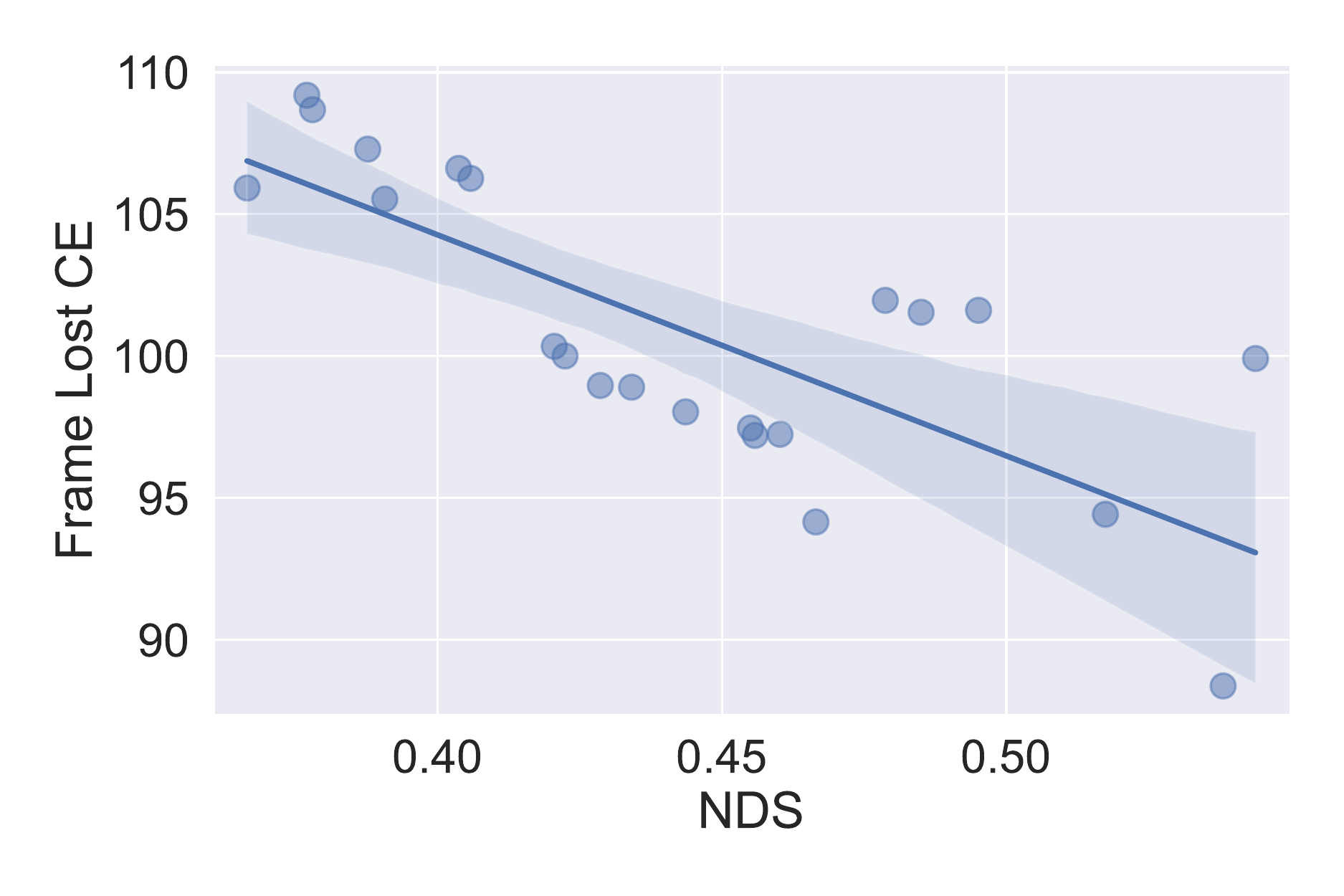}
        \label{fig:frame-ce-nds}
    }
    \subfigure[Frame Lost: RR \vs NDS.]{
        \includegraphics[width=0.23\linewidth]{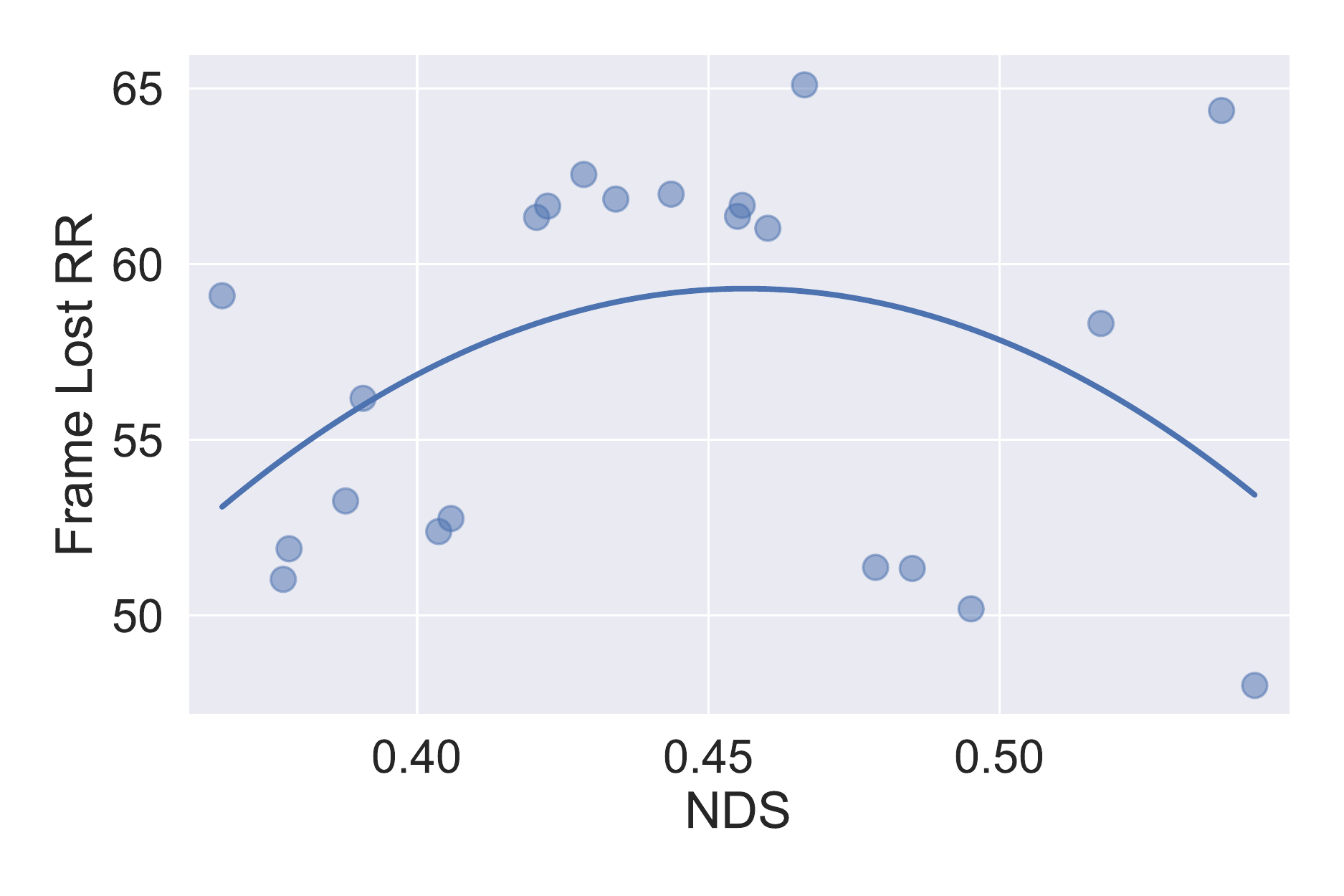}
    }
    \subfigure[Motion Blur: CE \vs NDS.]{
        \includegraphics[width=0.23\linewidth]{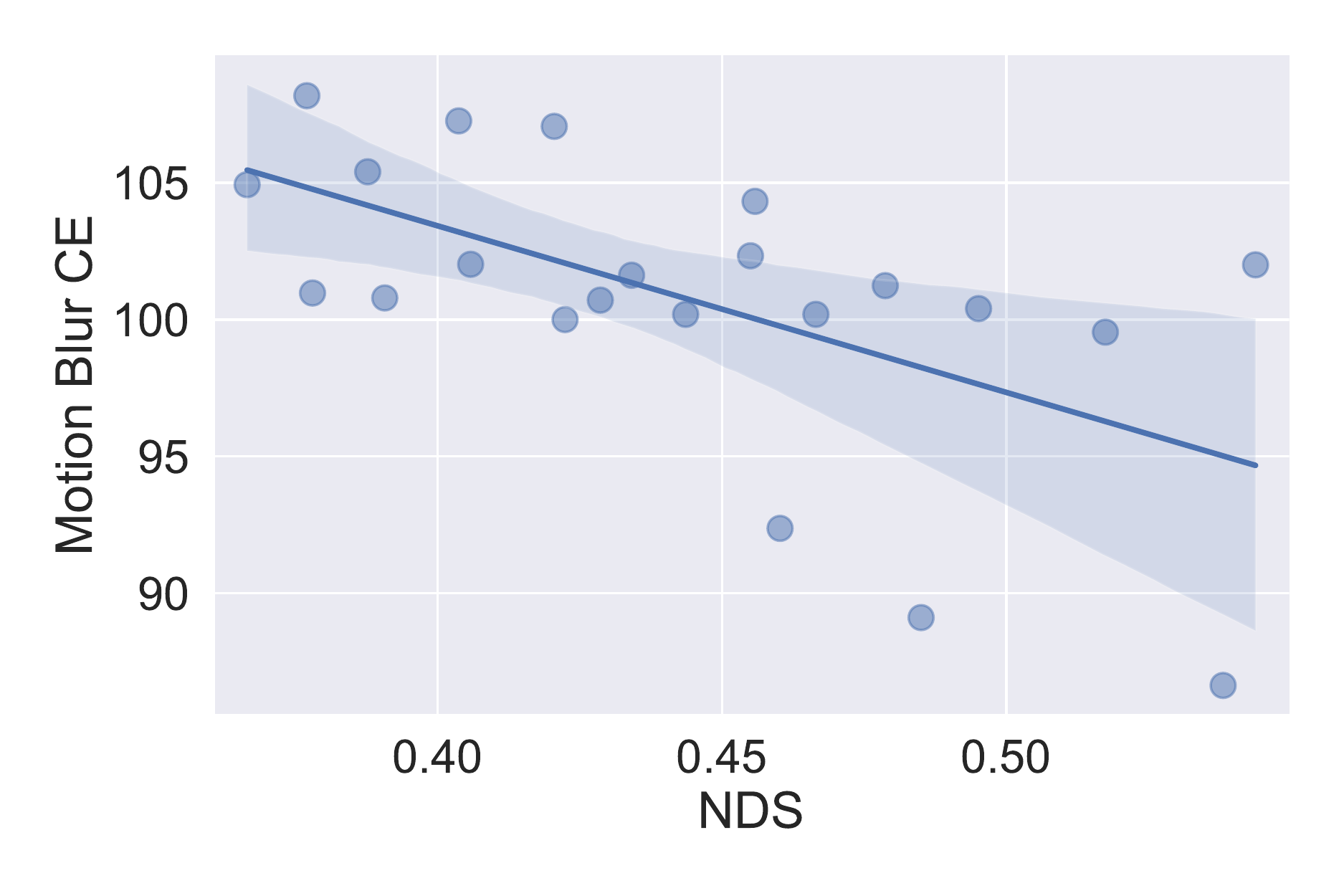}
    }
    \subfigure[Motion Blur: RR \vs NDS.]{
        \includegraphics[width=0.23\linewidth]{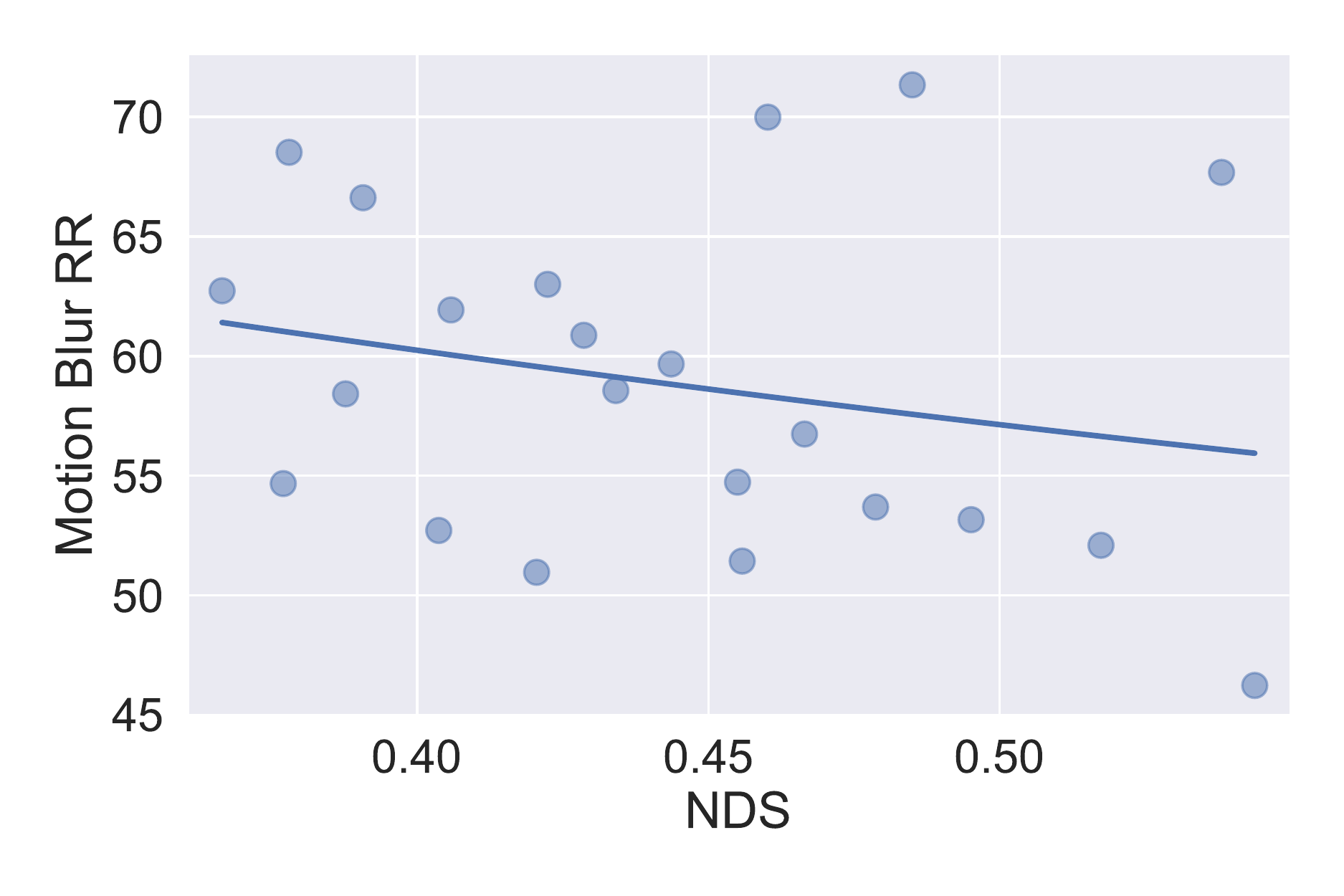}
        \label{fig:motion-rr-nds}
    }
    \subfigure[Color Quant: CE \vs NDS.]{
        \includegraphics[width=0.23\linewidth]{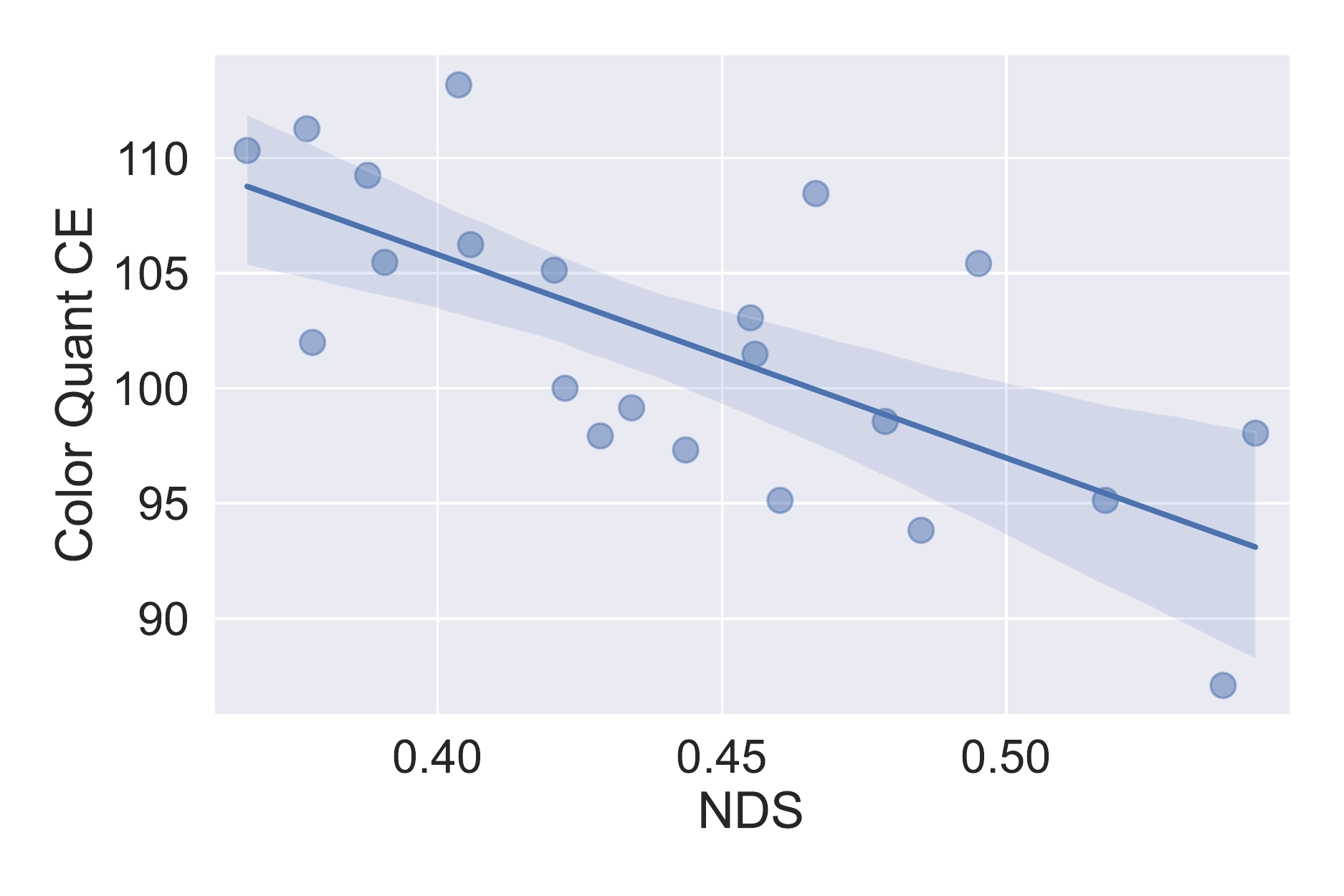}
    }
    \subfigure[Color Quant: RR \vs NDS.]{
        \includegraphics[width=0.23\linewidth]{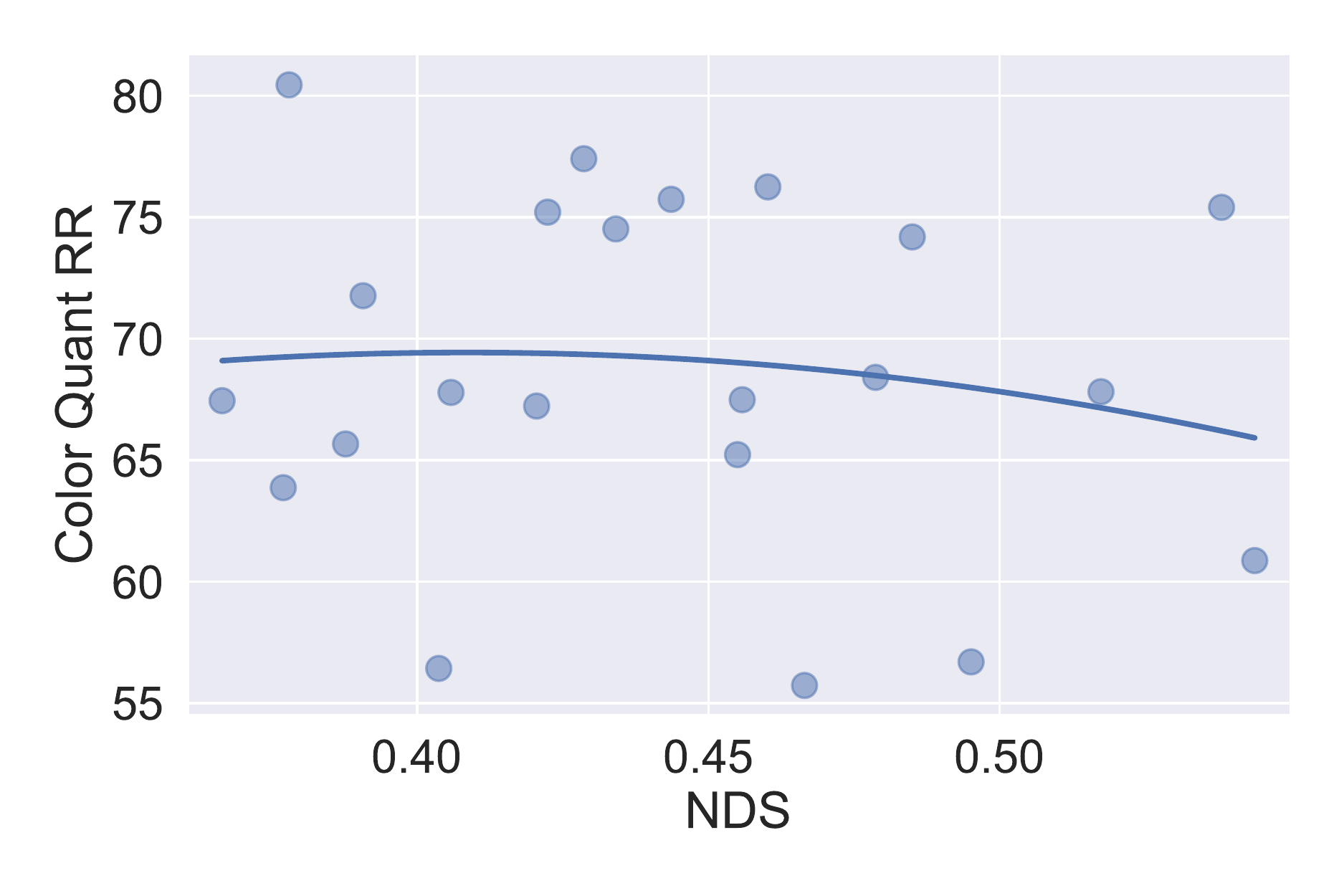}
        \label{fig:quant-rr-nds}
    }
    \subfigure[Bright: CE \vs NDS.]{
        \includegraphics[width=0.23\linewidth]{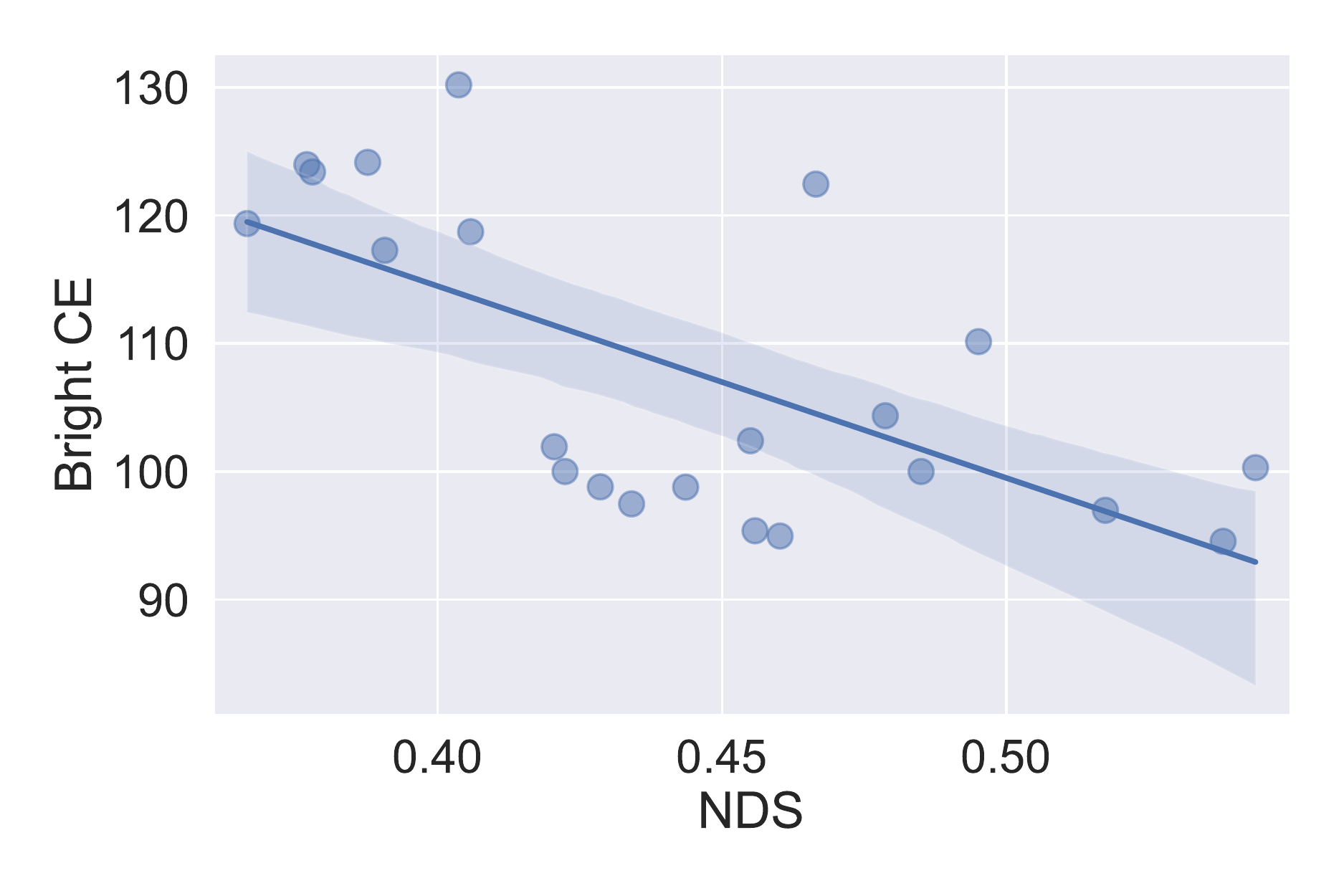}
    }
    \subfigure[Bright: RR \vs NDS.]{
        \includegraphics[width=0.23\linewidth]{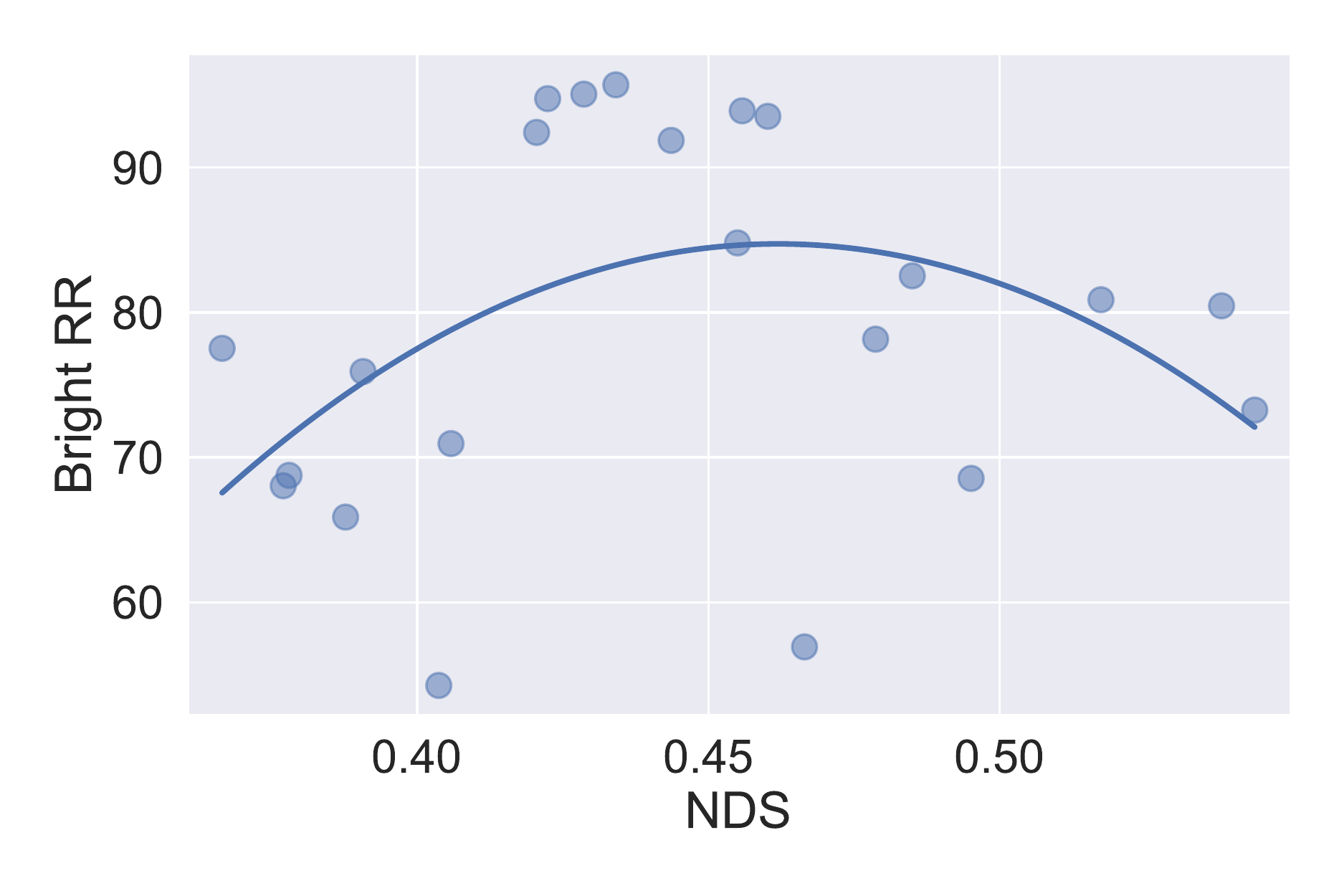}
    }
    \subfigure[Dark: CE \vs NDS.]{
        \includegraphics[width=0.23\linewidth]{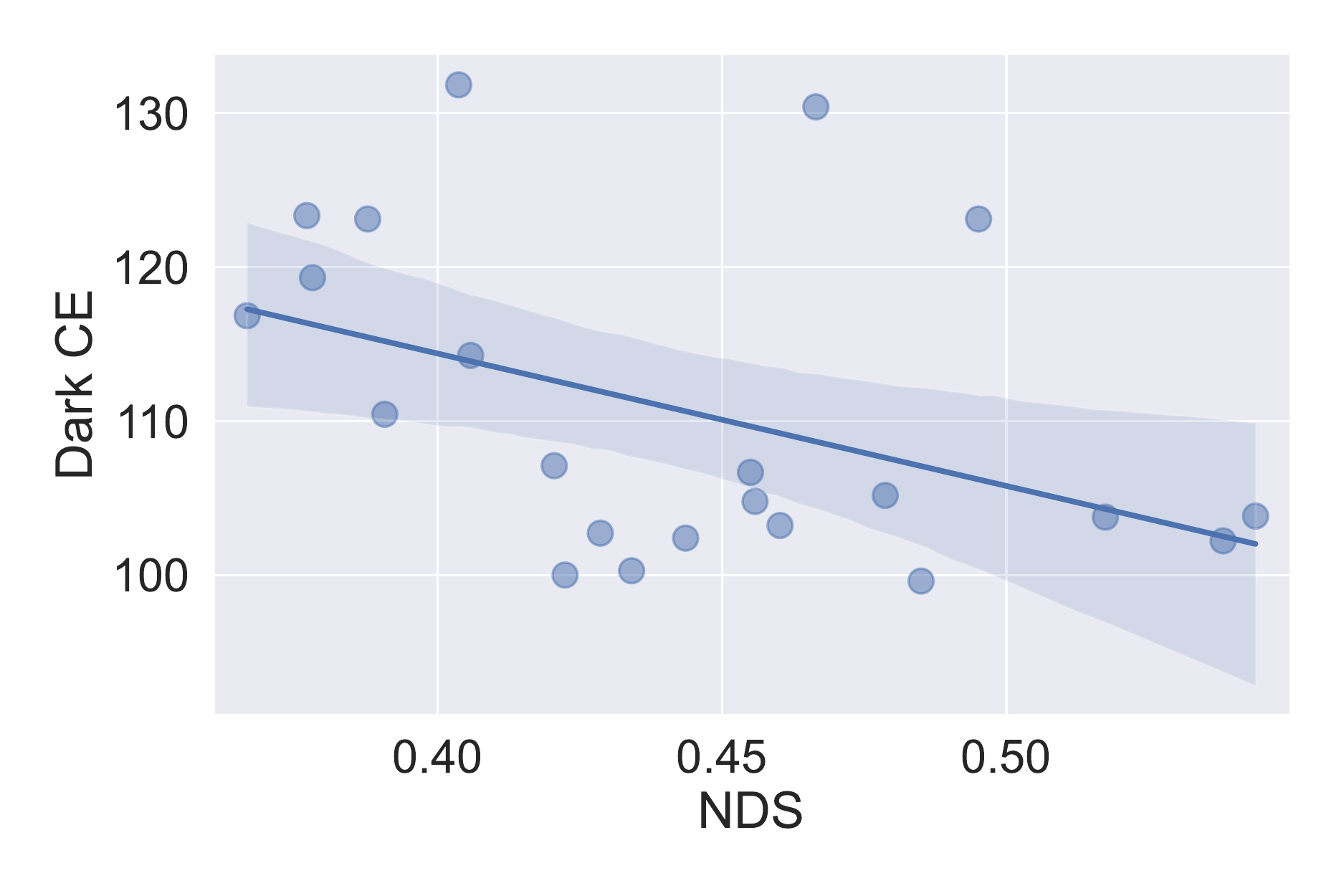}
    }
    \subfigure[Dark: RR \vs NDS.]{
        \includegraphics[width=0.23\linewidth]{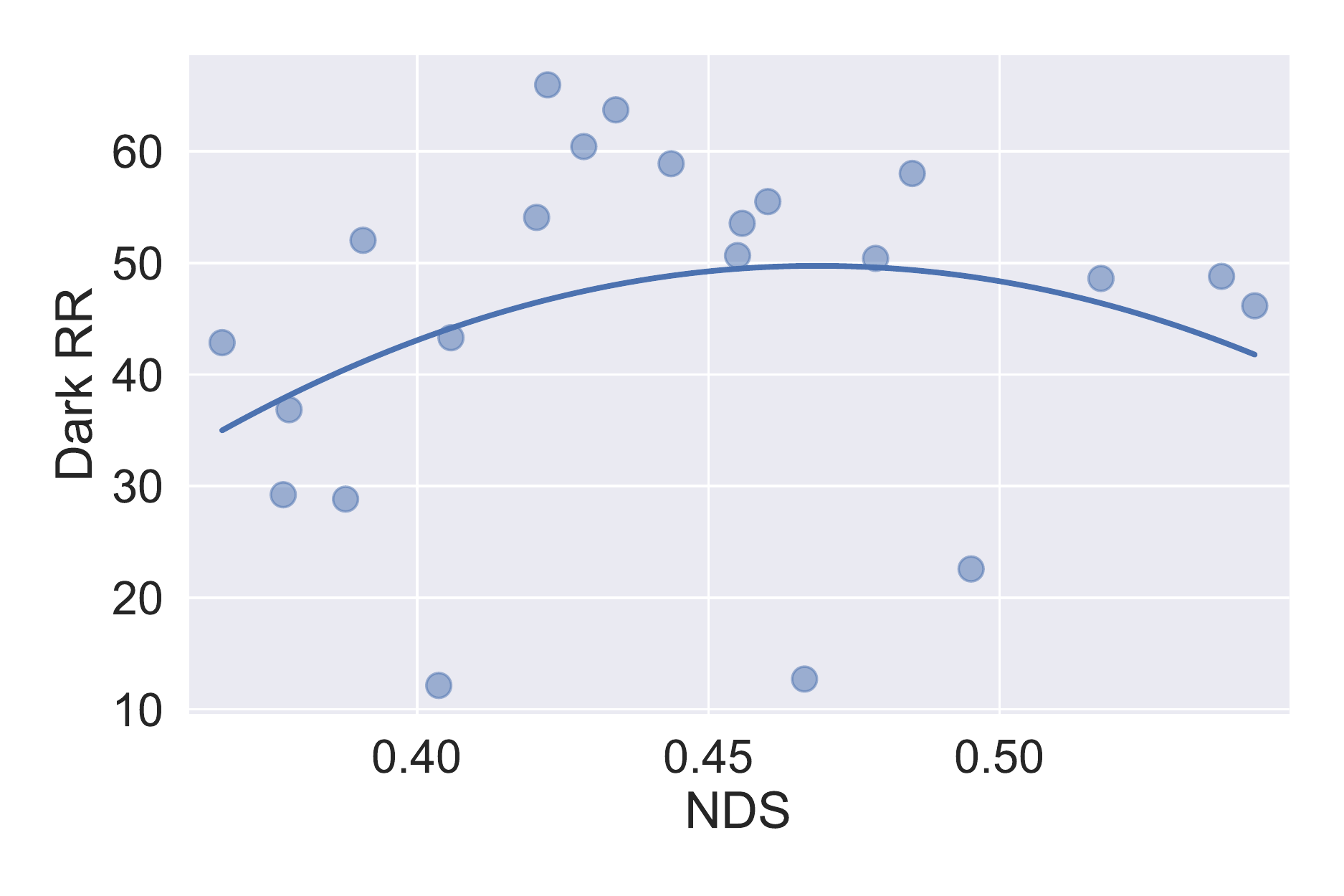}
    }
    \subfigure[Fog: CE \vs NDS.]{
        \includegraphics[width=0.23\linewidth]{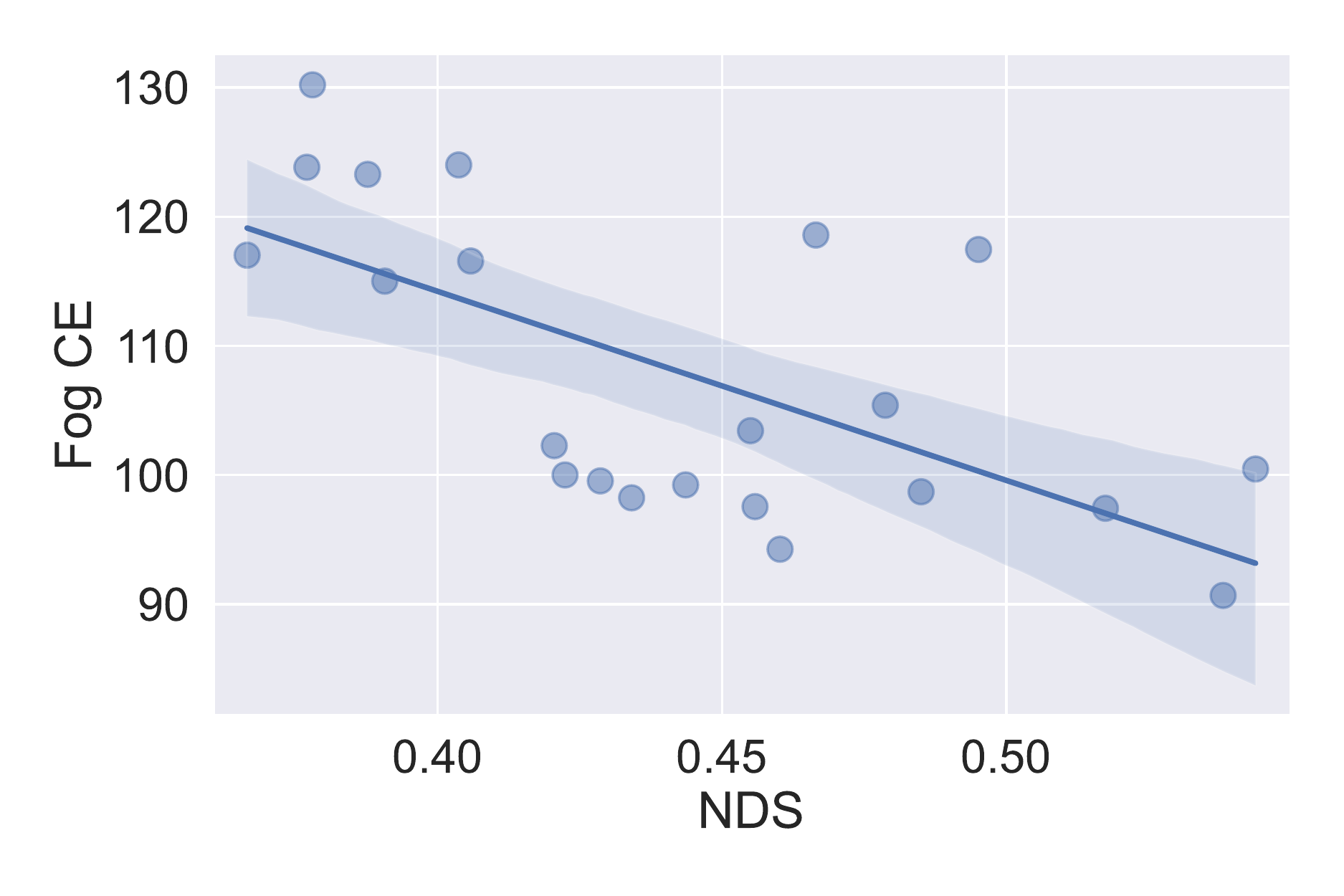}
    }
    \subfigure[Fog: RR \vs NDS.]{
        \includegraphics[width=0.23\linewidth]{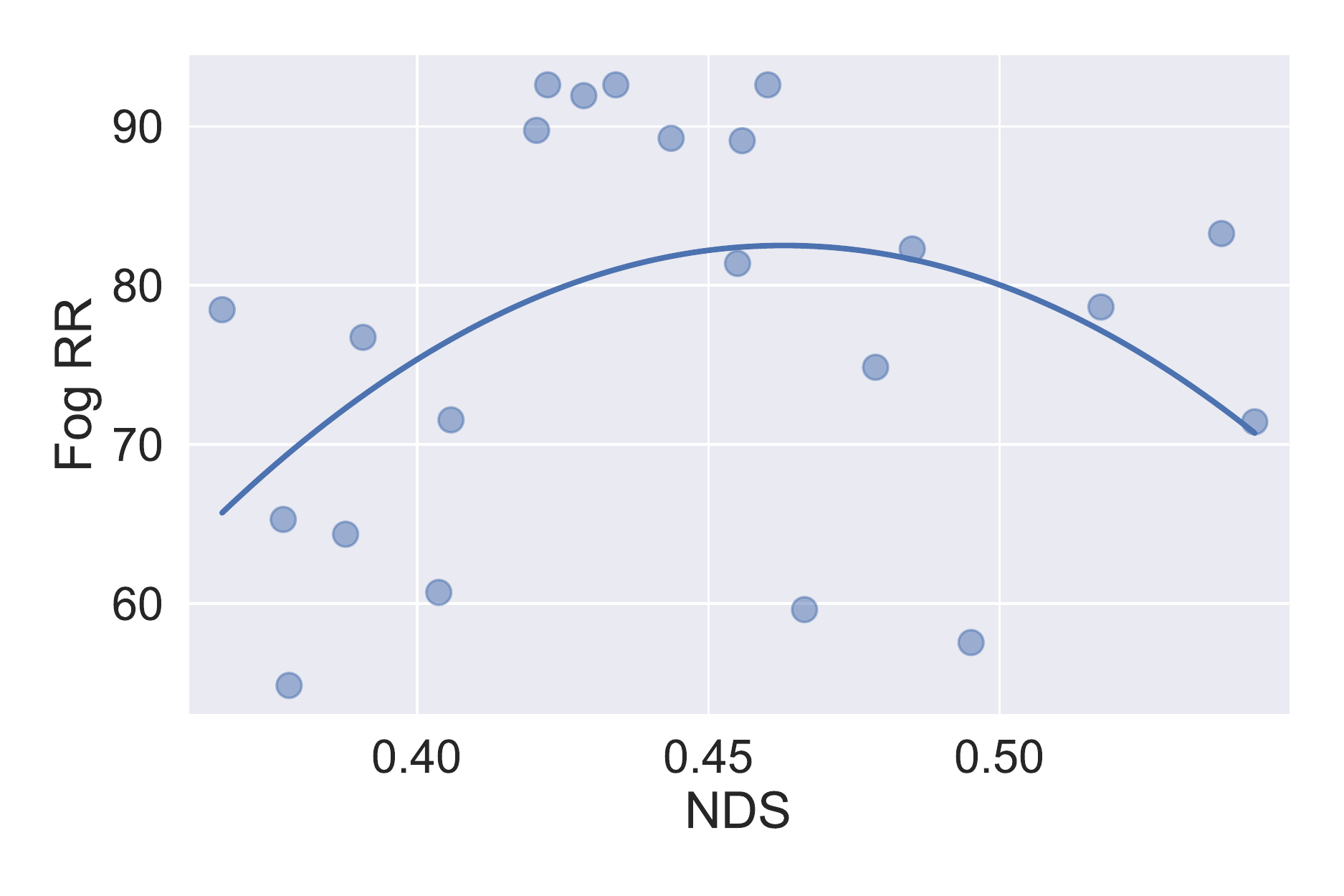}
    }
    \subfigure[Snow: CE \vs NDS.]{
        \includegraphics[width=0.23\linewidth]{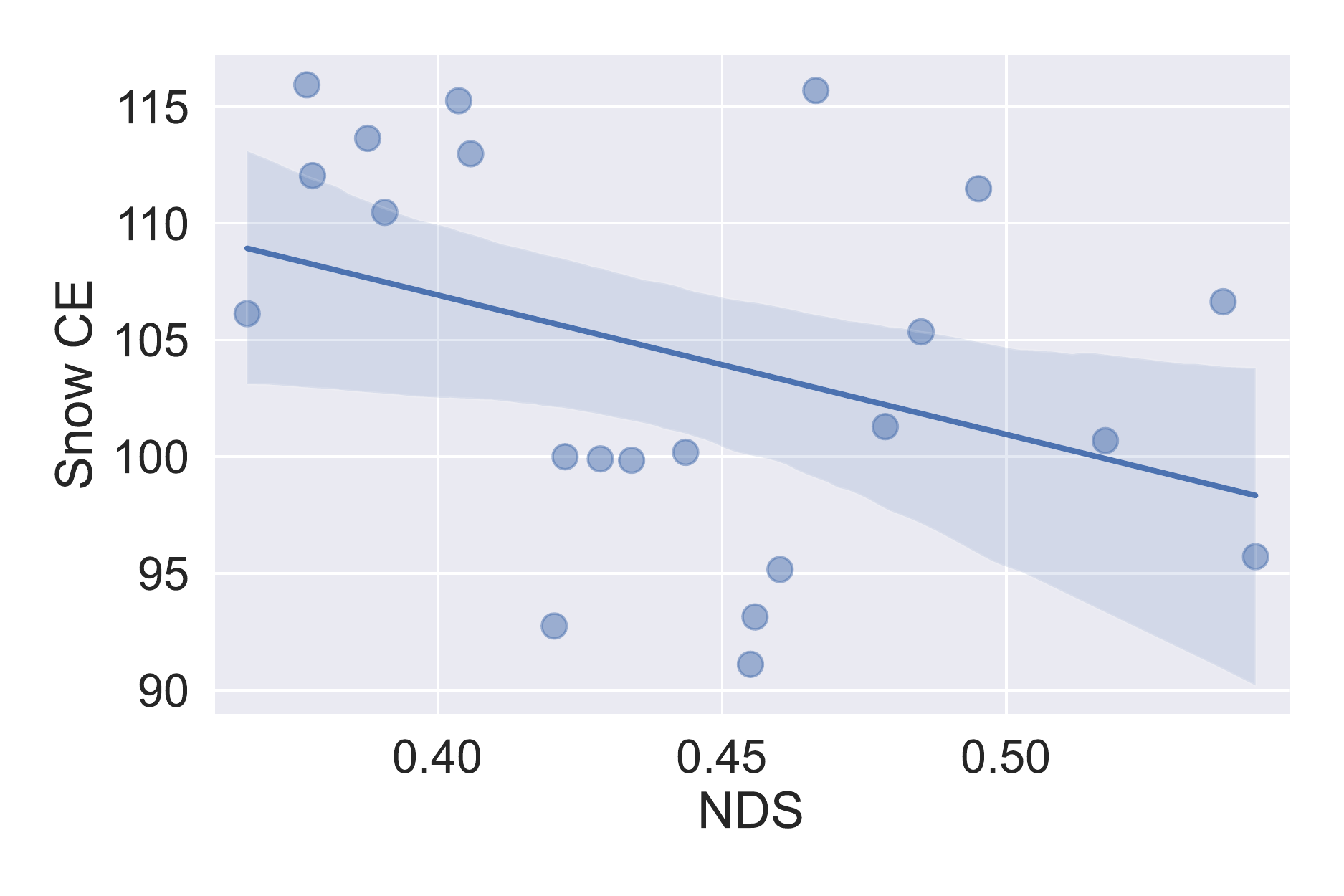}
    }
    \subfigure[Snow: RR \vs NDS.]{
        \includegraphics[width=0.23\linewidth]{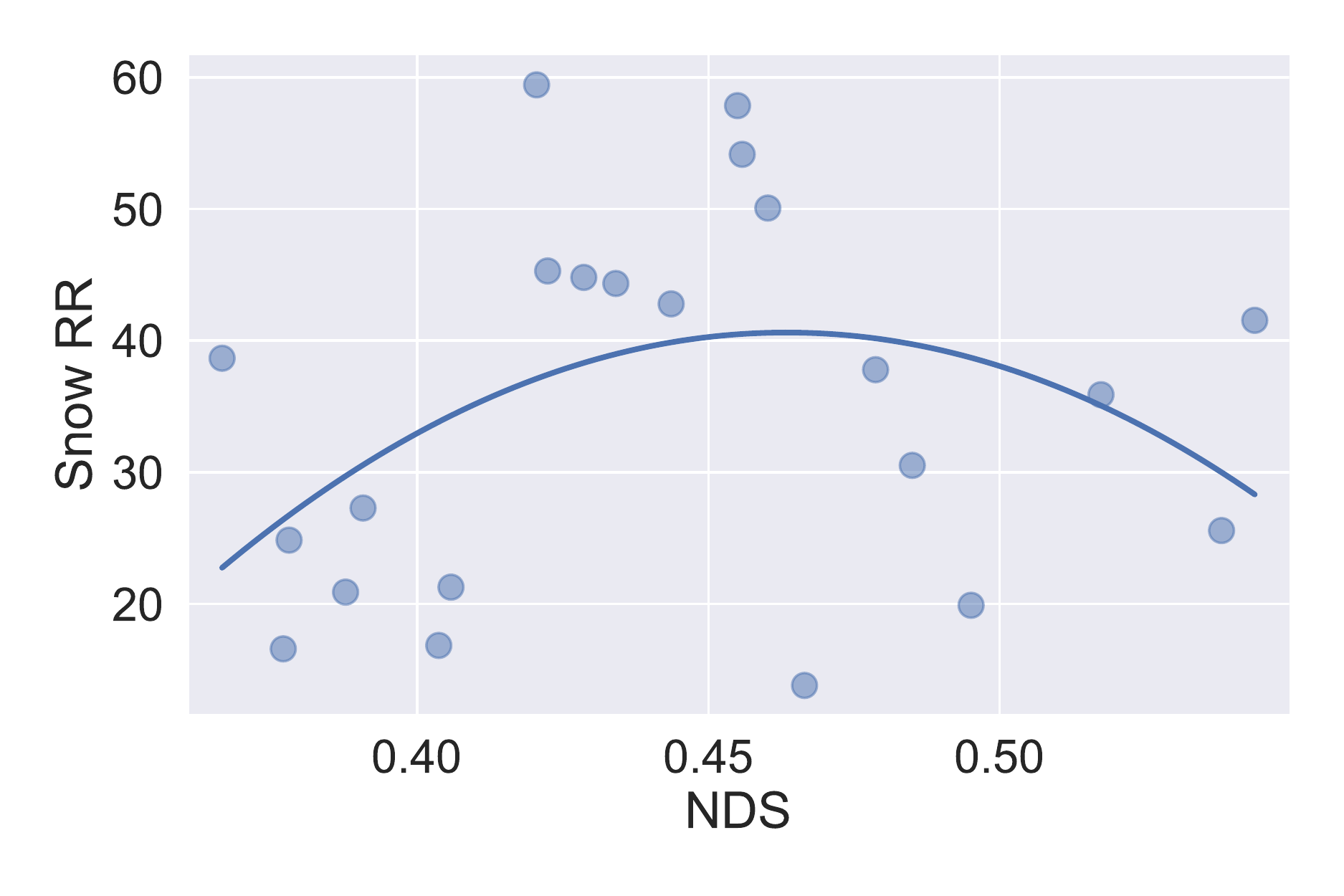}
    }
    \vspace{0.2cm}
    \caption{The corruption error (CE) and resilience rate (RR) \vs NDS under each corruption type. It is interesting to notice that the performances follow a strictly linear relationship between the standard dataset and the dataset with temporal corruptions. However, under other semantic corruptions, the robustness does not necessarily show a linear trend, which indicates models' different sensitivities to corruptions.}
    \label{fig:corruption-nds-ce-rr}
\end{figure*}

 We calculate the pixel distribution over 300 images sampled from the nuScenes dataset and visualize the pixel histograms shown in Figure~\ref{fig:hist}. Interestingly, the \textit{Motion Blur} causes the least pixel distribution shifts while causing a relatively large performance drop. On the other hand, \textit{Bright} shift the pixel distribution to higher values apparently and \textit{Fog} makes fine-grained features more indistinct by shifting the pixel value more agminated. However, these two corruptions only lead to the smallest performance gap, which reveals that model robustness is not simply correlated with pixel distribution. Additionally, we visualize corruption error (resilience rate) \vs nuScenes detection score (NDS) under every single corruption as shown in Figure~\ref{fig:corruption-nds-ce-rr}.

\begin{figure*}[ht]
    \centering
    \includegraphics[width=0.9\linewidth]{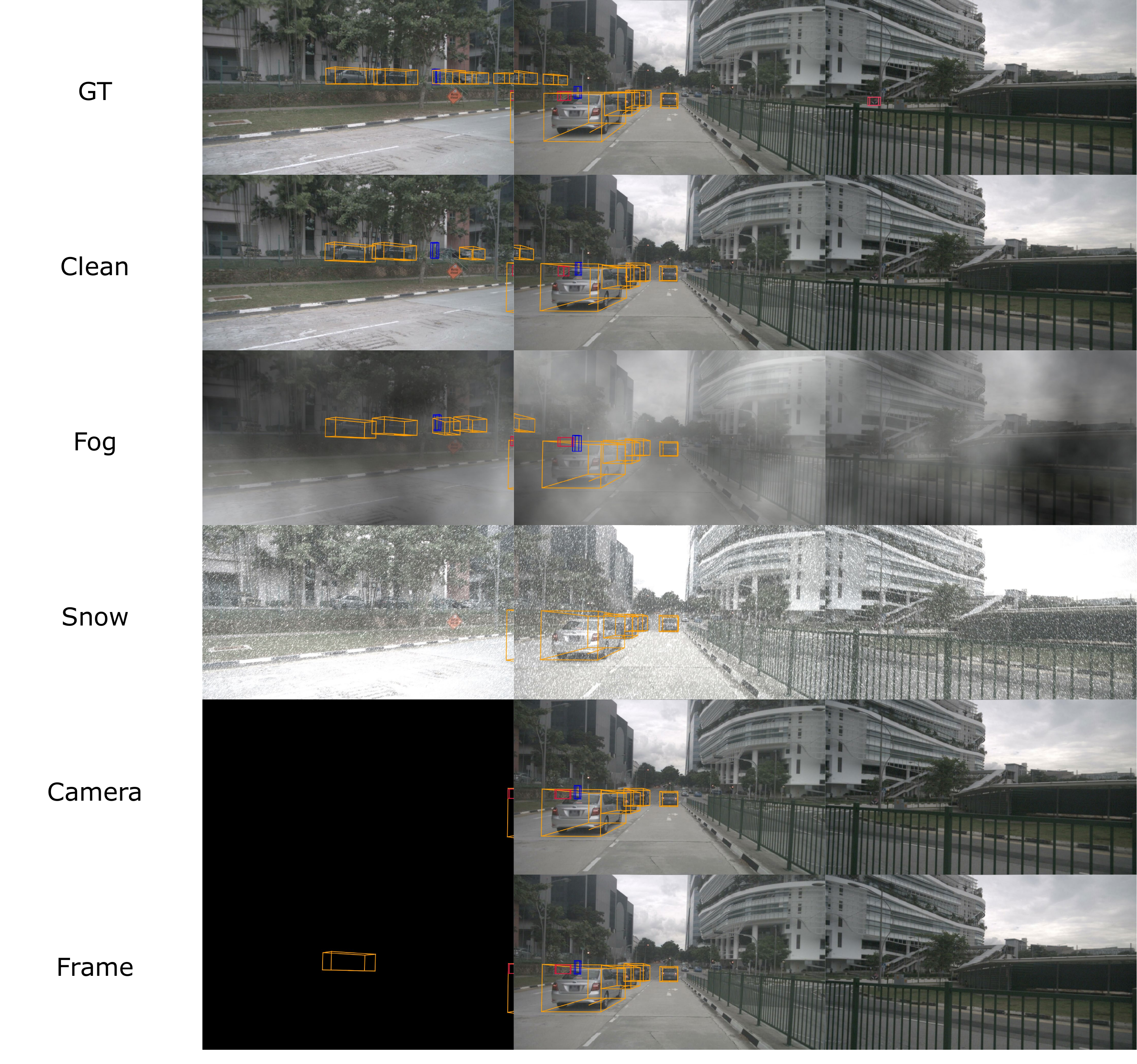}
    \caption{Visualization results of BEVFormer~\cite{li2022bevformer} under moderate corruptions.}
    \label{fig:demo-2}
\end{figure*}

\begin{figure*}[ht]
    \centering
    \includegraphics[width=\linewidth]{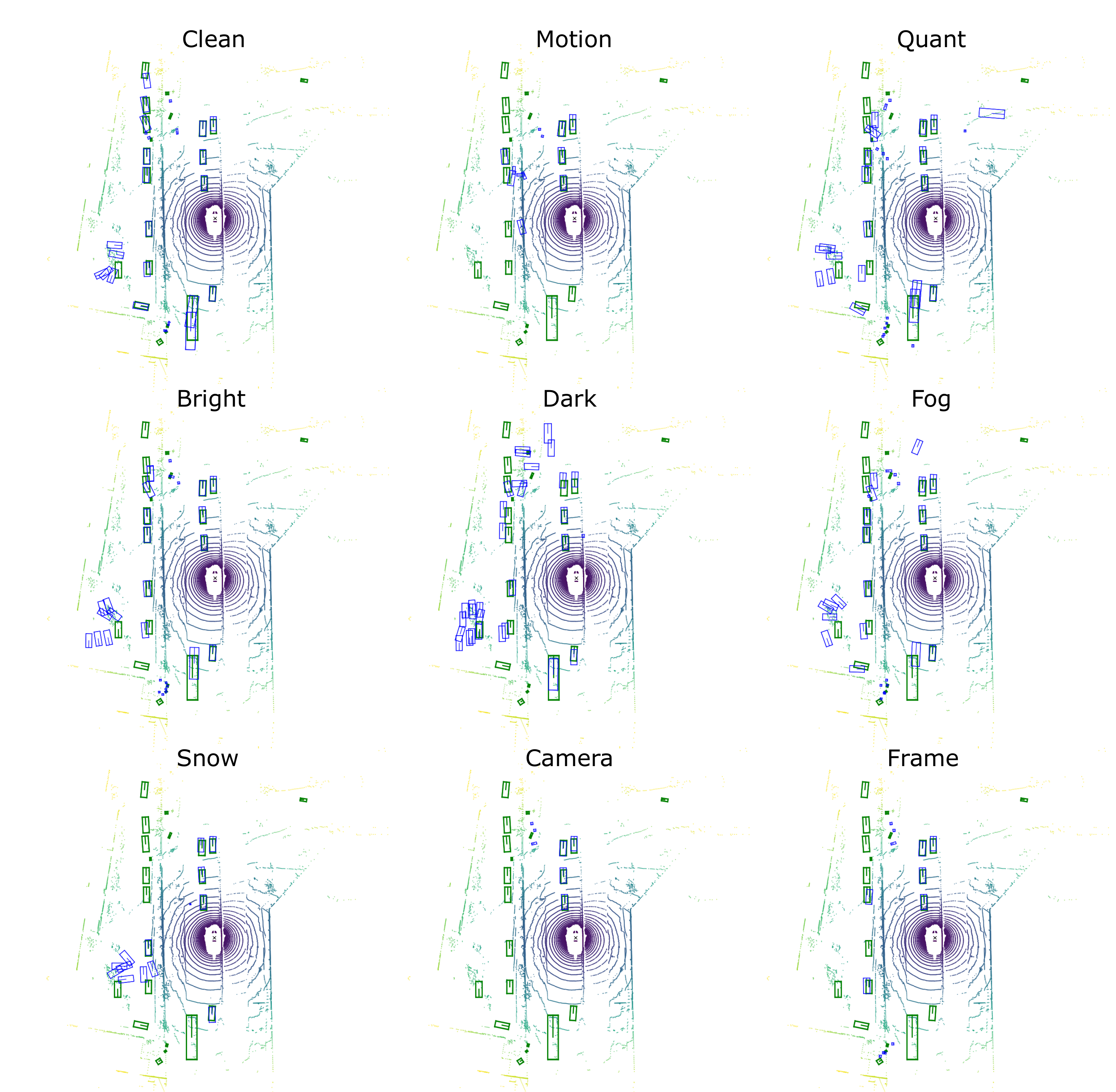}
    \vspace{0.2cm}
    \caption{The bird's eye view detection results of BEVFormer~\cite{li2022bevformer} under each corruption in the \textit{nuScenes-C} dataset. The green boxes represent ground truth and the blue boxes represent model predictions.}
    \label{fig:demo-3}
\end{figure*}

\paragraph{Temporal corruption.}
We observe a strong linear relationship between standard performance and performance on temporal corruptions as shown in Figure \ref{fig:camera-ce-nds} and \ref{fig:frame-ce-nds}. Furthermore,  We find models that utilize temporal information (\eg, BEVFormer) can output the prediction of a lost frame. However, these models still struggle to make predictions under \textit{Camera Crash}, where consecutive frames of one camera are lost. The visualization results can be seen in Figure~\ref{fig:demo-2} and Figrue~\ref{fig:demo-3}. The \textit{Camera Crash} and \textit{Frame Lost} both drop images from the front left sensor. It's interesting to see the model attempts to make predictions under \textit{Frame Lost} while failing to output any prediction under \textit{Camera Crash}. 

\paragraph{Motion and processing.}
However, in terms of other semantic corruption, models resent large variations. For example, the resilience rate of \textit{Motion Blur} and \textit{Color Quant} does not improve with the standard performance as shown in Figure \ref{fig:motion-rr-nds} and \ref{fig:quant-rr-nds}. This reveals that the relative robustness does not necessarily improve as the standard performance and approaches to increase the generalization ability to sensor distortion are still unsolved in the context of 3D perception.

\paragraph{Weather and lighting condition.}
For weather and light corruptions, the resilience rate presents more promising results, where improvement on the standard dataset can effectively bring a better resilience rate. However, this also confronts the risk of decreasing as the NDS score surpasses 0.5. Performance under \textit{Snow} corruption shows the greatest variation for models even under close standard performance. Thus, 3D perception algorithms are expected to be thoroughly examined from multiple perspectives (\eg, generalization ability to diverse environment conditions) to better understand their reliability.

\begin{figure}[ht]
    \centering
    \subfigure[Corruption Error.]{
        \label{fig:supp-backbone-ce}
        \includegraphics[width=\linewidth]{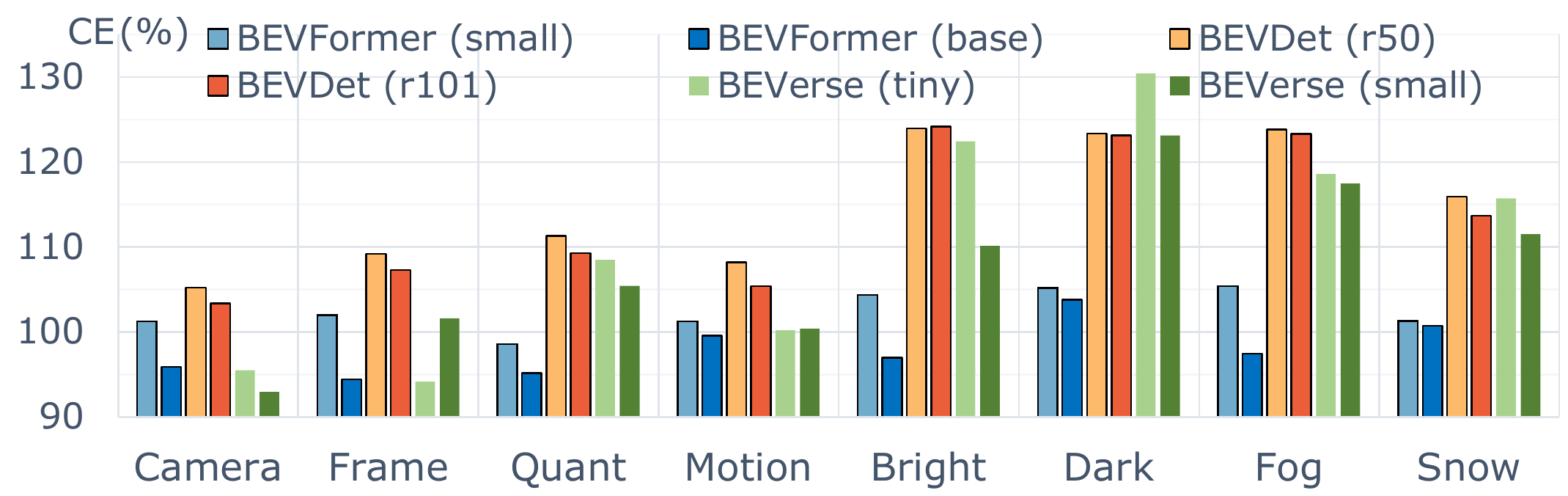}
    }
    \subfigure[Resilience Rate.]{
        \label{fig:supp-backbone-rr}
        \includegraphics[width=\linewidth]{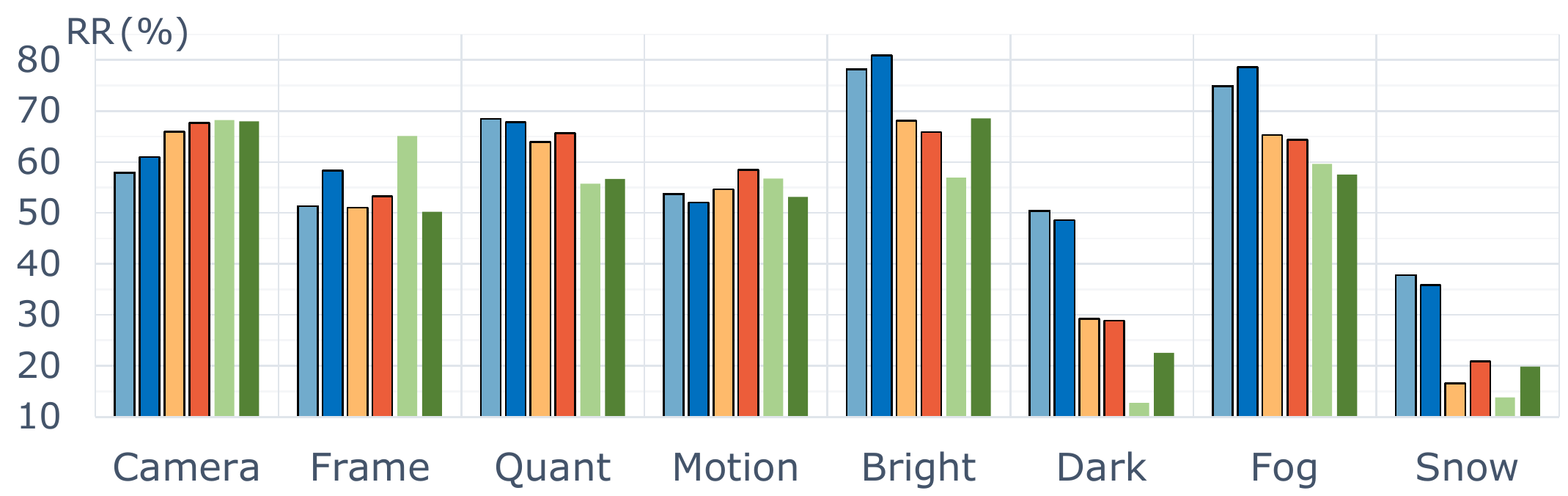}
    }
    \caption{Comparisons of different backbone sizes.}
    \label{fig:supp-backbone_size}
\end{figure}

\subsection{Backbone Size}
For most approaches, the final detection results will always improve if we enlarge the backbone network size (\eg, from ResNet50 to ResNet101). We hereby study how enlarging backbone network parameters influence model robustness. The comparison can be seen in Figure~\ref{fig:supp-backbone_size}. We find enlarging backbone size can always bring lower corruption error. However, in terms of resilience rate (RR), the benefits of larger backbone sizes are limited. The ResNet101 version BEVFormer and BEVDet only outperform their ResNet50 counterpart by a small margin. For BEVerse, the Swin-Small backbone shows better robustness towards \textit{Bright}, \textit{Dark}, and \textit{Snow}.

\section{Public Resources Used}
\label{sec:public-resources-used}

We acknowledge the use of the following public resources, during the course of this work:

\begin{itemize}
    \item nuScenes\footnote{\url{https://www.nuscenes.org/nuscenes}.} \dotfill CC BY-NC-SA 4.0
    \item nuScenes-devkit\footnote{\url{https://github.com/nutonomy/nuscenes-devkit}.} \dotfill Apache License 2.0
    \item mmdetection3d\footnote{\url{https://github.com/open-mmlab/mmdetection3d}.} \dotfill Apache License 2.0
    \item 
    imagecorruptions\footnote{\url{https://github.com/bethgelab/imagecorruptions}.} \dotfill Apache License 2.0
    \item detr3d\footnote{\url{https://github.com/WangYueFt/detr3d}.} \dotfill MIT License
    \item BEVFormer\footnote{\url{https://github.com/fundamentalvision/BEVFormer}.} \dotfill Apache License 2.0
    \item PolarFormer\footnote{\url{https://github.com/fudan-zvg/PolarFormer}.} \dotfill MIT License
    \item PETR\footnote{\url{https://github.com/megvii-research/PETR}.} \dotfill Apache License 2.0
    \item SRCN3D\footnote{\url{https://github.com/synsin0/SRCN3D}.} \dotfill MIT License
    \item BEVDet\footnote{\url{https://github.com/HuangJunJie2017/BEVDet}.} \dotfill Apache License 2.0
    \item BEVFusion\footnote{\url{https://github.com/mit-han-lab/bevfusion}.} \dotfill MIT License
    \item SOLOFusion\footnote{\url{https://github.com/Divadi/SOLOFusion}.} \dotfill Apache License 2.0
    \item Sparse4D\footnote{\url{https://github.com/linxuewu/Sparse4D}.} \dotfill Unknown
    \item BEVerse\footnote{\url{https://github.com/zhangyp15/BEVerse}.} \dotfill Unknown
    \item ora3d\footnote{\url{https://github.com/anonymous2776/ora3d}.} \dotfill Unknown

\end{itemize}

\begin{figure*}
    \centering
    \subfigure[DETR3D]{
        \includegraphics[width=0.36\linewidth]{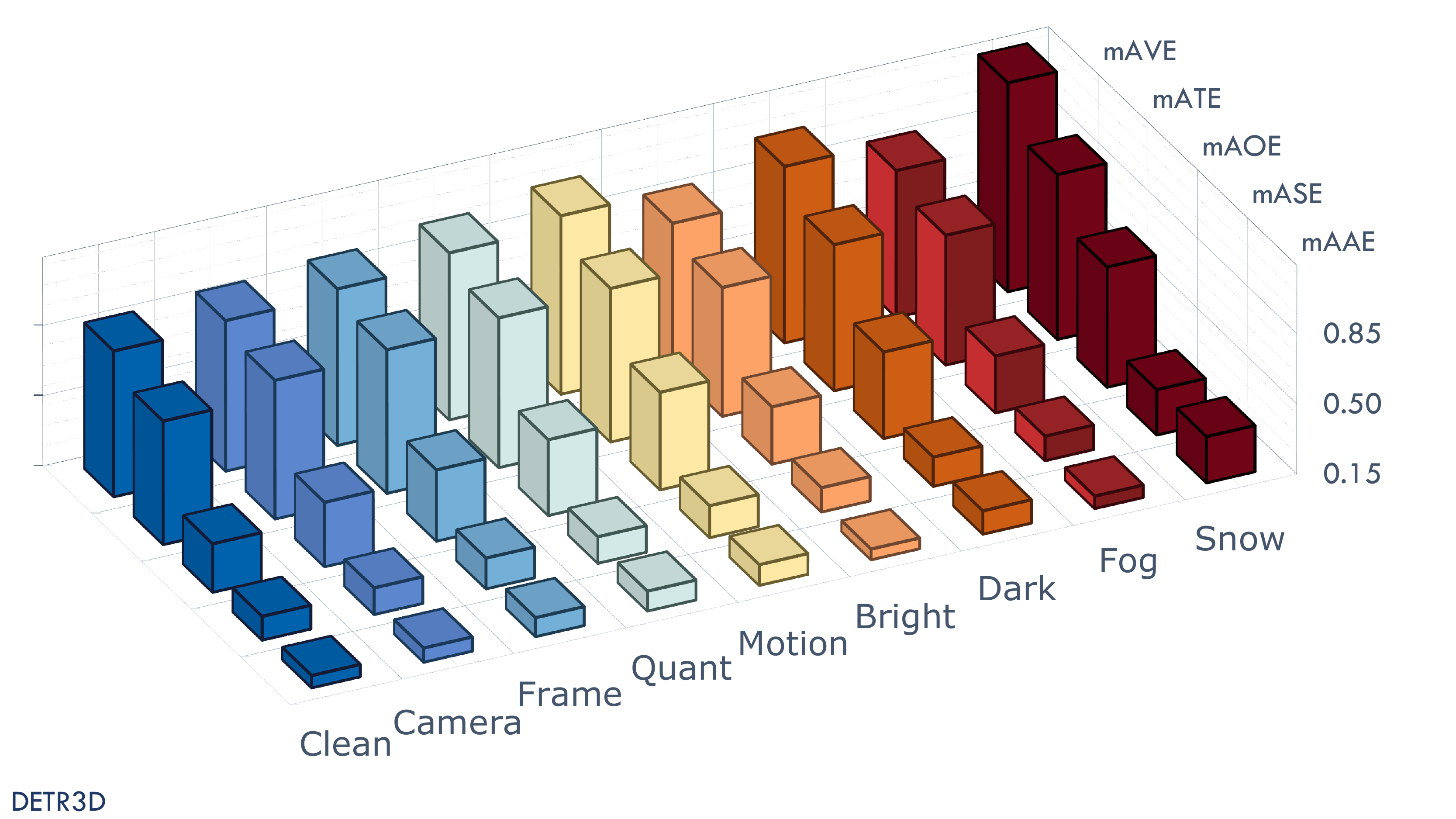}
    }
    \subfigure[BEVFormer]{
        \includegraphics[width=0.36\linewidth]{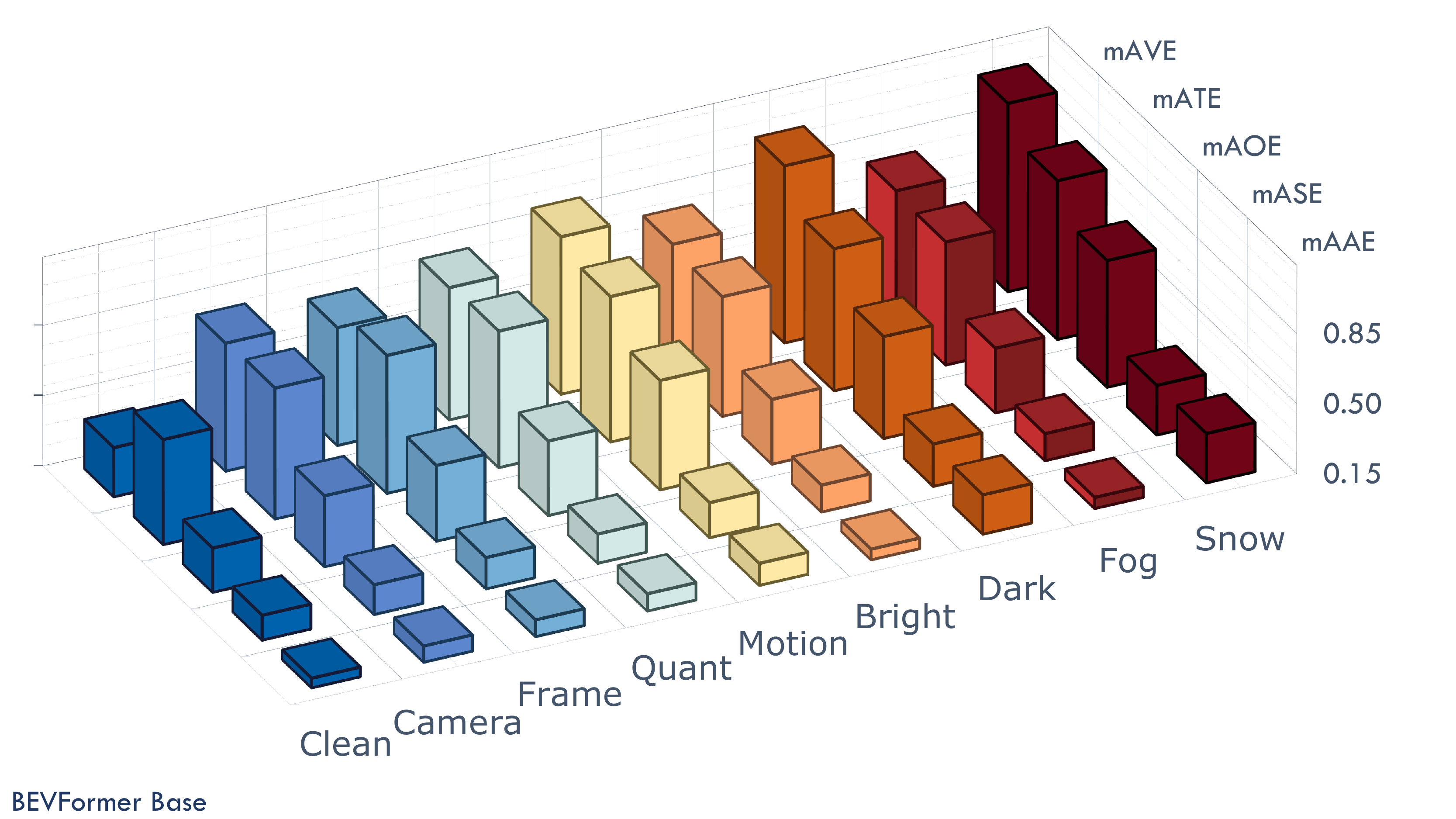}
    }
    \subfigure[PolarFormer]{
        \includegraphics[width=0.36\linewidth]{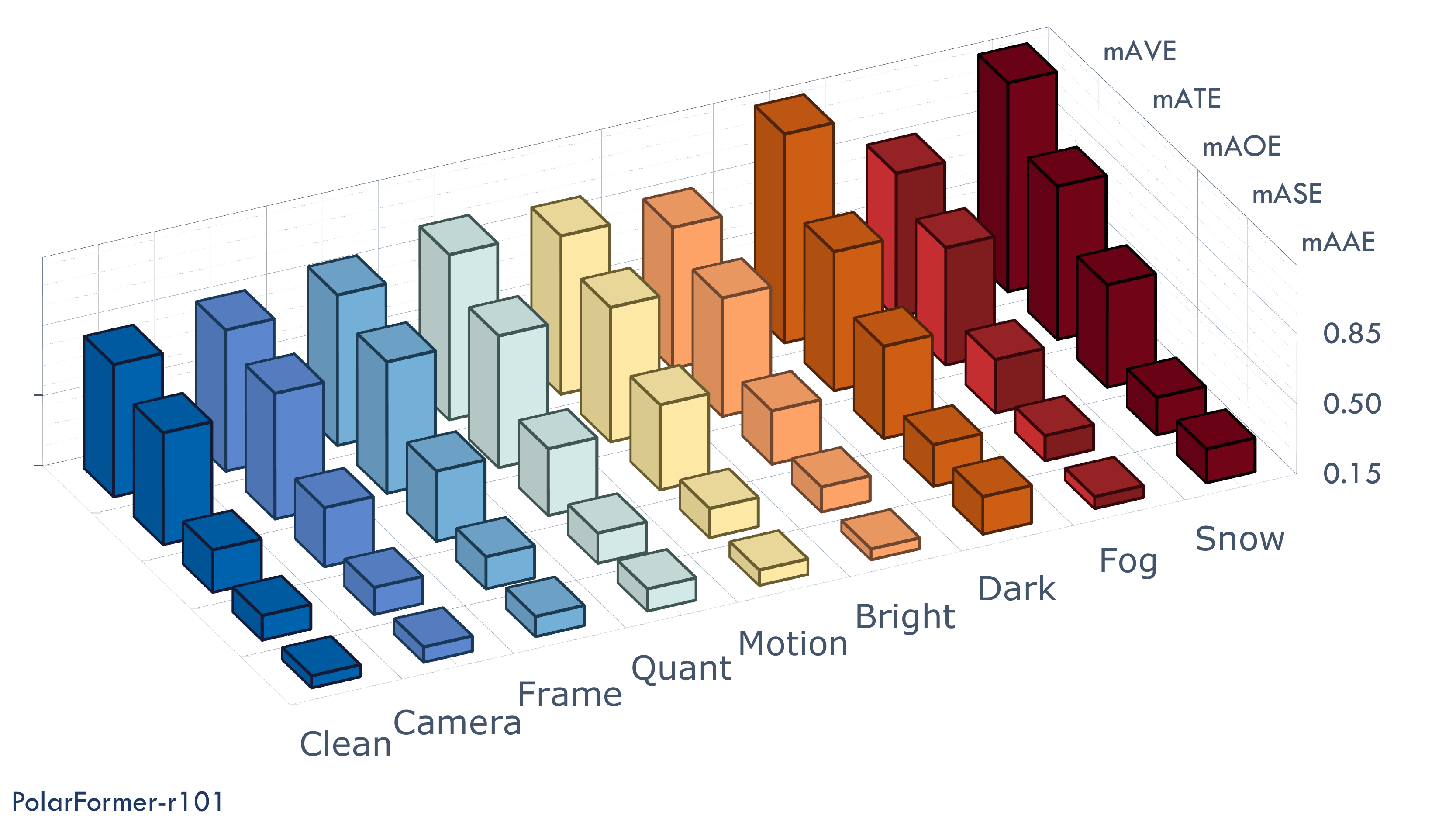}
    }
    \subfigure[PETR]{
        \includegraphics[width=0.36\linewidth]{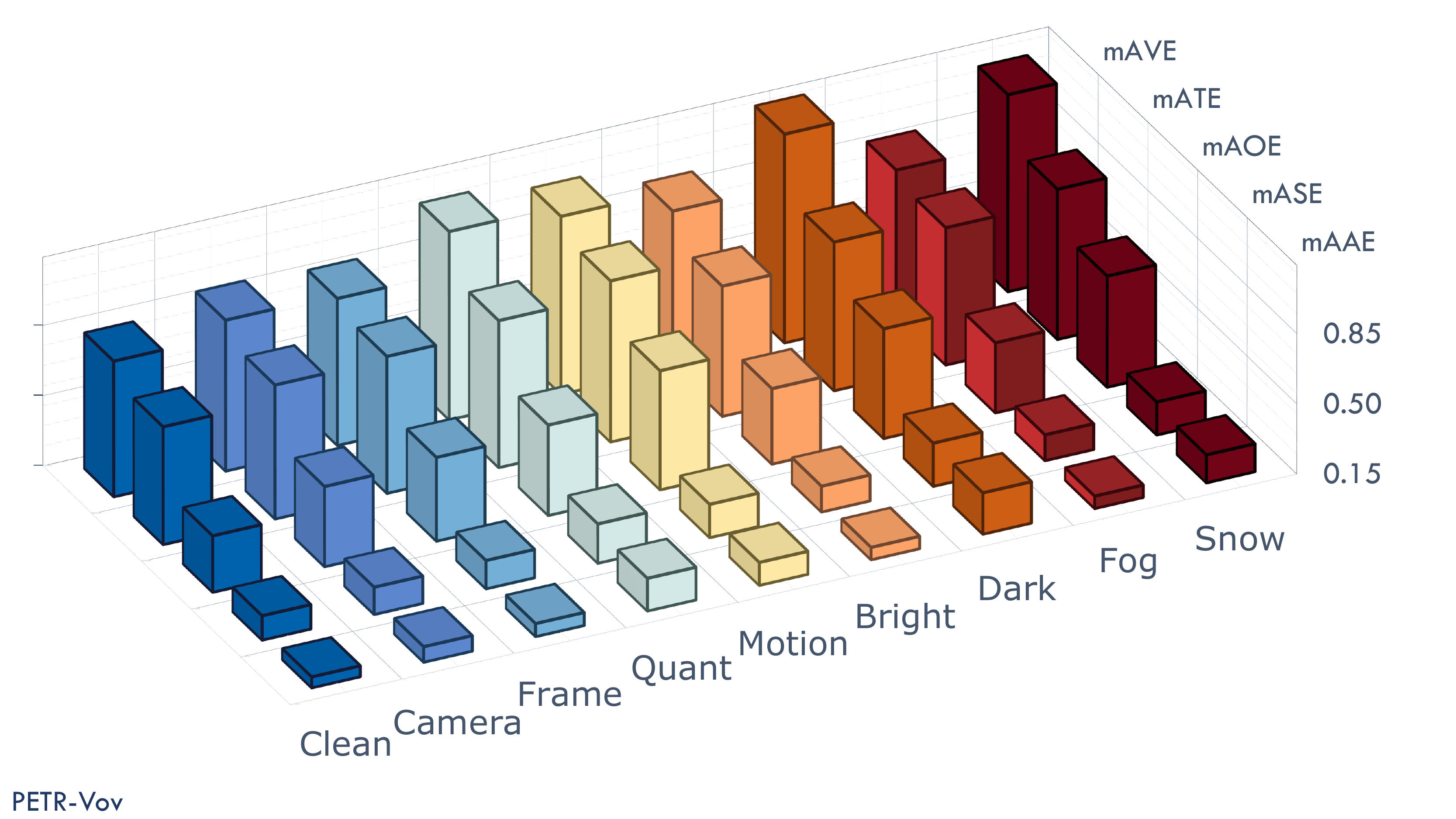}
    }
    \subfigure[ORA3D]{
        \includegraphics[width=0.36\linewidth]{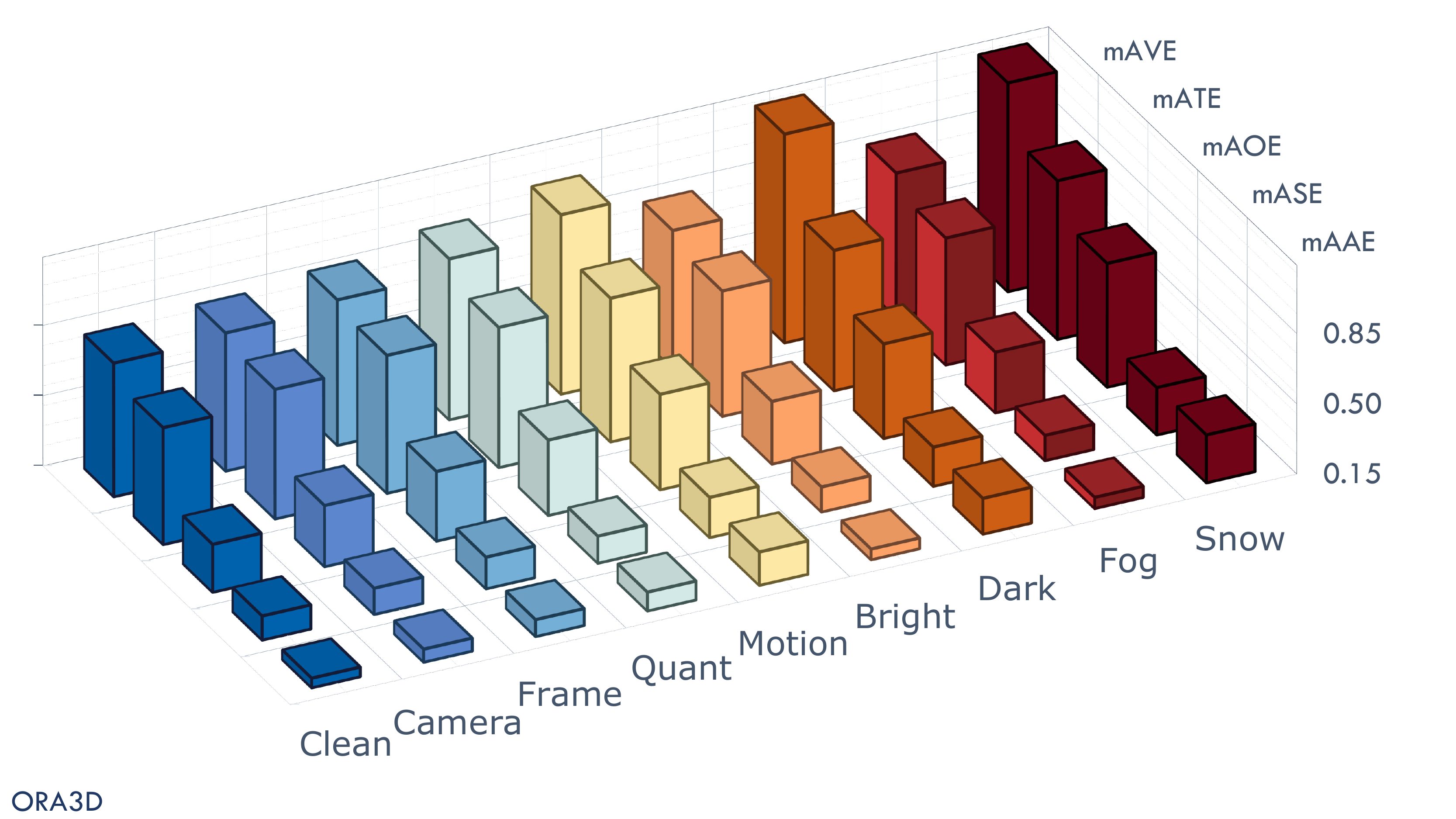}
    }
    \subfigure[Sparse4D]{
        \includegraphics[width=0.36\linewidth]{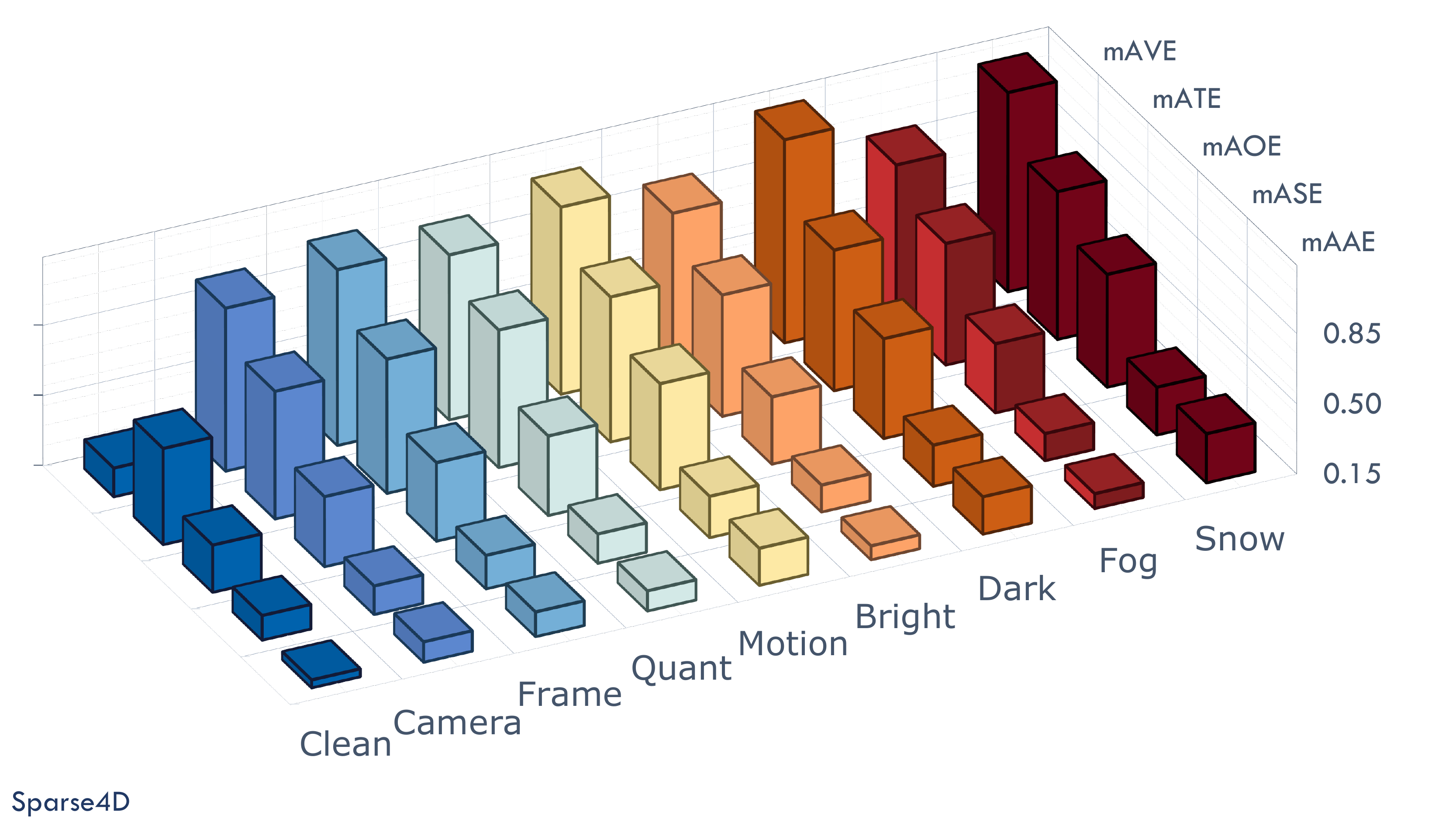}
    }
    \subfigure[SRCN3D]{
        \includegraphics[width=0.36\linewidth]{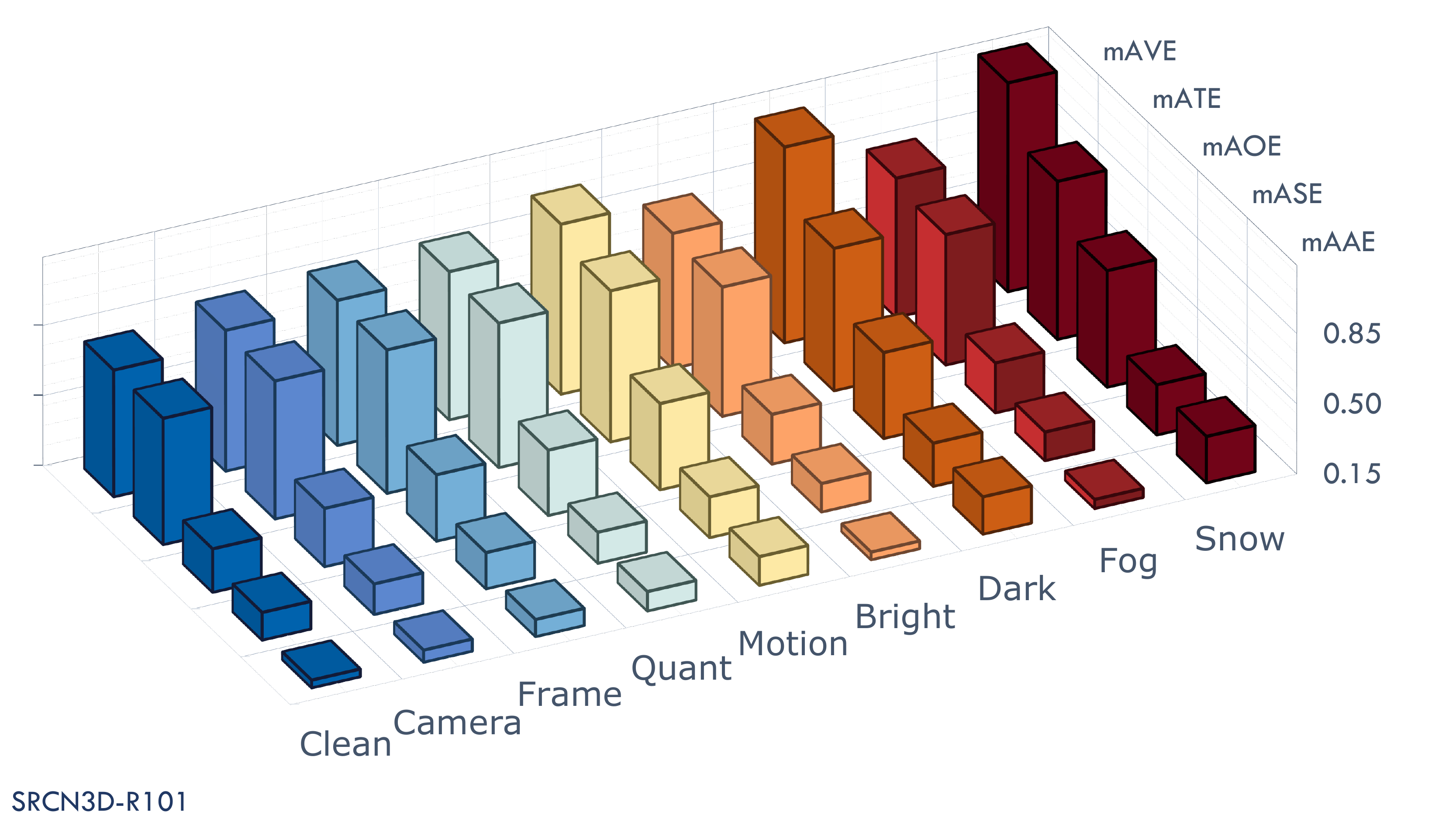}
    }
    \subfigure[BEVDet]{
        \includegraphics[width=0.36\linewidth]{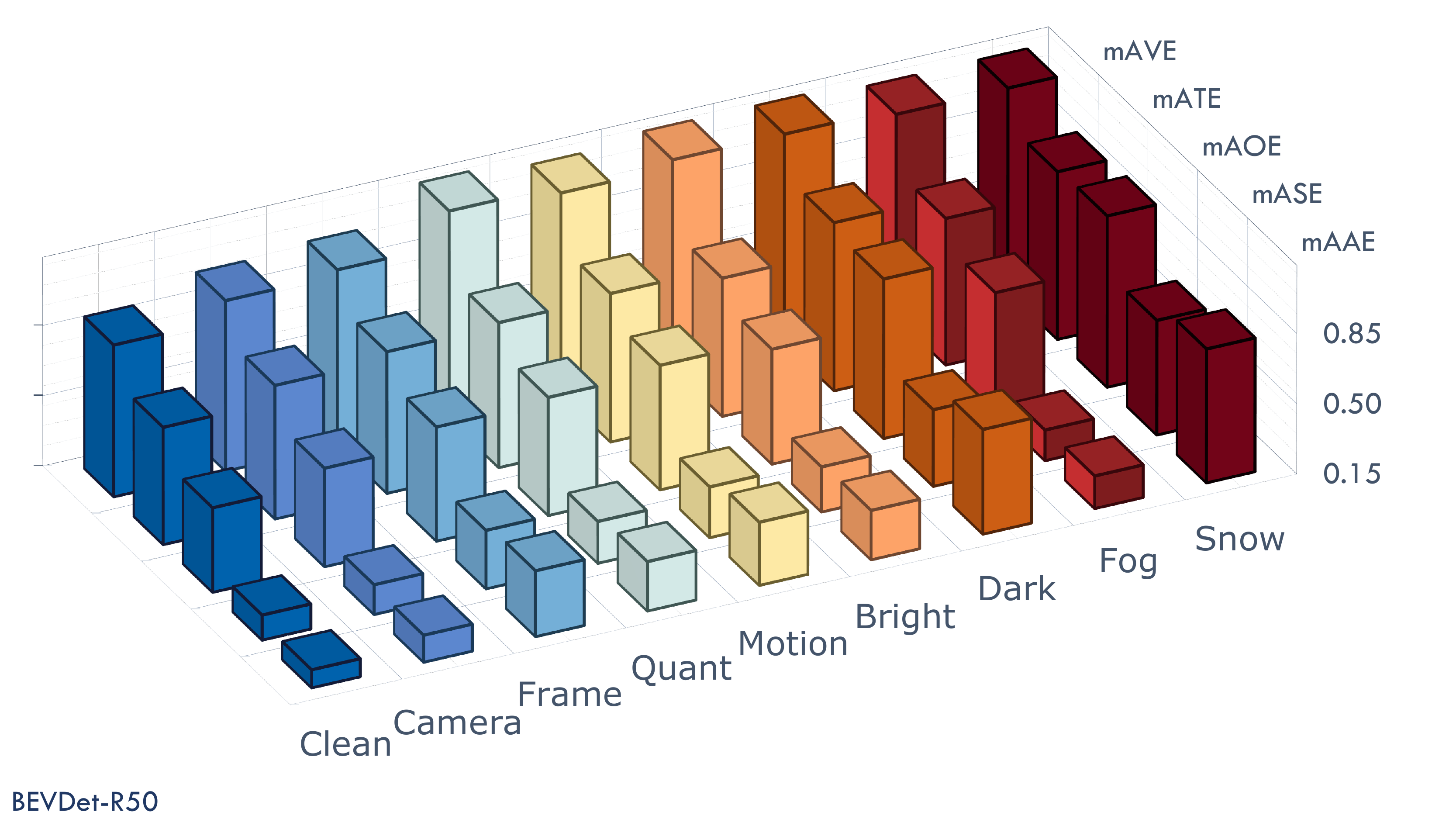}
    }
    \subfigure[BEVDepth]{
        \includegraphics[width=0.36\linewidth]{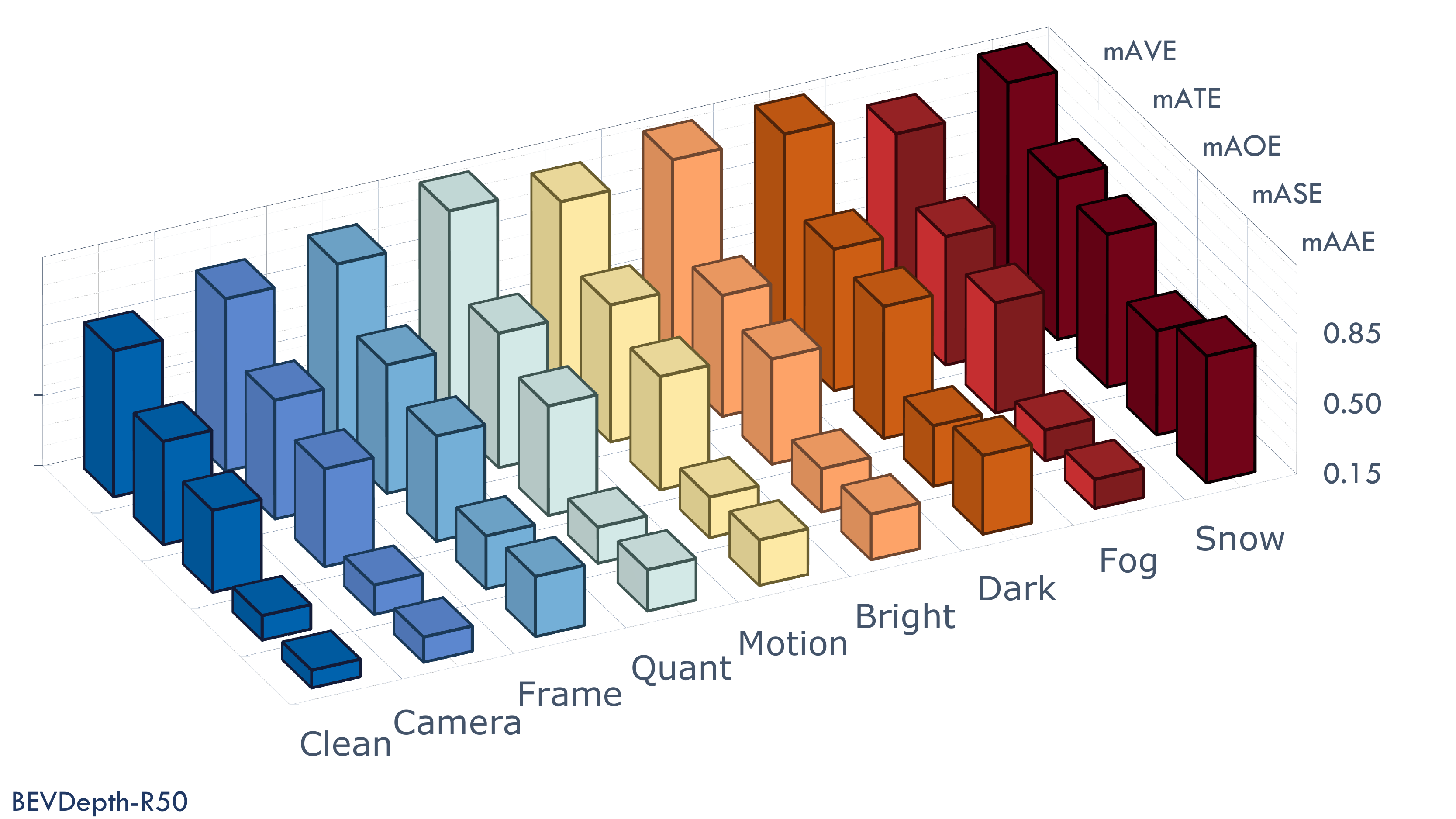}
    }
    \subfigure[BEVerse]{
        \includegraphics[width=0.36\linewidth]{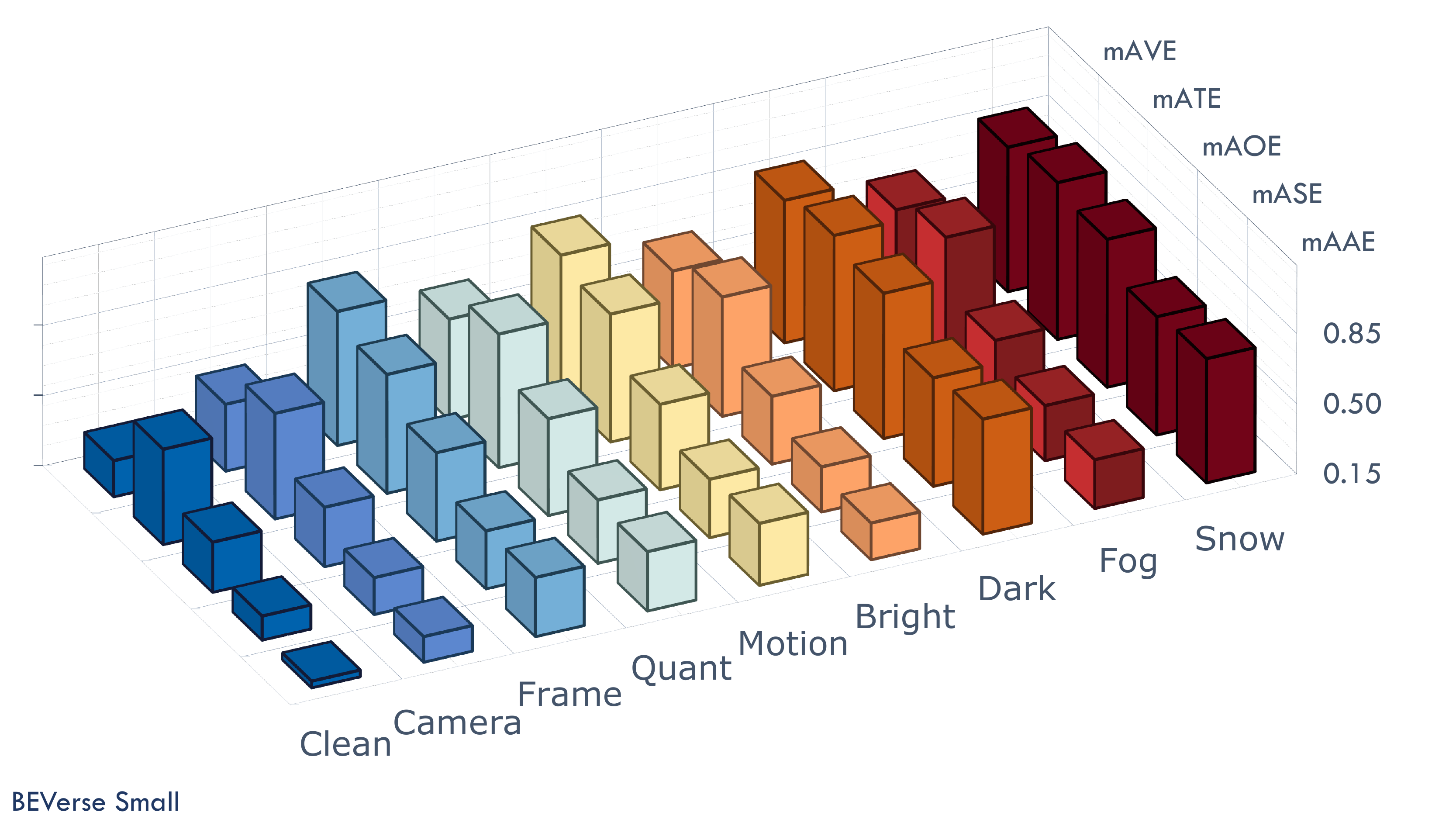}
    }
    \vspace{0.2cm}
    \caption{Benchmark results of other metric reported on \textit{nuScenes-C} other than NDS, under different corruption types.}
\end{figure*}

\clearpage

\input{tables/benchmark/detr3d}
\input{tables/benchmark/detr3d-cbgs}
\input{tables/benchmark/bevformer-small}
\input{tables/benchmark/bevformer-s-small}
\input{tables/benchmark/bevformer-base}
\input{tables/benchmark/bevformer-s-base}
\input{tables/benchmark/petr-r50}
\input{tables/benchmark/petr-vov}
\input{tables/benchmark/polarformer-r101}
\input{tables/benchmark/polarformer-vov}
\input{tables/benchmark/ora3d}
\input{tables/benchmark/bevdet-r50}
\input{tables/benchmark/bevdet-r101}
\input{tables/benchmark/bevdet-r101-fcos3d}
\input{tables/benchmark/bevdet-sttiny}
\input{tables/benchmark/bevdepth-r50}
\input{tables/benchmark/beverse-tiny}
\input{tables/benchmark/beverse-s-tiny}
\input{tables/benchmark/beverse-small}
\input{tables/benchmark/beverse-s-small}
\input{tables/benchmark/srcn3d-r101}
\input{tables/benchmark/srcn3d-vov}
\input{tables/benchmark/sparse4d}
\input{tables/benchmark/solofusion-s}
\input{tables/benchmark/solofusion-l}
\input{tables/benchmark/solofusion}
\input{tables/benchmark/fcos3d}

%% file: tables/corruption_setting.tex
\begin{table}[ht]
    \centering
    \vspace{0.1cm}
    \caption{The definition of corruption severity in \textit{RoboBEV}, with a mapping to the severity protocol in defined in ImageNet-C~\cite{hendrycks2019benchmarking}.}
    \vspace{0.2cm}
    \label{tab:corruption-setting}
    \footnotesize
    \begin{tabular}{|c|c|c|c|}
    \hline  
    \textbf{Corruption} & \textbf{Easy} & \textbf{Moderate} & \textbf{Hard} \\
    \hline  
    Motion Blur & 2 & 4 & 5 \\
    \hline  
    Color Quant & 1 & 2 & 3  \\
    \hline  
    Bright & 2 & 4 & 5  \\
    \hline  
    Dark & 2 & 3 & 4  \\
    \hline  
    Fog & 2 & 4 & 5  \\
    \hline  
    Snow & 1 & 2 & 3  \\
    \hline  
    Camera Crash & 2 & 4 & 5\\
    \hline  
    Frame Lost & 2 & 4 & 5 \\
    \hline  
    \end{tabular}
\end{table}

%% file: tables/map_seg.tex
\begin{table*}[t]
    \centering
    \caption{Map segmentation task results. The setting is the same as setting 1 mentioned in \cite{pan2020cross}.}
    \vspace{0.2cm}
    \label{tab:appe-map-seg}
    \scalebox{0.93}{
    \footnotesize
    \begin{tabular}{l|c|cccccccc}
    \toprule
    \textbf{Model} & \textbf{Clean} & \textbf{Camera} & \textbf{Frame} & \textbf{Quant} & \textbf{Motion} & \textbf{Bright} & \textbf{Dark} & \textbf{Fog} & \textbf{Snow} \\
    \midrule
    CVT\cite{pan2020cross} & $34.83$ & $20.00$ & $17.08$ & $29.42$ & $28.10$ & $27.53$ & $20.02$ & $24.72$ & $17.74$\\
    \bottomrule
    \end{tabular}
    }
\end{table*}

%% file: tables/benchmark/detr3d.tex
\begin{table*}[t]
    \centering
    \begin{tabular}{l|c|cccccc}
    \toprule
    \textbf{Corruption} & \textbf{NDS $\uparrow$} & \textbf{mAP $\uparrow$} & \textbf{mATE $\downarrow$} & \textbf{mASE $\downarrow$} & \textbf{mAOE $\downarrow$} & \textbf{mAVE $\downarrow$} & \textbf{mAAE $\downarrow$}\\
    \midrule
    Clean & $0.4224	$ & $0.3468$ & $0.7647$ & $0.2678$ & $0.3917$ & $0.8754$ & $0.2108$  \\
    Cam Crash & $0.2859$ & $0.1144$ & $0.8400$ & $0.2821$ & $0.4707$ & $0.8992$ & $0.2202$   \\
    Frame Lost & $0.2604$ & $0.0898$ & $0.8647	$ & $0.3030$ & $0.5041$ & $0.9297$ & $0.2439$   \\
    Color Quant & $0.3177$ & $0.2165$ & $0.8953	$ & $0.2816$ & $0.5266	$ & $0.9813$ & $0.2483$    \\
    Motion Blur & $0.2661$ & $0.1479$ & $0.9146$ & $0.3085$ & $0.6351$ & $1.0385$ & $0.2526$    \\
    Brightness & $0.4002$ & $0.3149$ & $0.7915$ & $0.2703$ & $0.4348$ & $0.8733$ & $0.2028$    \\
    Low Light & $0.2786$ & $0.1559$ & $0.8768$ & $0.2947$ & $0.5802$ & $1.0290$ & $0.2654$    \\
    Fog & $0.3912$ & $0.3007$ & $0.7961$ & $0.2711	$ & $0.4326$ & $0.8807$ & $0.2110$    \\
    Snow & $0.1913$ & $0.0776$ & $0.9714$ & $0.3752$ & $0.7486$ & $1.2478$ & $0.3797$    \\
    \bottomrule
    \end{tabular}
    \vspace{0.2cm}
    \caption{DETR3D~\cite{wang2022detr3d} results.}
    \label{tab:detr3d-results}
\end{table*}

%% file: tables/benchmark/detr3d-cbgs.tex
\begin{table*}[t]
    \centering
    \begin{tabular}{l|c|cccccc}
    \toprule
    \textbf{Corruption} & \textbf{NDS $\uparrow$} & \textbf{mAP $\uparrow$} & \textbf{mATE $\downarrow$} & \textbf{mASE $\downarrow$} & \textbf{mAOE $\downarrow$} & \textbf{mAVE $\downarrow$} & \textbf{mAAE $\downarrow$}\\
    \midrule
    Clean & $0.4341	$ & $0.3494$ & $0.7163$ & $0.2682$ & $0.3798$ & $0.8421$ & $0.1997$  \\
    Cam Crash & $0.2991$ & $0.1174$ & $0.7932$ & $0.2853	$ & $0.4575$ & $0.8471$ & $0.2131$   \\
    Frame Lost & $0.2685$ & $0.0923$ & $0.8268$ & $0.3135$ & $0.5042$ & $0.8867$ & $0.2455$   \\
    Color Quant & $0.3235$ & $0.2152$ & $0.8571	$ & $0.2875$ & $0.5350	$ & $0.9354$ & $0.2400$    \\
    Motion Blur & $0.2542$ & $0.1385$ & $0.8909$ & $0.3355$ & $0.6707$ & $1.0682$ & $0.2928$    \\
    Brightness & $0.4154$ & $0.3200$ & $0.7357$ & $0.2720$ & $0.4086$ & $0.8302	$ & $0.1990$    \\
    Low Light & $0.2766$ & $0.1539$ & $0.8419$ & $0.3262$ & $0.5682$ & $1.09520$ & $0.2847$    \\
    Fog & $0.4020$ & $0.3012$ & $0.7552$ & $0.2710	$ & $0.4237$ & $0.8302$ & $0.2054$    \\
    Snow & $0.1925$ & $0.0702$ & $0.9246$ & $0.3793$ & $0.7648$ & $1.2585$ & $0.3577$    \\
    \bottomrule
    \end{tabular}
    \vspace{0.2cm}
    \caption{DETR3D$_{\text{CBGS}}$~\cite{wang2022detr3d} results.}
    \label{tab:detr3d-cbgs-results}
\end{table*}

%% file: tables/benchmark/bevformer-small.tex
\begin{table*}[t]
    \centering
    \begin{tabular}{l|c|cccccc}
    \toprule
    \textbf{Corruption} & \textbf{NDS $\uparrow$} & \textbf{mAP $\uparrow$} & \textbf{mATE $\downarrow$} & \textbf{mASE $\downarrow$} & \textbf{mAOE $\downarrow$} & \textbf{mAVE $\downarrow$} & \textbf{mAAE $\downarrow$}\\
    \midrule
    Clean & $0.4787	$ & $0.3700$ & $0.7212$ & $0.2792$ & $0.4065$ & $0.4364$ & $0.2201$  \\
    Cam Crash & $0.2771$ & $0.1130$ & $0.8627$ & $0.3099	$ & $0.5398$ & $0.8376$ & $0.2446$   \\
    Frame Lost & $0.2459$ & $0.0933$ & $0.8959$ & $0.3411$ & $0.5742$ & $0.9154$ & $0.2804$   \\
    Color Quant & $0.3275$ & $0.2109$ & $0.8476	$ & $0.2943$ & $0.5234$ & $0.8539$ & $0.2601$    \\
    Motion Blur & $0.2570$ & $0.1344$ & $0.8995$ & $0.3264$ & $0.6774$ & $0.9625$ & $0.2605$    \\
    Brightness & $0.3741$ & $0.2697$ & $0.8064$ & $0.2830$ & $0.4796$ & $0.8162	$ & $0.2226$    \\
    Low Light & $0.2413$ & $0.1191$ & $0.8838$ & $0.3598$ & $0.6470$ & $1.0391$ & $0.3323$    \\
    Fog & $0.3583$ & $0.2486$ & $0.8131$ & $0.2862	$ & $0.5056$ & $0.8301	$ & $0.2251$    \\
    Snow & $0.1809$ & $0.0635$ & $0.9630$ & $0.3855	$ & $0.7741$ & $1.1002$ & $0.3863$    \\
    \bottomrule
    \end{tabular}
    \vspace{0.2cm}
    \caption{BEVFormer  (small)~\cite{li2022bevformer} results.}
    \label{tab:bevformer-small}
\end{table*}

%% file: tables/benchmark/bevformer-s-small.tex
\begin{table*}[t]
    \centering
    \begin{tabular}{l|c|cccccc}
    \toprule
    \textbf{Corruption} & \textbf{NDS $\uparrow$} & \textbf{mAP $\uparrow$} & \textbf{mATE $\downarrow$} & \textbf{mASE $\downarrow$} & \textbf{mAOE $\downarrow$} & \textbf{mAVE $\downarrow$} & \textbf{mAAE $\downarrow$}\\
    \midrule
     Clean  & $0.2622$  & $0.1324$  & $0.9352$  & $0.3024$  & $0.5556$  & $1.1106$  & $0.2466$  \\ 
     Camera Crash  & $0.2013$  & $0.0425$  & $0.9844$  & $0.3306$  & $0.6330$  & $1.0969$  & $0.2556$  \\
     Frame Lost  & $0.1638$  & $0.0292$  & $1.0051$  & $0.4294$  & $0.6963$  & $1.1418$  & $0.3954$  \\
     Color Quant  & $0.2313$  & $0.1041$  & $0.9625$  & $0.3131$  & $0.6435$  & $1.1686$  & $0.2882$  \\
     Motion Blur  & $0.1916$  & $0.0676$  & $0.9741$  & $0.3644$  & $0.7525$  & $1.3062$  & $0.3307$  \\
     Brightness  & $0.2520$  & $0.1250$  & $0.9484$  & $0.3034$  & $0.6046$  & $1.1318$  & $0.2486$  \\
     Low Light  & $0.1868$  & $0.0624$  & $0.9414$  & $0.3984$  & $0.7185$  & $1.3064$  & $0.3859$  \\
     Fog  & $0.2442$  & $0.1181$  & $0.9498$  & $0.3055$  & $0.6343$  & $1.1806$  & $0.2592$  \\
     Snow  & $0.1414$  & $0.0294$  & $1.0231$  & $0.4242$  & $0.8644$  & $1.3622$  & $0.4444$  \\
    \bottomrule
    \end{tabular}
    \vspace{0.2cm}
    \caption{BEVFormer-S (small)~\cite{li2022bevformer} results.}
    \label{tab:bevformer-s-small}
\end{table*}

%% file: tables/benchmark/bevformer-base.tex
\begin{table*}[t]
    \centering
    \begin{tabular}{l|c|cccccc}
    \toprule
    \textbf{Corruption} & \textbf{NDS $\uparrow$} & \textbf{mAP $\uparrow$} & \textbf{mATE $\downarrow$} & \textbf{mASE $\downarrow$} & \textbf{mAOE $\downarrow$} & \textbf{mAVE $\downarrow$} & \textbf{mAAE $\downarrow$}\\
    \midrule
     Clean  & $0.5174$  & $0.4164$  & $0.6726$  & $0.2734$  & $0.3704$  & $0.3941$  & $0.1974$  \\ 
     Cam Crash  & $0.3154$  & $0.1545$  & $0.8015$  & $0.2975$  & $0.5031$  & $0.7865$  & $0.2301$  \\
     Frame Lost  & $0.3017$  & $0.1307$  & $0.8359$  & $0.3053$  & $0.5262$  & $0.7364$  & $0.2328$  \\
     Color Quant  & $0.3509$  & $0.2393$  & $0.8294$  & $0.2953$  & $0.5200$  & $0.8079$  & $0.2350$  \\
     Motion Blur  & $0.2695$  & $0.1531$  & $0.8739$  & $0.3236$  & $0.6941$  & $0.9334$  & $0.2592$  \\
     Brightness  & $0.4184$  & $0.3312$  & $0.7457$  & $0.2832$  & $0.4721$  & $0.7686$  & $0.2024$  \\
     Low Light  & $0.2515$  & $0.1394$  & $0.8568$  & $0.3601$  & $0.6571$  & $1.0322$  & $0.3453$  \\
     Fog  & $0.4069$  & $0.3141$  & $0.7627$  & $0.2837$  & $0.4711$  & $0.7798$  & $0.2046$  \\
     Snow  & $0.1857$  & $0.0739$  & $0.9405$  & $0.3966$  & $0.7806$  & $1.0880$  & $0.3951$  \\
    \bottomrule
    \end{tabular}
    \vspace{0.2cm}
    \caption{BEVFormer (base)~\cite{li2022bevformer} results.}
    \label{tab:bevformer-base}
\end{table*}

%% file: tables/benchmark/bevformer-s-base.tex
\begin{table*}[t]
    \centering
    \begin{tabular}{l|c|cccccc}
    \toprule
    \textbf{Corruption} & \textbf{NDS $\uparrow$} & \textbf{mAP $\uparrow$} & \textbf{mATE $\downarrow$} & \textbf{mASE $\downarrow$} & \textbf{mAOE $\downarrow$} & \textbf{mAVE $\downarrow$} & \textbf{mAAE $\downarrow$}\\
    \midrule
     Clean  & $0.4129$  & $0.3461$  & $0.7549$  & $0.2832$  & $0.4520$  & $0.8917$  & $0.2194$  \\ 
     Cam Crash  & $0.2879$  & $0.1240$  & $0.8041$  & $0.2966$  & $0.5094$  & $0.8986$  & $0.2323$  \\
     Frame Lost  & $0.2642$  & $0.0969$  & $0.8352$  & $0.3093$  & $0.5748$  & $0.8861$  & $0.2374$  \\
     Color Quant  & $0.3207$  & $0.2243$  & $0.8488$  & $0.2992$  & $0.5422$  & $1.0003$  & $0.2522$  \\
     Motion Blur  & $0.2518$  & $0.1434$  & $0.8845$  & $0.3248$  & $0.7179$  & $1.1211$  & $0.2860$  \\
     Brightness  & $0.3819$  & $0.3093$  & $0.7761$  & $0.2861$  & $0.4999$  & $0.9466$  & $0.2201$  \\
     Low Light  & $0.2381$  & $0.1316$  & $0.8640$  & $0.3602$  & $0.6903$  & $1.2132$  & $0.3622$  \\
     Fog  & $0.3662$  & $0.2907$  & $0.7938$  & $0.2870$  & $0.5162$  & $0.9702$  & $0.2254$  \\
     Snow  & $0.1793$  & $0.0687$  & $0.9472$  & $0.3954$  & $0.8004$  & $1.2524$  & $0.4078$  \\
    \bottomrule
    \end{tabular}
    \vspace{0.2cm}
    \caption{BEVFormer-S (base)~\cite{li2022bevformer} results.}
    \label{tab:bevformer-s-base}
\end{table*}

%% file: tables/benchmark/petr-r50.tex
\begin{table*}[t]
    \centering
    \begin{tabular}{l|c|cccccc}
    \toprule
    \textbf{Corruption} & \textbf{NDS $\uparrow$} & \textbf{mAP $\uparrow$} & \textbf{mATE $\downarrow$} & \textbf{mASE $\downarrow$} & \textbf{mAOE $\downarrow$} & \textbf{mAVE $\downarrow$} & \textbf{mAAE $\downarrow$}\\
    \midrule
     Clean  & $0.3665$  & $0.3174$  & $0.8397$  & $0.2796$  & $0.6158$  & $0.9543$  & $0.2326$  \\ 
     Cam Crash  & $0.2320$  & $0.1065$  & $0.9383$  & $0.2975$  & $0.7220$  & $1.0169$  & $0.2585$  \\
     Frame Lost  & $0.2166$  & $0.0868$  & $0.9513$  & $0.3041$  & $0.7597$  & $1.0081$  & $0.2629$  \\
     Color Quant  & $0.2472$  & $0.1734$  & $0.9121$  & $0.3616$  & $0.7807$  & $1.1634$  & $0.3473$  \\
     Motion Blur  & $0.2299$  & $0.1378$  & $0.9587$  & $0.3164$  & $0.8461$  & $1.1190$  & $0.2847$  \\
     Brightness  & $0.2841$  & $0.2101$  & $0.9049$  & $0.3080$  & $0.7429$  & $1.0838$  & $0.2552$  \\
     Low Light  & $0.1571$  & $0.0685$  & $0.9465$  & $0.4222$  & $0.9201$  & $1.4371$  & $0.4971$  \\
     Fog  & $0.2876$  & $0.2161$  & $0.9078$  & $0.2928$  & $0.7492$  & $1.1781$  & $0.2549$  \\
     Snow  & $0.1417$  & $0.0582$  & $1.0437$  & $0.4411$  & $1.0177$  & $1.3481$  & $0.4713$  \\
    \bottomrule
    \end{tabular}
    \vspace{0.2cm}
    \caption{PETR (r50)~\cite{liu2022petr} results.}
    \label{tab:petr-r50}
\end{table*}

%% file: tables/benchmark/petr-vov.tex
\begin{table*}[t]
    \centering
    \begin{tabular}{l|c|cccccc}
    \toprule
    \textbf{Corruption} & \textbf{NDS $\uparrow$} & \textbf{mAP $\uparrow$} & \textbf{mATE $\downarrow$} & \textbf{mASE $\downarrow$} & \textbf{mAOE $\downarrow$} & \textbf{mAVE $\downarrow$} & \textbf{mAAE $\downarrow$}\\
    \midrule
     Clean  & $0.4550$  & $0.4035$  & $0.7362$  & $0.2710$  & $0.4316$  & $0.8249$  & $0.2039$  \\      
     Cam Crash  & $0.2924$  & $0.1408$  & $0.8167$  & $0.2854$  & $0.5492$  & $0.9014$  & $0.2267$  \\  
     Frame Lost  & $0.2792$  & $0.1153$  & $0.8311$  & $0.2909$  & $0.5662$  & $0.8816$  & $0.2144$  \\ 
     Color Quant  & $0.2968$  & $0.2089$  & $0.8818$  & $0.3455$  & $0.5997$  & $1.0875$  & $0.3123$  \\
     Motion Blur  & $0.2490$  & $0.1395$  & $0.9521$  & $0.3153$  & $0.7424$  & $1.0353$  & $0.2639$  \\
     Brightness  & $0.3858$  & $0.3199$  & $0.7982$  & $0.2779$  & $0.5256$  & $0.9342$  & $0.2112$  \\
     Low Light  & $0.2305$  & $0.1221$  & $0.8897$  & $0.3645$  & $0.6960$  & $1.2311$  & $0.3553$  \\
     Fog  & $0.3703$  & $0.2815$  & $0.8337$  & $0.2778$  & $0.4982$  & $0.8833$  & $0.2111$  \\
     Snow  & $0.2632$  & $0.1653$  & $0.8980$  & $0.3138$  & $0.7034$  & $1.1314$  & $0.2886$  \\
    \bottomrule
    \end{tabular}
    \vspace{0.2cm}
    \caption{PETR (vov)~\cite{liu2022petr} results.}
   \label{tab:petr-vov}
\end{table*}

%% file: tables/benchmark/polarformer-r101.tex
\begin{table*}[t]
    \centering
    \begin{tabular}{l|c|cccccc}
    \toprule
    \textbf{Corruption} & \textbf{NDS $\uparrow$} & \textbf{mAP $\uparrow$} & \textbf{mATE $\downarrow$} & \textbf{mASE $\downarrow$} & \textbf{mAOE $\downarrow$} & \textbf{mAVE $\downarrow$} & \textbf{mAAE $\downarrow$}\\
    \midrule
     Clean  & $0.4602$  & $0.3916$  & $0.7060$  & $0.2718$  & $0.3610$  & $0.8079$  & $0.2093$  \\ 
     Cam Crash  & $0.3133$  & $0.1425$  & $0.7746$  & $0.2840$  & $0.4440$  & $0.8524$  & $0.2250$  \\
     Frame Lost  & $0.2808$  & $0.1134$  & $0.8034$  & $0.3093$  & $0.4981$  & $0.8988$  & $0.2498$  \\
     Color Quant  & $0.3509$  & $0.2538$  & $0.8059$  & $0.2999$  & $0.4812$  & $0.9724$  & $0.2592$  \\
     Motion Blur  & $0.3221$  & $0.2117$  & $0.8196$  & $0.2946$  & $0.5727$  & $0.9379$  & $0.2258$  \\
     Brightness  & $0.4304$  & $0.3574$  & $0.7390$  & $0.2738$  & $0.4149$  & $0.8522$  & $0.2032$  \\
     Low Light  & $0.2554$  & $0.1393$  & $0.8418$  & $0.3557$  & $0.6087$  & $1.2004$  & $0.3364$  \\
     Fog  & $0.4262$  & $0.3518$  & $0.7338$  & $0.2735$  & $0.4143$  & $0.8672$  & $0.2082$  \\
     Snow  & $0.2304$  & $0.1058$  & $0.9125$  & $0.3363$  & $0.6592$  & $1.2284$  & $0.3174$  \\
    \bottomrule
    \end{tabular}
    \vspace{0.2cm}
    \caption{PolarFormer (r101)~\cite{jiang2022polarformer} results.}
    \label{tab:polarformer-r50}
\end{table*}

%% file: tables/benchmark/polarformer-vov.tex
\begin{table*}[t]
    \centering
    \begin{tabular}{l|c|cccccc}
    \toprule
    \textbf{Corruption} & \textbf{NDS $\uparrow$} & \textbf{mAP $\uparrow$} & \textbf{mATE $\downarrow$} & \textbf{mASE $\downarrow$} & \textbf{mAOE $\downarrow$} & \textbf{mAVE $\downarrow$} & \textbf{mAAE $\downarrow$}\\
    \midrule
     Clean  & $0.4558$  & $0.4028$  & $0.7097$  & $0.2690$  & $0.4019$  & $0.8682$  & $0.2072$  \\ 
     Cam Crash  & $0.3135$  & $0.1453$  & $0.7626$  & $0.2815$  & $0.4519$  & $0.8735$  & $0.2216$  \\
     Frame Lost  & $0.2811$  & $0.1155$  & $0.8019$  & $0.3015$  & $0.4956$  & $0.9158$  & $0.2512$  \\
     Color Quant  & $0.3076$  & $0.2000$  & $0.8846$  & $0.2962$  & $0.5393$  & $1.0044$  & $0.2483$  \\
     Motion Blur  & $0.2344$  & $0.1256$  & $0.9392$  & $0.3616$  & $0.6840$  & $1.0992$  & $0.3489$  \\
     Brightness  & $0.4280$  & $0.3619$  & $0.7447$  & $0.2696$  & $0.4413$  & $0.8667$  & $0.2065$  \\
     Low Light  & $0.2441$  & $0.1361$  & $0.8828$  & $0.3647$  & $0.6506$  & $1.2090$  & $0.3419$  \\
     Fog  & $0.4061$  & $0.3349$  & $0.7651$  & $0.2743$  & $0.4487$  & $0.9100$  & $0.2156$  \\
     Snow  & $0.2468$  & $0.1384$  & $0.9104$  & $0.3375$  & $0.6427$  & $1.1737$  & $0.3337$  \\
    \bottomrule
    \end{tabular}
    \vspace{0.2cm}
    \caption{PolarFormer (vov)~\cite{jiang2022polarformer} results.}
    \label{tab:polarformer-vov}
\end{table*}

%% file: tables/benchmark/ora3d.tex
\begin{table*}[t]
    \centering
    \begin{tabular}{l|c|cccccc}
    \toprule
    \textbf{Corruption} & \textbf{NDS $\uparrow$} & \textbf{mAP $\uparrow$} & \textbf{mATE $\downarrow$} & \textbf{mASE $\downarrow$} & \textbf{mAOE $\downarrow$} & \textbf{mAVE $\downarrow$} & \textbf{mAAE $\downarrow$}\\
    \midrule
     Clean  & $0.4436$  & $0.3677$  & $0.7319$  & $0.2698$  & $0.3890$  & $0.8150$  & $0.1975$  \\       
     Cam Crash  & $0.3055$  & $0.1275$  & $0.7952$  & $0.2803$  & $0.4549$  & $0.8376$  & $0.2145$  \\   
     Frame Lost  & $0.2750$  & $0.0997$  & $0.8362$  & $0.3075$  & $0.4963$  & $0.8747$  & $0.2340$  \\  
     Color Quant  & $0.3360$  & $0.2382$  & $0.8479$  & $0.2848$  & $0.5249$  & $0.9516$  & $0.2432$  \\ 
     Motion Blur  & $0.2647$  & $0.1527$  & $0.8656$  & $0.3497$  & $0.6251$  & $1.0433$  & $0.3160$  \\ 
     Brightness  & $0.4075$  & $0.3252$  & $0.7740$  & $0.2741$  & $0.4620$  & $0.8372$  & $0.2029$  \\  
     Low Light  & $0.2613$  & $0.1509$  & $0.8489$  & $0.3445$  & $0.6207$  & $1.2113$  & $0.3278$  \\   
     Fog  & $0.3959$  & $0.3084$  & $0.7822$  & $0.2753$  & $0.4515$  & $0.8685$  & $0.2048$  \\
     Snow  & $0.1898$  & $0.0757$  & $0.9404$  & $0.3857$  & $0.7665$  & $1.2890$  & $0.3879$  \\
    \bottomrule
    \end{tabular}
    \vspace{0.2cm}
    \caption{ORA3D~\cite{roh2022ora3d} results.}
    \label{tab:ora3d-results}
\end{table*}

%% file: tables/benchmark/bevdet-r50.tex
\begin{table*}[t]
    \centering
    \begin{tabular}{l|c|cccccc}
    \toprule
    \textbf{Corruption} & \textbf{NDS $\uparrow$} & \textbf{mAP $\uparrow$} & \textbf{mATE $\downarrow$} & \textbf{mASE $\downarrow$} & \textbf{mAOE $\downarrow$} & \textbf{mAVE $\downarrow$} & \textbf{mAAE $\downarrow$}\\
    \midrule
     Clean  & $0.3770$  & $0.2987$  & $0.7336$  & $0.2744$  & $0.5713$  & $0.9051$  & $0.2394$  \\ 
     Cam Crash  & $0.2486$  & $0.0990$  & $0.8147$  & $0.2975$  & $0.6402$  & $0.9990$  & $0.2842$  \\
     Frame Lost  & $0.1924$  & $0.0781$  & $0.8545$  & $0.4413$  & $0.7179$  & $1.0247$  & $0.4780$  \\
     Color Quant  & $0.2408$  & $0.1542$  & $0.8718$  & $0.3579$  & $0.7376$  & $1.2194$  & $0.3958$  \\
     Motion Blur  & $0.2061$  & $0.1156$  & $0.8891$  & $0.4020$  & $0.7693$  & $1.1521$  & $0.4645$  \\
     Brightness  & $0.2565$  & $0.1787$  & $0.8380$  & $0.3736$  & $0.7216$  & $1.2912$  & $0.3955$  \\
     Low Light  & $0.1102$  & $0.0470$  & $0.9867$  & $0.5308$  & $0.9443$  & $1.2841$  & $0.6708$  \\
     Fog  & $0.2461$  & $0.1404$  & $0.8801$  & $0.3018$  & $0.7483$  & $1.1610$  & $0.3112$  \\
     Snow  & $0.0625$  & $0.0254$  & $0.9853$  & $0.7204$  & $1.0029$  & $1.1642$  & $0.8160$  \\
    \bottomrule
    \end{tabular}
    \vspace{0.2cm}
    \caption{BEVDet (r50)~\cite{huang2021bevdet} results.}
    \label{tab:bevdet-r50-results}
\end{table*}

%% file: tables/benchmark/bevdet-r101.tex
\begin{table*}[t]
    \centering
    \begin{tabular}{l|c|cccccc}
    \toprule
    \textbf{Corruption} & \textbf{NDS $\uparrow$} & \textbf{mAP $\uparrow$} & \textbf{mATE $\downarrow$} & \textbf{mASE $\downarrow$} & \textbf{mAOE $\downarrow$} & \textbf{mAVE $\downarrow$} & \textbf{mAAE $\downarrow$}\\
    \midrule
    Clean  & $0.3877$  & $0.3008$  & $0.7035$  & $0.2752$  & $0.5384$  & $0.8715$  & $0.2379$  \\ 
     Cam Crash  & $0.2622$  & $0.1042$  & $0.7821$  & $0.3004$  & $0.6028$  & $0.9783$  & $0.2715$  \\
     Frame Lost  & $0.2065$  & $0.0805$  & $0.8248$  & $0.4175$  & $0.6754$  & $1.0578$  & $0.4474$  \\
     Color Quant  & $0.2546$  & $0.1566$  & $0.8457$  & $0.3361$  & $0.6966$  & $1.1529$  & $0.3716$  \\
     Motion Blur  & $0.2265$  & $0.1278$  & $0.8596$  & $0.3785$  & $0.7112$  & $1.1344$  & $0.4246$  \\
     Brightness  & $0.2554$  & $0.1738$  & $0.8094$  & $0.3770$  & $0.7228$  & $1.3752$  & $0.4060$  \\
     Low Light  & $0.1118$  & $0.0426$  & $0.9659$  & $0.5550$  & $0.8904$  & $1.3003$  & $0.6836$  \\
     Fog  & $0.2495$  & $0.1412$  & $0.8460$  & $0.3269$  & $0.7007$  & $1.1480$  & $0.3376$  \\
     Snow  & $0.0810$  & $0.0296$  & $0.9727$  & $0.6758$  & $0.9027$  & $1.1803$  & $0.7869$  \\
    \bottomrule
    \end{tabular}
    \vspace{0.2cm}
    \caption{BEVDet (r101)~\cite{huang2021bevdet} results.}
    \label{tab:bevdet-r101-results}
\end{table*}

%% file: tables/benchmark/bevdet-r101-fcos3d.tex
\begin{table*}[t]
    \centering
    \begin{tabular}{l|c|cccccc}
    \toprule
    \textbf{Corruption} & \textbf{NDS $\uparrow$} & \textbf{mAP $\uparrow$} & \textbf{mATE $\downarrow$} & \textbf{mASE $\downarrow$} & \textbf{mAOE $\downarrow$} & \textbf{mAVE $\downarrow$} & \textbf{mAAE $\downarrow$}\\
    \midrule
     Clean  & $0.3780$  & $0.2846$  & $0.7274$  & $0.2796$  & $0.5517$  & $0.8581$  & $0.2264$  \\       
     Cam Crash  & $0.2442$  & $0.0928$  & $0.8020$  & $0.3384$  & $0.5815$  & $1.0285$  & $0.3453$  \\   
     Frame Lost  & $0.1962$  & $0.0720$  & $0.8320$  & $0.4427$  & $0.6830$  & $1.0063$  & $0.4684$  \\  
     Color Quant  & $0.3041$  & $0.2064$  & $0.7815$  & $0.3247$  & $0.6251$  & $0.9955$  & $0.3212$  \\ 
     Motion Blur  & $0.2590$  & $0.1512$  & $0.7826$  & $0.3675$  & $0.6412$  & $1.1481$  & $0.3973$  \\ 
     Brightness  & $0.2599$  & $0.1714$  & $0.7910$  & $0.3963$  & $0.6828$  & $1.1539$  & $0.4242$  \\  
     Low Light  & $0.1393$  & $0.0613$  & $0.8761$  & $0.5631$  & $0.8235$  & $1.1739$  & $0.6510$  \\   
     Fog  & $0.2073$  & $0.0984$  & $0.8521$  & $0.4107$  & $0.6897$  & $1.2659$  & $0.4668$  \\
     Snow  & $0.0939$  & $0.0301$  & $0.9494$  & $0.6685$  & $0.8397$  & $1.2412$  & $0.7535$  \\
    \bottomrule
    \end{tabular}
    \vspace{0.2cm}
    \caption{BEVDet (r101) FCOS3D pre-trained~\cite{huang2021bevdet} results.}
    \label{tab:bevdet-r101-fcos3d-results}
\end{table*}

%% file: tables/benchmark/bevdet-sttiny.tex
\begin{table*}[t]
    \centering
    \begin{tabular}{l|c|cccccc}
    \toprule
    \textbf{Corruption} & \textbf{NDS $\uparrow$} & \textbf{mAP $\uparrow$} & \textbf{mATE $\downarrow$} & \textbf{mASE $\downarrow$} & \textbf{mAOE $\downarrow$} & \textbf{mAVE $\downarrow$} & \textbf{mAAE $\downarrow$}\\
    \midrule
     Clean  & $0.4037$  & $0.3080$  & $0.6648$  & $0.2729$  & $0.5323$  & $0.8278$  & $0.2050$  \\ 
     Cam Crash  & $0.2609$  & $0.1053$  & $0.7786$  & $0.3246$  & $0.5761$  & $0.9821$  & $0.2822$  \\
     Frame Lost  & $0.2115$  & $0.0826$  & $0.8174$  & $0.4207$  & $0.6710$  & $1.0138$  & $0.4294$  \\
     Color Quant  & $0.2278$  & $0.1487$  & $0.8236$  & $0.4518$  & $0.7461$  & $1.1668$  & $0.4742$  \\
     Motion Blur  & $0.2128$  & $0.1235$  & $0.8455$  & $0.4457$  & $0.7074$  & $1.1857$  & $0.5080$  \\
     Brightness  & $0.2191$  & $0.1370$  & $0.8300$  & $0.4523$  & $0.7277$  & $1.2995$  & $0.4833$  \\
     Low Light  & $0.0490$  & $0.0180$  & $0.9883$  & $0.7696$  & $1.0083$  & $1.1225$  & $0.8607$  \\
     Fog  & $0.2450$  & $0.1396$  & $0.8459$  & $0.3656$  & $0.6839$  & $1.2694$  & $0.3520$  \\
     Snow  & $0.0680$  & $0.0312$  & $0.9730$  & $0.7665$  & $0.8973$  & $1.2609$  & $0.8393$  \\
    \bottomrule
    \end{tabular}
    \vspace{0.2cm}
    \caption{BEVDet (swin-tiny)~\cite{huang2021bevdet} results.}
    \label{tab:bevdet-swint-results}
\end{table*}

%% file: tables/benchmark/bevdepth-r50.tex
\begin{table*}[t]
    \centering
    \begin{tabular}{l|c|cccccc}
    \toprule
    \textbf{Corruption} & \textbf{NDS $\uparrow$} & \textbf{mAP $\uparrow$} & \textbf{mATE $\downarrow$} & \textbf{mASE $\downarrow$} & \textbf{mAOE $\downarrow$} & \textbf{mAVE $\downarrow$} & \textbf{mAAE $\downarrow$}\\
    \midrule
     Clean  & $0.4058$  & $0.3328$  & $0.6633$  & $0.2714$  & $0.5581$  & $0.8763$  & $0.2369$  \\ 
     Cam Crash  & $0.2638$  & $0.1111$  & $0.7407$  & $0.2959$  & $0.6373$  & $1.0079$  & $0.2749$  \\
     Frame Lost  & $0.2141$  & $0.0876$  & $0.7890$  & $0.4134$  & $0.6728$  & $1.0536$  & $0.4498$  \\
     Color Quant  & $0.2751$  & $0.1865$  & $0.8190$  & $0.3292$  & $0.6946$  & $1.2008$  & $0.3552$  \\
     Motion Blur  & $0.2513$  & $0.1508$  & $0.8320$  & $0.3516$  & $0.7135$  & $1.1084$  & $0.3765$  \\
     Brightness  & $0.2879$  & $0.2090$  & $0.7520$  & $0.3646$  & $0.6724$  & $1.2089$  & $0.3766$  \\
     Low Light  & $0.1757$  & $0.0820$  & $0.8540$  & $0.4509$  & $0.8073$  & $1.3149$  & $0.5410$  \\
     Fog  & $0.2903$  & $0.1973$  & $0.7900$  & $0.3021$  & $0.6973$  & $1.0640$  & $0.2940$  \\
     Snow  & $0.0863$  & $0.0350$  & $0.9529$  & $0.6682$  & $0.9107$  & $1.2750$  & $0.7802$  \\
    \bottomrule
    \end{tabular}
    \vspace{0.2cm}
    \caption{BEVDepth (r50)~\cite{li2022bevdepth} results.}
    \label{tab:bevdepth-r50-results}
\end{table*}

%% file: tables/benchmark/beverse-tiny.tex
\begin{table*}[t]
    \centering
    \begin{tabular}{l|c|cccccc}
    \toprule
    \textbf{Corruption} & \textbf{NDS $\uparrow$} & \textbf{mAP $\uparrow$} & \textbf{mATE $\downarrow$} & \textbf{mASE $\downarrow$} & \textbf{mAOE $\downarrow$} & \textbf{mAVE $\downarrow$} & \textbf{mAAE $\downarrow$}\\
    \midrule
     Clean  & $0.4665$  & $0.3214$  & $0.6807$  & $0.2782$  & $0.4657$  & $0.3281$  & $0.1893$  \\ 
     Cam Crash  & $0.3181$  & $0.1218$  & $0.7447$  & $0.3545$  & $0.5479$  & $0.4974$  & $0.2833$  \\
     Frame Lost  & $0.3037$  & $0.1466$  & $0.7892$  & $0.3511$  & $0.6217$  & $0.6491$  & $0.2844$  \\
     Color Quant  & $0.2600$  & $0.1497$  & $0.8577$  & $0.4758$  & $0.6711$  & $0.6931$  & $0.4676$  \\
     Motion Blur  & $0.2647$  & $0.1456$  & $0.8139$  & $0.4269$  & $0.6275$  & $0.8103$  & $0.4225$  \\
     Brightness  & $0.2656$  & $0.1512$  & $0.8120$  & $0.4548$  & $0.6799$  & $0.7029$  & $0.4507$  \\
     Low Light  & $0.0593$  & $0.0235$  & $0.9744$  & $0.7926$  & $0.9961$  & $0.9437$  & $0.8304$  \\
     Fog  & $0.2781$  & $0.1348$  & $0.8467$  & $0.3967$  & $0.6135$  & $0.6596$  & $0.3764$  \\
     Snow  & $0.0644$  & $0.0251$  & $0.9662$  & $0.7966$  & $0.8893$  & $0.9829$  & $0.8464$  \\
    \bottomrule
    \end{tabular}
    \vspace{0.2cm}
    \caption{BEVerse (tiny)~\cite{zhang2022beverse} results.}
    \label{tab:beverse-small-results}
\end{table*}

%% file: tables/benchmark/beverse-s-tiny.tex
\begin{table*}[t]
    \centering
    \begin{tabular}{l|c|cccccc}
    \toprule
    \textbf{Corruption} & \textbf{NDS $\uparrow$} & \textbf{mAP $\uparrow$} & \textbf{mATE $\downarrow$} & \textbf{mASE $\downarrow$} & \textbf{mAOE $\downarrow$} & \textbf{mAVE $\downarrow$} & \textbf{mAAE $\downarrow$}\\
    \midrule
     Clean  & $0.1603$  & $0.0826$  & $0.8298$  & $0.5296$  & $0.8771$  & $1.2639$  & $0.5739$  \\ 
     Cam Crash  & $0.0639$  & $0.0165$  & $0.9135$  & $0.7574$  & $0.9522$  & $1.1890$  & $0.8201$  \\
     Frame Lost  & $0.0508$  & $0.0141$  & $0.9455$  & $0.8181$  & $0.9221$  & $1.1765$  & $0.8765$  \\
     Color Quant  & $0.0642$  & $0.0317$  & $0.9478$  & $0.7735$  & $0.9723$  & $1.2508$  & $0.8397$  \\
     Motion Blur  & $0.0540$  & $0.0230$  & $0.9556$  & $0.8028$  & $0.9339$  & $1.2137$  & $0.8826$  \\
     Brightness  & $0.0683$  & $0.0360$  & $0.9369$  & $0.7315$  & $0.9878$  & $1.3048$  & $0.8531$  \\
     Low Light  & $0.0100$  & $0.0005$  & $1.0097$  & $0.9474$  & $1.0048$  & $1.1073$  & $0.9561$  \\
     Fog  & $0.0402$  & $0.0179$  & $0.9789$  & $0.8230$  & $1.0094$  & $1.3083$  & $0.8962$  \\
     Snow  & $0.0107$  & $0.0017$  & $1.0021$  & $0.9468$  & $0.9968$  & $1.1652$  & $0.9612$  \\
    \bottomrule
    \end{tabular}
    \vspace{0.2cm}
    \caption{BEVerse-SingleFrame (tiny)~\cite{zhang2022beverse} results.}
    \label{tab:beverse-s-tiny-results}
\end{table*}

%% file: tables/benchmark/beverse-small.tex
\begin{table*}[t]
    \centering
    \begin{tabular}{l|c|cccccc}
    \toprule
    \textbf{Corruption} & \textbf{NDS $\uparrow$} & \textbf{mAP $\uparrow$} & \textbf{mATE $\downarrow$} & \textbf{mASE $\downarrow$} & \textbf{mAOE $\downarrow$} & \textbf{mAVE $\downarrow$} & \textbf{mAAE $\downarrow$}\\
    \midrule
     Clean  & $0.4951$  & $0.3512$  & $0.6243$  & $0.2694$  & $0.3999$  & $0.3292$  & $0.1827$  \\ 
     Cam Crash  & $0.3364$  & $0.1156$  & $0.6753$  & $0.3331$  & $0.4460$  & $0.4823$  & $0.2772$  \\
     Frame Lost  & $0.2485$  & $0.0959$  & $0.7413$  & $0.4389$  & $0.5898$  & $0.8170$  & $0.4445$  \\
     Color Quant  & $0.2807$  & $0.1630$  & $0.8148$  & $0.4651$  & $0.6311$  & $0.6511$  & $0.4455$  \\
     Motion Blur  & $0.2632$  & $0.1455$  & $0.7866$  & $0.4399$  & $0.5753$  & $0.8424$  & $0.4586$  \\
     Brightness  & $0.3394$  & $0.1935$  & $0.7441$  & $0.3736$  & $0.4873$  & $0.6357$  & $0.3326$  \\
     Low Light  & $0.1118$  & $0.0373$  & $0.9230$  & $0.6900$  & $0.8727$  & $0.8600$  & $0.7223$  \\
     Fog  & $0.2849$  & $0.1291$  & $0.7858$  & $0.4234$  & $0.5105$  & $0.6852$  & $0.3921$  \\
     Snow  & $0.0985$  & $0.0357$  & $0.9309$  & $0.7389$  & $0.8864$  & $0.8695$  & $0.7676$  \\
    \bottomrule
    \end{tabular}
    \vspace{0.2cm}
    \caption{BEVerse (small)~\cite{zhang2022beverse} results.}
    \label{tab:beverse-tiny-results}
\end{table*}

%% file: tables/benchmark/beverse-s-small.tex
\begin{table*}[t]
    \centering
    \begin{tabular}{l|c|cccccc}
    \toprule
    \textbf{Corruption} & \textbf{NDS $\uparrow$} & \textbf{mAP $\uparrow$} & \textbf{mATE $\downarrow$} & \textbf{mASE $\downarrow$} & \textbf{mAOE $\downarrow$} & \textbf{mAVE $\downarrow$} & \textbf{mAAE $\downarrow$}\\
    \midrule
     Clean  & $0.2682$  & $0.1513$  & $0.6631$  & $0.4228$  & $0.5406$  & $1.3996$  & $0.4483$  \\ 
     Cam Crash  & $0.1305$  & $0.0340$  & $0.8028$  & $0.6164$  & $0.7475$  & $1.2273$  & $0.6978$  \\
     Frame Lost  & $0.0822$  & $0.0274$  & $0.8755$  & $0.7651$  & $0.8674$  & $1.1223$  & $0.8107$  \\
     Color Quant  & $0.1002$  & $0.0495$  & $0.8923$  & $0.7228$  & $0.8517$  & $1.1570$  & $0.7850$  \\
     Motion Blur  & $0.0716$  & $0.0370$  & $0.9117$  & $0.7927$  & $0.8818$  & $1.1616$  & $0.8833$  \\
     Brightness  & $0.1336$  & $0.0724$  & $0.8340$  & $0.6499$  & $0.8086$  & $1.2874$  & $0.7333$  \\
     Low Light  & $0.0132$  & $0.0041$  & $0.9862$  & $0.9356$  & $1.0175$  & $0.9964$  & $0.9707$  \\
     Fog  & $0.0910$  & $0.0406$  & $0.8894$  & $0.7200$  & $0.8700$  & $1.0564$  & $0.8140$  \\
     Snow  & $0.0116$  & $0.0066$  & $0.9785$  & $0.9385$  & $1.0000$  & $1.0000$  & $1.0000$  \\
    \bottomrule
    \end{tabular}
    \vspace{0.2cm}
    \caption{BEVerse-SingleFrame (small)~\cite{zhang2022beverse} results.}
    \label{tab:beverse-s-small-results}
\end{table*}

%% file: tables/benchmark/srcn3d-r101.tex
\begin{table*}[t]
    \centering
    \begin{tabular}{l|c|cccccc}
    \toprule
    \textbf{Corruption} & \textbf{NDS $\uparrow$} & \textbf{mAP $\uparrow$} & \textbf{mATE $\downarrow$} & \textbf{mASE $\downarrow$} & \textbf{mAOE $\downarrow$} & \textbf{mAVE $\downarrow$} & \textbf{mAAE $\downarrow$}\\
    \midrule
     Clean  & $0.4286$  & $0.3373$  & $0.7783$  & $0.2873$  & $0.3665$  & $0.7806$  & $0.1878$  \\ 
     Cam Crash  & $0.2947$  & $0.1172$  & $0.8369$  & $0.3017$  & $0.4403$  & $0.8506$  & $0.2097$  \\
     Frame Lost  & $0.2681$  & $0.0924$  & $0.8637$  & $0.3303$  & $0.4798$  & $0.8725$  & $0.2349$  \\
     Color Quant  & $0.3318$  & $0.2199$  & $0.8696$  & $0.3041$  & $0.4747$  & $0.8877$  & $0.2458$  \\
     Motion Blur  & $0.2609$  & $0.1361$  & $0.9026$  & $0.3524$  & $0.5788$  & $0.9964$  & $0.2927$  \\
     Brightness  & $0.4074$  & $0.3133$  & $0.7936$  & $0.2911$  & $0.3974$  & $0.8227$  & $0.1877$  \\
     Low Light  & $0.2590$  & $0.1406$  & $0.8586$  & $0.3642$  & $0.5773$  & $1.1257$  & $0.3353$  \\
     Fog  & $0.3940$  & $0.2932$  & $0.7993$  & $0.2919$  & $0.3978$  & $0.8428$  & $0.1944$  \\
     Snow  & $0.1920$  & $0.0734$  & $0.9372$  & $0.3996$  & $0.7302$  & $1.2366$  & $0.3803$  \\
    \bottomrule
    \end{tabular}
    \vspace{0.2cm}
    \caption{SRCN3D (r101)~\cite{shi2022srcn3d} results.}
    \label{tab:srcn3d-r101}
\end{table*}

%% file: tables/benchmark/srcn3d-vov.tex
\begin{table*}[t]
    \centering
    \begin{tabular}{l|c|cccccc}
    \toprule
    \textbf{Corruption} & \textbf{NDS $\uparrow$} & \textbf{mAP $\uparrow$} & \textbf{mATE $\downarrow$} & \textbf{mASE $\downarrow$} & \textbf{mAOE $\downarrow$} & \textbf{mAVE $\downarrow$} & \textbf{mAAE $\downarrow$}\\
    \midrule
     Clean  & $0.4205$  & $0.3475$  & $0.7855$  & $0.2994$  & $0.4099$  & $0.8352$  & $0.2030$  \\ 
     Cam Crash  & $0.2875$  & $0.1252$  & $0.8435$  & $0.3139$  & $0.4879$  & $0.8897$  & $0.2165$  \\
     Frame Lost  & $0.2579$  & $0.0982$  & $0.8710$  & $0.3428$  & $0.5324$  & $0.9194$  & $0.2458$  \\
     Color Quant  & $0.2827$  & $0.1755$  & $0.9167$  & $0.3443$  & $0.5574$  & $1.0077$  & $0.2747$  \\
     Motion Blur  & $0.2143$  & $0.1102$  & $0.9833$  & $0.3966$  & $0.7434$  & $1.1151$  & $0.3500$  \\
     Brightness  & $0.3886$  & $0.3086$  & $0.8175$  & $0.3018$  & $0.4660$  & $0.8720$  & $0.2001$  \\
     Low Light  & $0.2274$  & $0.1142$  & $0.9192$  & $0.3866$  & $0.6475$  & $1.2095$  & $0.3435$  \\
     Fog  & $0.3774$  & $0.2911$  & $0.8227$  & $0.3045$  & $0.4646$  & $0.8864$  & $0.2034$  \\
     Snow  & $0.2499$  & $0.1418$  & $0.9299$  & $0.3575$  & $0.6125$  & $1.1351$  & $0.3176$  \\
    \bottomrule
    \end{tabular}
    \vspace{0.2cm}
    \caption{SRCN3D (vov)~\cite{shi2022srcn3d} results.}
    \label{tab:srcn3d-vov}
\end{table*}

%% file: tables/benchmark/sparse4d.tex
\begin{table*}[t]
    \centering
    \begin{tabular}{l|c|cccccc}
    \toprule
    \textbf{Corruption} & \textbf{NDS $\uparrow$} & \textbf{mAP $\uparrow$} & \textbf{mATE $\downarrow$} & \textbf{mASE $\downarrow$} & \textbf{mAOE $\downarrow$} & \textbf{mAVE $\downarrow$} & \textbf{mAAE $\downarrow$}\\
    \midrule
     Clean  & $0.5438$  & $0.4409$  & $0.6282$  & $0.2721$  & $0.3853$  & $0.2922$  & $0.1888$  \\ 
     Cam Crash  & $0.2873$  & $0.1319$  & $0.7852$  & $0.2917$  & $0.4989$  & $0.9611$  & $0.2510$  \\
     Frame Lost  & $0.2611$  & $0.1050$  & $0.8175$  & $0.3166$  & $0.5404$  & $1.0253$  & $0.2726$  \\
     Color Quant  & $0.3310$  & $0.2345$  & $0.8348$  & $0.2956$  & $0.5452$  & $0.9712$  & $0.2496$  \\
     Motion Blur  & $0.2514$  & $0.1438$  & $0.8719$  & $0.3553$  & $0.6780$  & $1.0817$  & $0.3347$  \\
     Brightness  & $0.3984$  & $0.3296$  & $0.7543$  & $0.2835$  & $0.4844$  & $0.9232$  & $0.2187$  \\
     Low Light  & $0.2510$  & $0.1386$  & $0.8501$  & $0.3543$  & $0.6464$  & $1.1621$  & $0.3356$  \\
     Fog  & $0.3884$  & $0.3097$  & $0.7552$  & $0.2840$  & $0.4933$  & $0.9087$  & $0.2229$  \\
     Snow  & $0.2259$  & $0.1275$  & $0.8860$  & $0.3875$  & $0.7116$  & $1.1418$  & $0.3936$  \\
    \bottomrule
    \end{tabular}
    \vspace{0.2cm}
    \caption{Sparse4D~\cite{lin2022sparse4d} results.}
    \label{tab:sparse4d-results}
\end{table*}

%% file: tables/benchmark/solofusion-s.tex
\begin{table*}[t]
    \centering
    \begin{tabular}{l|c|cccccc}
    \toprule
    \textbf{Corruption} & \textbf{NDS $\uparrow$} & \textbf{mAP $\uparrow$} & \textbf{mATE $\downarrow$} & \textbf{mASE $\downarrow$} & \textbf{mAOE $\downarrow$} & \textbf{mAVE $\downarrow$} & \textbf{mAAE $\downarrow$}\\
    \midrule
     Clean  & $0.3907$  & $0.3438$  & $0.6691$  & $0.2809$  & $0.6638$  & $0.8803$  & $0.3180$  \\       
     Cam Crash  & $0.2541$  & $0.1132$  & $0.7542$  & $0.2848$  & $0.7337$  & $0.9248$  & $0.3273$  \\   
     Frame Lost  & $0.2195$  & $0.0848$  & $0.8066$  & $0.3285$  & $0.7407$  & $1.0092$  & $0.3785$  \\  
     Color Quant  & $0.2804$  & $0.2013$  & $0.7790$  & $0.3214$  & $0.7702$  & $0.9825$  & $0.3706$  \\ 
     Motion Blur  & $0.2603$  & $0.1717$  & $0.8145$  & $0.2968$  & $0.8353$  & $0.9831$  & $0.3414$  \\ 
     Brightness  & $0.2966$  & $0.2339$  & $0.7497$  & $0.3258$  & $0.8038$  & $1.0663$  & $0.3433$  \\
     Low Light  & $0.2033$  & $0.1138$  & $0.7744$  & $0.3716$  & $0.9146$  & $1.1518$  & $0.4757$  \\
     Fog  & $0.2998$  & $0.2260$  & $0.7556$  & $0.2908$  & $0.7761$  & $1.0074$  & $0.3238$  \\
     Snow  & $0.1066$  & $0.0427$  & $0.9399$  & $0.5888$  & $0.9026$  & $1.1212$  & $0.7160$  \\
    \bottomrule
    \end{tabular}
    \vspace{0.2cm}
    \caption{SOLOFusion (short)~\cite{Park2022TimeWT} results.}
    \label{tab:solofusion-s}
\end{table*}

%% file: tables/benchmark/solofusion-l.tex
\begin{table*}[t]
    \centering
    \begin{tabular}{l|c|cccccc}
    \toprule
    \textbf{Corruption} & \textbf{NDS $\uparrow$} & \textbf{mAP $\uparrow$} & \textbf{mATE $\downarrow$} & \textbf{mASE $\downarrow$} & \textbf{mAOE $\downarrow$} & \textbf{mAVE $\downarrow$} & \textbf{mAAE $\downarrow$}\\
    \midrule
     Clean  & $0.4850$  & $0.3862$  & $0.6292$  & $0.2840$  & $0.6387$  & $0.3151$  & $0.2141$  \\       
     Cam Crash  & $0.3159$  & $0.1173$  & $0.7462$  & $0.2938$  & $0.6939$  & $0.4614$  & $0.2327$  \\   
     Frame Lost  & $0.2490$  & $0.1121$  & $0.7824$  & $0.3529$  & $0.8133$  & $0.8167$  & $0.3249$  \\  
     Color Quant  & $0.3598$  & $0.2233$  & $0.7704$  & $0.3206$  & $0.7326$  & $0.4266$  & $0.2681$  \\ 
     Motion Blur  & $0.3460$  & $0.1969$  & $0.7765$  & $0.2973$  & $0.7849$  & $0.4262$  & $0.2395$  \\ 
     Brightness  & $0.4002$  & $0.2726$  & $0.7163$  & $0.3113$  & $0.6754$  & $0.4251$  & $0.2328$  \\  
     Low Light  & $0.2814$  & $0.1301$  & $0.7669$  & $0.3701$  & $0.7913$  & $0.5548$  & $0.3534$  \\   
     Fog  & $0.3991$  & $0.2570$  & $0.7230$  & $0.2947$  & $0.7084$  & $0.3678$  & $0.2002$  \\
     Snow  & $0.1480$  & $0.0590$  & $0.8901$  & $0.5666$  & $0.9179$  & $0.7932$  & $0.6480$  \\
    \bottomrule
    \end{tabular}
    \vspace{0.2cm}
    \caption{SOLOFusion (long)~\cite{Park2022TimeWT} results.}
    \label{tab:solofusion-l}
\end{table*}

%% file: tables/benchmark/solofusion.tex
\begin{table*}[t]
    \centering
    \begin{tabular}{l|c|cccccc}
    \toprule
    \textbf{Corruption} & \textbf{NDS $\uparrow$} & \textbf{mAP $\uparrow$} & \textbf{mATE $\downarrow$} & \textbf{mASE $\downarrow$} & \textbf{mAOE $\downarrow$} & \textbf{mAVE $\downarrow$} & \textbf{mAAE $\downarrow$}\\
    \midrule
     Clean  & $0.5381$  & $0.4299$  & $0.5842$  & $0.2747$  & $0.4564$  & $0.2426$  & $0.2103$  \\ 
     Cam Crash  & $0.3806$  & $0.1590$  & $0.6607$  & $0.2773$  & $0.5186$  & $0.3152$  & $0.2176$  \\
     Frame Lost  & $0.3464$  & $0.1671$  & $0.7161$  & $0.3042$  & $0.5557$  & $0.5292$  & $0.2668$  \\
     Color Quant  & $0.4058$  & $0.2572$  & $0.6910$  & $0.3200$  & $0.6217$  & $0.3434$  & $0.2514$  \\
     Motion Blur  & $0.3642$  & $0.2019$  & $0.7191$  & $0.3244$  & $0.6643$  & $0.3834$  & $0.2762$  \\
     Brightness  & $0.4329$  & $0.2959$  & $0.6532$  & $0.3238$  & $0.5353$  & $0.3808$  & $0.2577$  \\
     Low Light  & $0.2626$  & $0.1237$  & $0.7258$  & $0.4567$  & $0.7598$  & $0.5910$  & $0.4597$  \\
     Fog  & $0.4480$  & $0.2923$  & $0.6502$  & $0.2883$  & $0.5496$  & $0.2958$  & $0.1973$  \\
     Snow  & $0.1376$  & $0.0561$  & $0.8722$  & $0.6480$  & $0.8219$  & $0.8363$  & $0.7255$  \\
    \bottomrule
    \end{tabular}
    \vspace{0.2cm}
    \caption{SOLOFusion~\cite{Park2022TimeWT} results.}
    \label{tab:solofusion}
\end{table*}

%% file: tables/benchmark/fcos3d.tex
 \begin{table*}[t]
    \centering
    \begin{tabular}{l|c|cccccc}
    \toprule
    \textbf{Corruption} & \textbf{NDS $\uparrow$} & \textbf{mAP $\uparrow$} & \textbf{mATE $\downarrow$} & \textbf{mASE $\downarrow$} & \textbf{mAOE $\downarrow$} & \textbf{mAVE $\downarrow$} & \textbf{mAAE $\downarrow$}\\
    \midrule
     Clean  & $0.3949$  & $0.3214$  & $0.7538$  & $0.2603$  & $0.4864$  & $1.3321$  & $0.1574$  \\       
     Cam Crash  & $0.2849$  & $0.1169$  & $0.7842$  & $0.2693$  & $0.5134$  & $1.2993$  & $0.1684$  \\   
     Frame Lost  & $0.2479$  & $0.0915$  & $0.7912$  & $0.3521$  & $0.5367$  & $1.3668$  & $0.2989$  \\  
     Color Quant  & $0.2574$  & $0.1548$  & $0.8851$  & $0.3631$  & $0.6378$  & $1.3906$  & $0.3157$  \\ 
     Motion Blur  & $0.2570$  & $0.1459$  & $0.8460$  & $0.3318$  & $0.6894$  & $1.2404$  & $0.2920$  \\ 
     Brightness  & $0.3218$  & $0.2237$  & $0.8243$  & $0.2801$  & $0.6179$  & $1.4902$  & $0.1778$  \\  
     Low Light  & $0.1468$  & $0.0491$  & $0.8845$  & $0.5287$  & $0.7911$  & $1.3388$  & $0.5729$  \\   
     Fog  & $0.3321$  & $0.2319$  & $0.8087$  & $0.2702$  & $0.5719$  & $1.2989$  & $0.1879$  \\
     Snow  & $0.1136$  & $0.0448$  & $0.9656$  & $0.6321$  & $0.7768$  & $1.2827$  & $0.7141$  \\
    \bottomrule
    \end{tabular}
    \vspace{0.2cm}
    \caption{FCOS3D~\cite{wang2021fcos3d} results.}
    \label{tab:fcos3d-results}
\end{table*}